\documentclass[final,5p,times,twocolumn,authoryear]{elsarticle} 
\DeclareUnicodeCharacter{2212}{-}
\PassOptionsToPackage{numbers}{natbib}

\usepackage{amssymb}
\usepackage{lipsum} 
\usepackage{hyperref}
\usepackage{amsmath}
\usepackage{graphicx}
\usepackage{subcaption}
\usepackage{booktabs} 
\usepackage{array}
\usepackage{lscape}
\usepackage{siunitx}
\usepackage{upgreek}

\journal{arXiv preprint}

\setcitestyle{numbers}
\begin{document}

\begin{frontmatter}



\title{Traditional and Recent Approaches for Infrared Small Target Detection: Overview, Challenges and Perspectives (Part 1)}


\author[first]{Thierry Bouwmans}
\author[first]{Anastasia Zakharova}

\affiliation[first]{
    departement={Laboratoire MIA},
    organization={La Rochelle Université},
    city={La Rochelle},
    postcode={17000}, 
    country={France}
}

\author[second]{Belmar Garcia}
\affiliation[second]{
    departement={ISEN Toulon},
    city={Toulon},
    postcode={83000}, 
   country={France}
}

\author[third]{Marwa Chendeb El Rai}
\affiliation[third]{
    departement={Mathematics Division},
    organization={American University in Dubai},
    city={Dubai},
    country={United Arab Emirates}
}

\author[fourth]{Meghna Kapoor}
\affiliation[fourth]{
    departement={Laboratoire L3i},
    organization={La Rochelle Université},
    city={La Rochelle},
    postcode={17000}, 
    country={France}
}

\author[fifth]{Badri Narayan Subudhi}
\affiliation[fifth]{
    departement={Department of Electrical Engineering},
    organization={IIT Jammu},
    city={Jagti},
    postcode={181221}, 
    country={India}
}

\begin{abstract}
Infrared small target detection is often used in different civilian and military applications to  detect static or moving targets in scenes like in ground, air and sea surveillance. Research in this field started more than three decades ago, and the last two decades have witnessed very significant publications in this field following the rise of low-rank and sparse decomposition (model-based methods) and deep learning models (data-based methods).  Furthermore, several surveys can be found in the literature but none of them addresses an overall review in this field. In this context, the aim of this paper is to provide a complete survey of the traditional and recent approaches in infrared small target detection. First, we categorize the different approaches found in the literature in terms of the modeling models, i.e mathematical models, machine learning models and signal processing models. Then, we have discussed them in terms of the critical situations that they claim to handle. Furthermore, we present the available resources, datasets and libraries. Then, we conclude with several promising directions for future research. For the sake of reading, this survey is divided into two parts: 1) "Traditional and Recent Approaches for Infrared Small Target Detection: Overview,
Challenges and Perspectives (Part 1)" about conventional and low-rank and sparse decomposition (model-based methods), and 2) "Traditional and Recent Approaches for Infrared Small Target Detection: Overview, Challenges and Perspectives (Part 2)" about deep learning models (data-based methods).
\end{abstract}



\begin{keyword}
Small Target Detection \sep Infrared Imaging \sep Local/Global Contrast Measurement \sep Low-rank plus Sparse Decomposition \sep Tensor Decomposition \sep Performance Evaluation



\end{keyword}

\end{frontmatter}



\section{Introduction}
\label{introduction}
Infrared small target detection is a crucial task in military and civilian fields using Infrared Search and Track (IRST), Forward Looking Infrared System (FLIR), Space-based Infrared System (SBIRS), Space Tracking and Surveillance Systems (STSS), early-warning systems, and missile tracking systems. All these applications require to detect target as early as possible and the size of target is often with only a few pixels because of long imaging distance.  Thus, infrared target detection plays a pivotal role in various long-distance detection and surveillance tasks owing to its exceptional sensitivity to thermal radiation and independence from illumination conditions. However, the infrared target radiation energy received by imaging sensor is very low and with dim spotlike feature because of the atmospheric refraction, dispersion and absorption. Thus, infrared small target detection in complex clutter background is difficult because of a low Signal-to-Clutter Ratio (SCR), and a lack of concrete shape or texture information. In this context, Infrared Small Target Detection (IRSTD) is a significant research field with high challenges. In literature, this field is also reffered as infrared dim small target detection \cite{SFD-Miscellaneous-5}, infrared weak target detection \cite{SFD-TH-13}, infrared tiny target detection \cite{DL-599-191,Applications-S-40} and far infrared small target detection \cite{Miscellaneous-300}. When detection is performed on single frames, as in most works, the task is called Single-frame InfraRed Small Target (SIRST) detection. When the targets are in motion, it is also called Moving Infrared Small Target detection (MIRST) \cite{DL-38,DL-67,DL-117,DL-199,DL-241,DL-261,DL-271,DL-275,VLM-1,GM-2}. \\

\indent Two closely related problems to infrared small target detection are closely-spaced infrared small target unmixing \cite{DL-305,DL-307,DL-315,DL-332,DL-599-375} and the moving object detection task using background subtraction in video surveillance \cite{MOD-Survey-1,MOD-Survey-1-1} with visible spectrum or IR sensors.  In closely-spaced infrared small target unmixing, the radiation intensity obtained from remote targets is inherently weak due to their long-range distance from the imaging system. This challenge is exacerbated when targets appear in spatially close dense clusters, as the Closely-Spaced Objects (CSO), resulting in overlapping blob-like spots. The core of CSO is to recover the intrinsic attributes of each constituent target from the mixed spot, which requires the simultaneous achievement of three objectives \cite{DL-315}: sub-pixel target detection, sub-pixel localization, and recovery of intrinsic target information. Instead of IRSTD that focuses on pixel-level target detection, CSO requires accurate estimate of the number of targets within the mixed light spot rather than mere target-level detection. Moreover, distinct from IRSTD which achieves pixel-level target localization, CSO determines the sub-pixel coordinates of each individual target. Using the signal processing theory, CSO is formulated within the framework of Compressed Sensing (CS) and reconstruction. It aims to reconstruct a high resolution target distribution from low-resolution observations. Recent advances concern the deep neural networks approaches \cite{DL-305,DL-307}. In the moving object detection task, the background can be modeled and the size of the moving object is sufficient to extract the foreground comparing the background with the current image \cite{MOD-Survey-200}. The background model is based on either mathematical tools (statistical  concepts \cite{MOD-Survey-2}, fuzzy concepts \cite{MOD-Method-3,MOD-Survey-3}), machine learning tools (subspace learning \cite{MOD-Method-4,MOD-Survey-4}, deep neural networks \cite{MOD-Method-5,MOD-Survey-5}) or signal processing tools (graph signal processing \cite{MOD-Method-6,MOD-Survey-6}. Even if several approaches have been developed in the context of infrared moving object detection to model the background allowing to detect persons \cite{MOD-Thermal-1}, cars \cite{MOD-Thermal-2} or boats \cite{MOD-Thermal-3} in challenging environments, none of these methods are suitable in the case of infrared small target detection \cite{MOD-Survey-1-1}. However, several methods address the challenge of small moving object detection \cite{MOD-Small-1,MOD-Small-2,MOD-Small-3,DL-106} in the visible spectrum and the size is bigger than the size defined by the Society of Photo-Optical Instrumentation Engineers (SPIE) about small target in infrared images. Finally, small target motion detection methods based on bio-inspired visual networks \cite{DL-599-169,DL-599-169-1,DL-599-169-2,DL-599-169-3,DL-599-169-4,DL-599-169-5,DL-599-169-6} works well in visible spectrum and are not adaptable for IRSTD. In summary, these two close related fields employ dedicated methods that can not be used in IRSTD.

\subsection{Related Works}
\label{Related_Works}
In literature, the infrared small target detection has thus attracted numerous works since the last three decades with several key periods related to the rise of sensors in more diverse environments and to the mathematical, machine learning and signal processing tools employed to deal with the met challenges. In this context, several surveys \cite{Survey-1,Survey-2,Survey-3,Survey-4,Survey-5,Survey-6,Survey-7} about infrared small target detection can be found in the literature but none of them address an overall review in this field. Indeed, Bai et al. \cite{Survey-1} survey in 2011 wavelet based algorithms, inter-frame difference based algorithms and filter based algorithms. In 2019, Zhang and Peng  \cite{Survey-2} review low-rank and sparse decomposition models. In 2019, Eysa and Hamdulla \cite{Survey-3} focus on typical algorithm principles and characteristics for infrared small target detection and tracking. In 2022, Zhao et al.  \cite{Survey-5} review only single frame models. In 2024, Cheng et al. \cite{Survey-6} focus on deep learning methods from the past three years and classify them according to six key issues: 1) enhancing the representation capability of small targets, 2) improving the accuracy of bounding box regression, 3) resolving the issue of target information loss in the deep network, 4) balancing missed detections and false alarms, 5) adapting for complex backgrounds and 6) lightweight design and deployment issues of the network. In 2025,  Kumar and  Singh \cite{Survey-7} provide a comprehensive survey but the survey reviews only 90 papers lacking in terms of exhaustivity, misses key recent future works based on graph learning representation and misses also key issues about performance evaluation. Table \ref{OverviewSurvey} compares these surveys with the proposed survey highlighting our contributions. \\

Note that the small space target detection \cite{DL-600} and the infrared small target tracking task \cite{DL-562,DL-599-73} are out of the scope of this review. 

\begin{table}
\scalebox{0.79}{
\begin{tabular}{|l|l|}
\hline
\scriptsize{Surveys [Authors (Date)]} &\scriptsize{Surveyed Categories (Number of Reviewed Papers)}  \\
\hline
\hline
\scriptsize{Bai et al. \cite{Survey-1} (2011)} &\scriptsize{Traditional Unsupervised Models (30 papers)} \\
\scriptsize{Zhang and Peng. \cite{Survey-2} (2019)} &\scriptsize{Low-rank and Sparse Decomposition Models (70 papers)} \\
\scriptsize{Eysa and  Hamdulla \cite{Survey-3} (2019)} &\scriptsize{Typical Algorithm Principles and Characteristics} \\
\scriptsize{Zhao et al. \cite{Survey-5} (2022)} &\scriptsize{Single Frame Models} \\
\scriptsize{Cheng et al. \cite{Survey-6} (2024)} &\scriptsize{Deep Learning Supervised Models (90 papers) } \\
\scriptsize{Kumar and  Singh \cite{Survey-7} (2025)} &\scriptsize{All Categories (90 papers) + Datasets} \\
\scriptsize{\textbf{This Survey}} &\scriptsize{All Categories (1035 papers) + Datasets + Performance Evaluation Metrics} \\
\hline
\end{tabular}}
\caption{Existing Surveys in Infrared Small Target Detection: An Overview.}
\label{OverviewSurvey}
\end{table}

\subsection{Motivation and Contributions}
Infrared small target detection has emerged as a topic of interest 30 years ago with the development of infrared sensors which are more and more efficient and deployable over time. In addition, machine learning methods has also witnessed key developments with model-based models and data-based models to deal with diverse kinds of data. More specifically, with the rise of deep learning, infrared small target detection has made rapid progress in diverse civilian and military fields like marine resource, high-precision navigation, autonomous driving, and ecological environment monitoring. With the growing demands for edge-side applications like micro-satellites, micro-UAVs, autonomous vehicles and ships, existing methods fails to achieve efficient response in edge-side scenarios with limited annotated samples and computational resources. \\ 

In this context, the aim of this survey is then to provide a first complete overview of all infrared small target detection methods over more than 1000 papers from the origin until 2026  for (1) novices who could be students or engineers beginning in the field of remote sensing, (2) experts as we put forward the recent advances that need to be improved, and (3) reviewers to evaluate papers in journals (IEEE Transactions on Geoscience and Remote Sensing, IEEE Geoscience and Remote Sensing Letters, IEEE Journal of Selected Topics in Applied Earth Observations and Remote Sensing, ELSEVIER Remote Sensing of Environment, MDPI Remote Sensing, etc.) and conferences (IEEE International Geoscience and Remote Sensing Symposium, etc.). So, this survey is intended to be a reference for researchers and developers in industries, as well as graduate students, interested in infrared small target detection methods applied to ground, air and sea environments. It can also be suggested as a reading text to teach graduate courses in subjects such as computer vision and remote sensing. \\

Although there are numerous works on infrared small target detection and the fact that it can give the feeling that this issue is solved or is over-studied, no algorithm today seems to be able to simultaneously address all the key challenges that we can meet in infrared small target detection and the substantial progress remain constrained \cite{DL-334}. This is due to the four main reasons:
\begin{enumerate}
\item The lack of common framework, that is, each method was developed in several different contexts such as ground vehicles, airplanes and ship detection.
\item  The lack of scientific progress, that is, there are mainly improvements of the state-of-art methods such as saliency based methods, low-rank plus sparse decomposition, and deep neural networks but recent representation models are investigated insufficiently.
\item The absence of a single realistic large-scale dataset with accurate ground-truth providing a balanced coverage of the range of challenges present in the real world.
\item The last reason for this technological stagnation is the fundamental bias in the current evaluation protocol: A critical “evaluation gap” in
identifying the intrinsic difficulty of detection tasks. Indeed, existing research has mainly focused on post detection performance
metrics. Even if needed for benchmarking final outputs, these metrics are posterior and result-oriented. Thus, they can not quantify the a priori difficulty of the input data, leaving the field overly reliant on the SCR as the sole authoritative indicator of difficulty \cite{DL-334}.
\end{enumerate}
These four problems have been partially addressed by (1) some recent surveys on the field \cite{Survey-6,Survey-7} but limited to deep learning methods, (2) the application of graph-based learning tools \cite{GM-1,GM-2,GM-3,GM-4,DL-291}, (3) the creation of large-scale datasets such as the IRDST dataset \cite{DL-33}, the TensorISTD dataset \cite{PE-2}, the NCHU-Seg dataset \cite{Dataset-700} and the IRSTDID dataset \cite{Dataset-600} and (4)  new evaluation protocols based on hybrid level metrics \cite{PE-1} and a target detectability metric \cite{DL-334}. In addition of the annual top ranked IEEE IGARSS conferences, a ICPR 2024 Resource-Limited Infrared Small Target Detection Challenge (LimitIRSTD\footnote{https://limitirstd.github.io/}) facilitates the development of infrared small object detection algorithms under limited resources including two tracks about weakly supervised algorithms under single point supervision and lightweight algorithms under pixel-level supervision. \\

In this context, we believe that we are living in a key transition in the field of infrared small target detection and the aim of this survey is to review these key new points. Thus, it reviews all the models since the first works in the field to the recent ones. By reviewing both existing and new ideas, this survey gives a complete overview of the concepts, theories, algorithms, and applications related to infrared small target detection. First, an infrared small target detection definition for beginners is provided. Then, we review the different challenges and which steps and issues are concerned to deal with them. Furthermore, existing and recent methods for infrared small target detection are presented. A description of recent complete datasets and codes are given. Moreover, an accompanying website called the Target Detection Website\footnote{https://sites.google.com/site/targetdetectionwebsite/} is provided. It allows the reader to have a quick access to the main resources, datasets and codes in the field. Finally, with this survey, we aim to bring a one-stop solution, i.e., access to a number of different models, algorithms, implementations and benchmarking techniques in a single paper. \\

Thus, we present the different models in three main categories to strengthen this transition:
\begin{itemize}
\item Traditional Unsupervised Models which present the following characteristics: (1) These models are the first models used in the field and are generally basics in terms of mathematical concepts. (2) These models allow to handle some specific challenges in normal environments, and are generally easy to implement. (3) Numerous improvements concern these models and their limitations seem to be reached.
\item Low-rank and Sparse Decomposition Unsupervised Models which present the following characteristics: (1) These models present a disruptive methodology after the seminal work of Candes et al. \cite{MOD-Theory-1} published in 2011. (2) These models allow to handle more challenges than traditional unsupervised models and are generally easy to implement. (3) Numerous improvements concern these models in terms of modeling of background, accuracy of detection, robustness in presence of noise and computation time.
\item Deep Neural Networks Supervised Models which present on the other hand the following characteristics: (1) These models are more sophisticated to robustly handle a lot of challenges, (2) Most of them need improvements to be deployable and to achieve real-time requirements. To this end, fast and lightweight deep learning algorithms as well as inductive learning algorithms and graph based learning algorithms requiring less labeled data should be developed.
\end{itemize}
Infrared small target detection can be also classified into model-driven methods and data driven methods as done in the work of Li et al. \cite{SFD-TH-12}. The model-driven algorithms require fewer computing resources and do not require training samples as they are unsupervised. Moreover, they are easy to implement on platforms with limited resources and hold great potential for practical implementations The model-driven approaches can be classified into three categories: morphological filtering-drive, human visual system-driven and sparse representation-driven algorithms. The data-driven methods rely heavily on computational resources and labeled training samples as they are supervised. However, according on the spatial and temporal information used, the model-driven methods and data driven methods can be further classified into single-frame and multi-frame detection (also called sequential-frame detection). Single-frame detection methods are enough for SIRST detection while MIRST detection often requires multi-frame detection methods to employ the motion information. Contributions of this paper can be summarized as follows:
\begin{itemize}
\item A first complete overview of infrared small target detection methods concerning more than 1000 papers covering 30 years of research using mathematical tools, machine learning tools and signal processing tools. For each of these three categories, the methods are classified into two groups: single-frame detection and multi-frame detection. Each method is also labeled as static or moving targets detection methods. However, we present for each category in a rationale and historical manner the corresponding first works, the similar research, their variants/improvements and their last developments to provide the reader an overview of where we come from, where we are and where we go. Subsequently, for each category, the methods are classified into single-frame and multi-frame detection, and also labeled as static or moving targets detection methods.
\item A description of the different resources available to allow fair comparisons of the algorithms. More specifically, datasets and performance evaluation protocol using conventional metrics and disruptive metrics are detailed and discussed.
\item An analysis of solved and unsolved challenges and theoretical perspectives to handle the current unsolved challenges. The idea is to provide the reader perspectives how the field will progress in the next years.
\end{itemize}
The rest of this paper is organized as follows: In Section \ref{GeneralOverview}, we first explain the methodology employed for the survey and we give a general overview in terms of papers published per journals/conferences, per year and per category. In Section \ref{Applications}, we investigate the various applications in terms of ground, maritime and air environments. In Section \ref{Challenges}, we review the different challenges met in infrared small target detection, and which steps and issues are involved to deal with them. Then, in Section \ref{ConventionalMethods}, we give an overview of the conventional unsupervised models classified following the mathematical tools. In Section \ref{LRSD}, we review the low-rank and sparse decomposition methods using the matrix representation and the tensor representation achieving single-frame detection and multi-frame detection, respectively. In  the paper titled "Traditional and Recent Approaches for Infrared Small Target Detection: Overview, Challenges and Perspectives (Part 2)", we investigate the recent supervised models based on deep neural networks and the recent promising methods based on graph/hypergraph representation and learning. In Section \ref{RDC}, we present resources, datasets and codes publicly available to test and evaluate algorithms.  Finally, we provide a conclusion and perspectives for future research. 

\section{General Overview}
\label{GeneralOverview}

\subsection{Factual Overview}
The methodology used to collect all the papers since 30 years about infrared small target detection is based on research on IEEE Xplore\footnote{https://ieeexplore.ieee.org/Xplore/home.jsp}, Scopus \footnote{https://www.scopus.com/} and GoogleScholar.The first papers specific to infrared small target detection \cite{SFD-Consistency-1,SFD-TDLMS-1-1,MFD-1} were published in 1994. We also investigate the number of papers per journals and conferences, the number of papers per years and the number of papers per categories.
\begin{itemize}
\item \textbf{Publications per journals and conferences:} We investigate in which journals and conferences papers about infrared small target detection are mainly published. The main stream for journals is practically as follows: IEEE Transactions on Geoscience and Remote Sensing (152 papers), IEEE Geoscience and Remote Sensing Letters (66 papers), IEEE Journal of Selected Topics in Applied Earth Observations (61 papers), ELSEVIER Infrared Physics and Technology (72 papers), ELSEVIER Pattern Recognition (39 papers) and MDPI Remote Sensing (54 papers). The main stream for conferences concerns IEEE IGARSS conferences, IAPR ICPR conferences, and  AAAI Conference on Artificial Intelligence. 87 papers are recent preprint papers available on ArXiv\footnote{https://arxiv.org/} and SSRN \footnote{https://www.ssrn.com/ssrn/} since 2025 and 2026.
\item \textbf{Publications per year:} We investigate the number of publications per year to highlight the attraction of the field over time. The stream for the last decade is as follows: 2017 (23 papers), 2018 (19 papers), 2019 (28 papers), 2020 (39 papers), 2021 (33 papers), 2022 (34 papers), 2023 (110 papers), 2024 (143 papers), 2025 (255 papers), and 2026 (320 papers). 
\item \textbf{Publications per scientific category:} We investigate the number of publications per category to highlight the respective attraction of mathematical tools, machine learning tools and signal processing tools. The stream is as follows: mathematical unsupervised models with 157 papers, low-rank and sparse decomposition models (model-based models) with 159 papers, and deep learning methods (data-based models) with 558 papers, as listed in the overview tables. Thus, data-based methods represent approximately $64\%$ of the reviewed methods.\\
\end{itemize}
In summary, the evolution over the last decade demonstrates the increasing attraction of infrared small target detection. This phenomenon is owing to the progress in cooled and uncooled infrared sensors \cite{IR-Sensors-1,IR-Sensors-20} increasing their deployment of infrared cameras in diverse challenging environments \cite{IR-Sensors-100} and to the rise of deep learning methods which allows to tackle more challenges both in sparse and dense target detection. We can note also that at least $70\%$ publications come from China laboratories confirming their international leadership in the field. In this context, Wei An\footnote{https://dblp.org/pid/74/8455-3.html} (National University of Defense Technology, Changsha, China) emerges as the most active researcher with more than 30 published papers in the field. \\

Note that the current research activities in 2026 generate three up to five publications per week and also generate duplicate acronyms (SANet \cite{DL-32} (2023), SANet \cite{DL-566} (2026) and SANet \cite{DL-599-436} (2026), LESPS \cite{DL-364} (2023) and LESPS \cite{DL-364} (2024), SAMamba (Xu et al. \cite{DL-268} (2025) and SAMamba \cite{DL-320} (2026), IRS-DETR \cite{Applications-S-46} (2025) and IRS-DETR \cite{Applications-S-30} (2026)) providing confusion.

\subsection{Conceptual Overview}
\begin{figure*}
\begin{center}
\includegraphics[width=17cm]{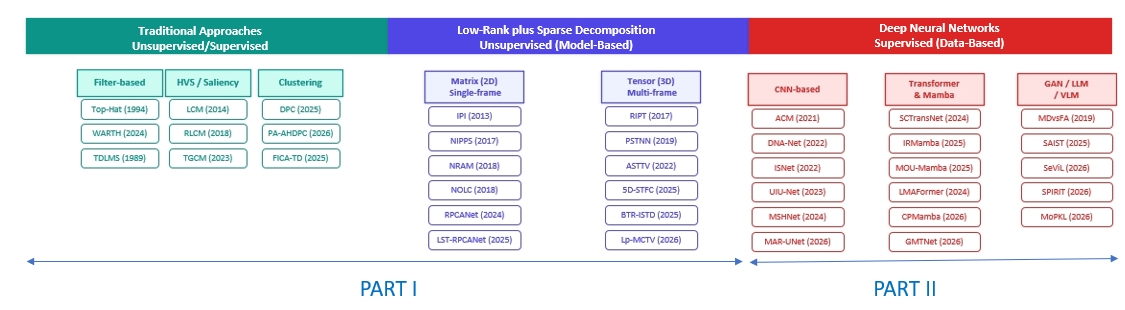}
\caption{Traditional Approaches, Low-rank plus Sparse Decomposition (Model-based) and Deep Neural Networks (Data-based models) : An Overview (with representative models).} 
\label{Overview}
\end{center}
\end{figure*}

Figure \ref{Overview} shows the representative models in the three main categories.  The first two categories namely traditional approaches and low-rank plus sparse decomposition are reviewed in the Part I while deep neural networks are surveyed in Part II. Moreover, we investigate the impact of the conceptual development of mathematical tools, machine learning tools and signal processing on the research about infrared small target detection. Figure \ref{Overview} shows the timeline of the progress in the data modeling tools, in computer vision and infrared small target detection.

\section{Applications of Infrared Small Target Detection}
\label{Applications}
\begin{figure}
\begin{center}
\includegraphics[width=2.5cm]{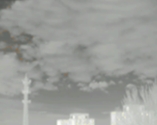}
\includegraphics[width=2.5cm]{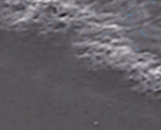}
\includegraphics[width=2.5cm]{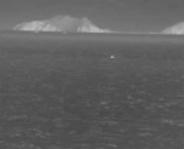} \\
\caption{From left to right: Ground environment, Air environment, Sea environment.} 
\label{Environments}
\end{center}
\end{figure}

Applications of infrared small target detection are done in three main environments: Ground environments (Land-based), air environments (Aerial-based), and sea environments (Maritime-based). Each environment has its own properties and challenges. Figure \ref{Environments} shows a sample image of each environment.

\subsection{Ground Environment}
The targets are vehicles \cite{DL-599-133}, stopped airplanes, etc.  The challenges vary according to urban environments and to natural environments. In urban environments, complex geometric layouts of urban landscapes (buildings, etc.) disturb the detection. In natural environments, dynamic spatial changes owing to the vegetation increase the challenge of detection. Moreover, the visual similarities between background and targets, including similarity in shape and energy, complicate the differentiation process and add layers of complexity to detection algorithms \cite{DL-334}. 

\subsection{Sea Environment}
The targets are ships, boats, vessels, etc. The three main challenges in sea environments are the following ones \cite{Applications-S-21}:
\begin{enumerate}
\item Because of the resolution limitation of remote sensing images and the small size of ships, the targets show weak edges in the images and are difficult to accurately identify.
\item The sea surface environment is changeable with interference information such as ripples, clouds, and wake waves. These interferences easily cause false detection and missed detection.
\item The length-width ratio of ships is very different, and the directions are different.
\end{enumerate}
In literature, several methods \cite{Applications-S-1,Applications-S-2,Applications-S-3,Applications-S-15,Applications-S-16} have addressed small target detection covering the three categories defined in this survey:
\begin{itemize}
\item Traditional unsupervised models with filters \cite{Applications-S-2,Applications-S-10}, consistency of backgrounds \cite{Applications-S-6,Applications-S-7,Applications-S-13}, saliency of targets \cite{Applications-S-5,Applications-S-14,Applications-S-17,Applications-S-34}, SVM \cite{Applications-S-3}, and Isolation Forest \cite{Applications-S-4}.
\item Low-rank and sparse decomposition unsupervised models \cite{Applications-S-11,Applications-S-12,Applications-S-17}.
 \item Deep Neural Networks Supervised Models with detection based methods with YOLO series (EISD-YOLO \cite{DL-705}, EG-YOLO \cite{DL-706}, SeaInfraNet \cite{DL-707}, YCENet \cite{DL-707-1}, GT-YOLO \cite{DL-707-2}, InfraTiny-YOLO \cite{DL-707-3}, DSEE-YOLO \cite{DL-707-4}, DCST-YOL \cite{DL-707-5}, IRMultiFuseNet \cite{Applications-S-47}, InfraShip-Net \cite{Applications-S-63}) and DETR series \cite{MOD-Theory-206} (IRS-DETR \cite{Applications-S-30}, FGOM-RTDETR \cite{Applications-S-33}, FP-RTDETR \cite{DL-711}, KDFF-DET  \cite{DL-712}, MAFF-DETR \cite{DL-717}, SASE-DETR \cite{DL-718}, SG-DETR \cite{DL-719}, Physics-Driven DETR \cite{DL-599-363}), and segmentation based methods (Dedicated Architectures \cite{DL-572,Applications-S-62}, Attention Networks \cite{Applications-S-20,Applications-S-35,Applications-S-36,Applications-S-40,Applications-S-51},  Flexible Networks \cite{Applications-S-21}, Transformers \cite{Applications-S-33,Applications-S-30,DL-340,Applications-S-41,Applications-S-64,DL-575,DL-599-361}, Feature Guidance \cite{Applications-S-31,DL-573,DL-599-46}, Semantic Guidance \cite{Applications-S-42,DL-574,Applications-S-65}), and spatial–temporal graph neural networks \cite{Applications-S-19,Applications-S-32}.
\end{itemize}
Each method is reviewed in each corresponding section. Note that infrared and visible image fusion is a key technique for all-weather ship surveillance in maritime security applications  \cite{MOD-Survey-10,MOD-Survey-11,Applications-S-100,Applications-S-101,Applications-S-102}. Indeed, this technique allows to compensate the deficiencies of single-modal imaging under low-illumination and complex sea-state conditions \cite{Applications-S-100}. 

\subsection{Air Environment}
Airborne IRST and FLIR based systems acquire infrared image by imaging systems that sense infrared radiation \cite{Applications-A-3}. The targets are aircrafts, missiles, etc. for military applications and airplanes, birds, etc. for civil applications. In numerous situations, these targets need to be detected at considerable distances, which can extend to several hundred kilometers. Moreover, airplanes can be submerged in a thick cloud and birds may fly in the jungle. Working in three main infrared atmospheric ranges (1--2.7~\si{\micro\meter}, 3--5~\si{\micro\meter} and 8--14~\si{\micro\meter}), the thermal imaging cameras have two basic infrared wavelength ranges: medium-wave infrared (3--5~\si{\micro\meter}) and long-wave infrared (8--12~\si{\micro\meter}). These characteristics allow to detect aircrafts with medium-wave infrared radiation characteristic of smoke plume and long-wave infrared radiation characteristic of aircraft skin when it is flying with high-speed. But, the images in the form of visible light include both interested target and background and noise. More specifically, the infrared radiation characteristic of sky background is related to time, space and climate. The radiation comes from two main parts: diffuse solar radiation ($<3$~\si{\micro\meter}) and atmospheric thermal radiation ($>4$~\si{\micro\meter}) affecting the infrared image. In literature, several authors \cite{SFD-LCM-17,Applications-A-1,Applications-A-2,Applications-A-4,Applications-A-5,SFD-RPCAMatrix-6} have addressed small target detection in air environments.  More specifically, Li et al. \cite{SFD-LCM-17} design a saliency based method. Wang et al. \cite{Applications-A-1} and  Gao et al. \cite{SFD-RPCAMatrix-6} employ low-rank plus sparse matrix decomposition methods whilst Sasikaladevi et al. \cite{Applications-A-2} design in a deep neural networks method.

\section{Preliminaries, Challenges and Issues}
\label{Challenges}
\subsection{Preliminaries}
In the field of infrared small target detection, the Society of Photo-Optical Instrumentation Engineers defines a small target as an entity with a spatial extent typically smaller than $9 \times 9$ pixels. In 2003, Zhang et al. \cite{Theory-RS-1} consider a target as a small target if it occupies less than $0.12$ percent of the total pixels in an image. In the case of $256 \times 256$ images, targets smaller than $80$ pixels are categorized as small. These targets usually exhibit a prominent central intensity that attenuates symmetrically in all directions, providing as weak isotropic responses with negligible shape or textural structure. On the other hand, background clutter often displays pronounced directional characteristics, especially in edge-rich regions or parallax-dominated areas. The local structure refers to specific features of the target and its surroundings: 
\begin{itemize}
\item Target Morphology: Small targets (e.g., aircraft, vehicles, or people) appear as bright spots with distinct shapes and sizes.
\item Edge Information: The boundary between the target and the background is clear and has high contrast, making it a key local feature.
\item Texture Features: Even in low-resolution images, texture differences provide useful information. For example, the sky and flat ground have uniform textures, while vegetation and buildings show more complex texture
\end{itemize}
In summary, the infrared small targets present the following characteristics \cite{DL-12}:
\begin{itemize}
\item Small: Because of the long imaging distance, targets are generally small, ranging from one pixel to tens of pixels in the images.
\item Dim: Infrared targets usually present low Signal-to-Clutter Ratio (SCR) and are often immersed in heavy noise and clutter background.
\item Shapeless: Infrared small targets have limited shape characteristics. 
\item Changeable: The sizes and shapes of infrared targets vary a lot among different scenarios.
\end{itemize}
Sometimes researchers refer to infrared dim small target detection. The term “dim” mean the low intensity or weak contrast of the target signal. Even if there is no unified quantitative consensus for “dimness” as researchers continually push the limits of algorithm sensitivity, it generally mean targets that are submerged in background clutter with low Signal-to-Clutter Ratios (SCR), making them visually indiscernible. In literature, a target is considered as dim when its local SCR is below 3, a threshold where effective detection becomes extremely challenging for algorithms.

\subsection{Challenges}
The different challenges met in infrared small target detection can be classified as follows.

\subsubsection{Sensor Challenges}
\begin{figure}
\begin{center}
\includegraphics[width=7cm]{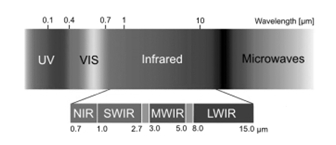} \\
\caption{Near-Infrared (NIR), Short-Wave Infrared (SWIR), Mid-Wave Infrared (MWIR), and Long-Wave Infrared (LWIR).} 
\label{WaveSpectrum}
\end{center}
\end{figure}

The challenges related to the sensors can be categorized as follows:
\begin{itemize}
\item \textbf{Sensor Locations:} IR sensors are either fixed or moving sensors. In the last case, they can be mounted on drones, on UAVs (Unmanned Aerial Vehicles) \cite{DL-710,DL-715,DL-716,DL-599-450} or on satellites (spaceborne \cite{DL-712,DL-599-46}).
\item \textbf{Wavelengths Spectrum:} In practice, the wavelengths captured by sensors are classified into four main categories: Near-Infrared (NIR), Short-Wave Infrared (SWIR), Mid-Wave Infrared (MWIR), and Long-Wave Infrared (LWIR). Figure \ref{WaveSpectrum} shows an illustration of these four categories with their range.
\item \textbf{Sensors Sensibility:} Infrared images are typically affected by low signal-to noise ratios. Indeed, infrared sensors should identify targets at long distances (usually up to hundreds of kilometers), causing them to appear small and difficult to distinguish from the background. Second, strong vignetting can be observed in various scenarios \cite{SFD-RPCAMatrix-48}. Vignetting in infrared imagery is characterized by a pixel intensity from the  image center toward the edges.The brightness distribution of vignetting is inconsistent, with the center region being nearly saturated and significantly brighter than the other regions. Third, noise fluctuations and some nonuniform streaks generated during the photoelectric conversion process have the common characteristic of weak intensity with the target. These problems generate significant challenges for small target detection against a strong vignetting background \cite{SFD-RPCAMatrix-48}. 
\end{itemize}

\subsubsection{Target Challenges} 
Due to their extremely small size and lack of distinct texture, such targets often provide insufficient structural features for reliable identification. In addition, disparities in shape and texture characteristics between infrared and visible light imagery led to suboptimal performance. In literature, targets are classified into three main categories: extended targets, spotted targets and point targets.

\subsubsection{Background Challenges}
Strong background noise complicates target extraction. Natural interference such as cloud boundaries, sea surface clutter and man-made interference further increases false positive rates and reduces detection performance. The main scenarios can be classified into the following types: air-based, ground-based, and air–ground interfaces \cite{DL-334}.
\begin{itemize}
\item Cloud clutter: Air exploration often takes place from the
ground up or from the air to a higher altitude and scene elements are often relatively simple, being mainly composed
of sky clouds. However, cloud clutter generate challenges to detect air targets because clouds are naturally formed with blurry edge contours and
have no fixed shapes.
\item Blurred water: Water surfaces present strong reflection
and scattering abilities because they are affected by the optical properties of water molecules. In addition, water zones are complex with various shadows and light spots. Finally, lines and waves on water surfaces may distort the outlines of a scene and interfere with the detection, recognition, and positioning of targets.
\item Vegetation cover: Vegetation generate complex scenes. Ground
vegetation areas consist of the same types of plants typically exhibit a high degree of similarity. They contain fine texture features, which can lead to differences in texture features in the presence of different plant species.
\item  Urban ground: Urban ground scenes present man made objects with various shapes and densities. For example, buildings present significant geometric features and the diversity of scene elements leads to large differences in the imaging responses of ground objects, which interferes with target detection.
\end{itemize}
Various backgrounds collectively present these characteristics that increase detection difficulty. The main challenges are the irregularity and dynamic aspects of the backgrounds, and the potential similarity in visual features between targets and their backgrounds (camouflage) \cite{DL-334}.

\subsection{Issues}
The infrared data are thus affected by low signal-to noise ratios. Furthermore, data corresponding to the small target lack of distinct texture. Modeling tools that can be employed are mathematical tools, machine learning tools and signal processing tools. For mathematical tools, statistic, fuzzy, and Dempster-Shäfer methods take into account imprecision, uncertainty, and incompleteness in the data, which appear due to different challenges, and because of the quality of the sensor. Regarding robust representation learning methods (i.e models-based methods), unsupervised techniques seek for low-rank structure in the data and attempt to be robust to outliers. Deep Learning models (i.e data-based methods) seek to be able to face all the challenges by learning all the situations. Signal processing models make an estimation of the background based on the temporal distribution of the pixels in image frames. In literature, these various tools have been employed: traditional and low-rank and sparse decomposition models are reviewed in this paper, and deep learning models in the paper titled "Traditional and Recent Approaches for Infrared Small Target Detection: Overview, Challenges and Perspectives (Part 2)".

\section{Conventional Unsupervised Methods}
\label{ConventionalMethods}

\subsection{Preliminaries}
Conventional unsupervised methods employed mathematics tools based on crisp, statistical and fuzzy concepts and signal processing tools using filters. Note that statistical and fuzzy concepts allows to take into account the imprecision, the uncertainty and the incompleteness that may occur in the data.

\subsection{Single-Frame Detection}
Single-frame infrared small target detection methods can be classified  into two main categories based on different assumptions about the relationship between the target and the background. More specifically, the categories can be identified as follows:
\begin{enumerate}
\item The first category assumes that targets disrupt local background correlations, where backgrounds exhibit slow variations and strong interpixel correlations. The methods are based on prior information and they are based either on local prior filtering and nonlocal prior optimization methods as follows
\begin{itemize}
 \item \textbf{Local Prior based Methods:} Their core idea is to focus on the pixel neighborhood (usually  the range of $3 \times 3$ up to $15 \times 15$), and to recognize the target by analyzing local contrast, gradient changes, and other features. Its theoretical basis is that weak and small targets will form significant extreme points or abnormal points in local areas. This type of algorithm mainly includes background estimation
techniques (i.e. filters techniques), directional gradient and local contrast. First, specialized filters have been proposed to adapt to different backgrounds and targets. These filtering-based methods operate in the intensity or derivative domain suppressing clutter while enhancing targets. More specifically, these filtering methods employ prior information on background consistency to estimate the background. However, these methods often amplify non-target edges providing to false alarms in complex scenes. To address this limitation, the second kind of methods enhance targets by exploiting local contrast differences. Local contrast measurement and multiscale contrast measurement  are  based on the same methodology by enhancing the saliency of small targets. The directional gradient technology distinguishes the target from the background by using the local gradient feature of the image, which can effectively capture the difference between the target and the background in the edge direction In addition, different variants of enhancement insights of small target have been proposed and achieved advanced performance. In summary, these methods increase the separability between targets and background by refining contrast measurement. 
\item \textbf{Non-Local Prior based Methods:} The non-local information methods are based on the difference between the target image and the non-target in the  frequency spectrum data to achieve the extraction of the target image. It can use the dependence on local characteristics and highlight the global nature. It addresses the challenge that the background characteristics are too large, resulting in the difficulty of target detection. The core idea of this kind of algorithm is thus to establish a global feature model and to suppress the background by mining the long range dependence relationship of the spatial-frequency domain of the image. It breaks through the limitations of local windows and uses the sparsity and continuity characteristics of the target in the entire image. 
\end{itemize}
These methods based on local and non-local priors are surveyed in Section \ref{Consistency}, Section \ref{Saliency}, Section \ref{Classification}, and Section \ref{Discriminative}. In addition, Table \ref{OverviewSFD-1} and Table \ref{OverviewSFD-2} show an overview of these different single-frame detection methods.\\
\item Unlike the first category based on strong local pixel correlations, the second category assumes similarities even among nonadjacent pixels and employs low-rank decomposition models for target detection. These methods reformulate detection as an iterative optimization problem using sparse and low-rank representations. These methods have been the SOTA algorithms between 2012 and the rise of deep learning methods and present the advantages to be model-based requiring no labels data. These methods are reviewed in Section \ref{LRSD-SFD}.
\end{enumerate}

\subsubsection{Classification/Clustering Methods}
\label{Classification}

\begin{itemize}
\item \textbf{Classification Methods:} In 2017, Bi et al. \cite{SFD-Classification-1} propose a multiple novel features approach. Each feature reflects specific characteristic of small target. The best feature vector is selected to apply these features for detection. Then, a learning-based classifier is trained to scan candidate targets which are obtained by initial segmentation. Experiments show that the proposed features method discriminates effectively small targets from various clutters. In an other work, Liu et al. \cite{SFD-Classification-4} design an Hierarchical Structure Dependency Whitening (HSDW). First, Liu et al. \cite{SFD-Classification-4} characterize domain discrepancies in infrared images as differences in structural information. Then, Liu et al. \cite{SFD-Classification-4} use feature whitening to mitigate the dependency on domain-specific structure information, whose distribution is diversely simulated by a dual-branch nonlinear transformation module. Liu et al. \cite{SFD-Classification-4} adopt a hierarchical suppression mechanism to alleviate the structural biases across multiple decoding stages, thereby facilitating more generalized target understanding across domains. \\
\item \textbf{Density Peak Clustering:} In 2025, Wu et al. \cite{SFD-Classification-3} propose a method based on improved Density Peak Clustering (DPC) and gray-level contribution. First, Wu et al. \cite{SFD-Classification-3} apply an attribute filtering to quickly extract candidate targets. This attribute cannot only guide the parameter setting of the improved DPC (IDPC) but also derive the weights of feature fusion. After,  Wu et al. \cite{SFD-Classification-3} build an unsupervised clustering model based on IDPC, which accurately represent the local features of these targets. A gray-level model allows to extract the global features of small targets, leveraging the statistical properties of the gray level of small targets. By the weighted fusion of local and global features, the clustered targets are effectively enhanced, while the background clutter is further suppressed. Experiments show that IDPC provides a superior clustered target enhancement effect and a higher probability of multitarget detection compared to previous SOTA algorithms. 
\item \textbf{Monte Carlo linear Clustering:} In 2023, Li et al. \cite{PSTD-37} propose an approach using Monte Carlo linear clustering with single-point supervision  to avoid a large amount of training data with expensive pixel-level annotations. This method aims to recover the per-pixel mask of each target from a single point label using Monte Carlo linear clustering. This method employs randomness to the clustering process by adding noise to the input images. By averaging the clustered results, more reliable pseudo masks are obtained. This Monte Carlo clustering approach allows the method to accurately recover pseudo masks and convert fully supervised SIRST detection networks into weakly supervised ones with only single-point annotation. 
\item \textbf{Riemannian manifold Clustering:} Infrared small target detection present the challenge of center drift in the case of  unsupervised clustering which is misled by heavy-tailed background clutter in flat Euclidean spaces. To address this limitation, Zhu et al. \cite{SFD-Classification-10} design a Spatio-Temporal prior-guided Semi-supervised Riemannian manifold Detector and Tracker (ST-SRDT\footnote{https://github.com/llx-xupt/ST-SRDT}). First, spatial coordinates and high-order optical structural features are separated and mapped onto a symmetric positive definite (SPD) covariance manifold. Second, the curved matrices are projected onto a flat Lie algebra tangent space using the Log-Euclidean metric (LEM). Third, a semi-supervised Riemannian fuzzy C-means (SSFCM) framework robustly resists hierarchical clutter and steers the trajectory toward the true target. Experiments show that ST-SRDT outperforms WSLCM \cite{SFD-LCM-18} and PSTNN \cite{SFD-RPCATensor-5} while running in real time at 24 frames per second.
\item \textbf{Hyperbolic Density Clustering:} All the detection algorithms typically map hierarchical background clutter into a flat Euclidean space and results in a “feature crowding effect” because of the geometric mismatch between spatial capacity and data structure. In 2026, Zhu et al. \cite{Miscellaneous-200}  propose a hyperbolic density clustering method based on a training-free detection paradigm which integrates non-Euclidean geometric reconstruction with optical mechanism verification. In a further work, Zhu et al. \cite{Miscellaneous-201} design a Physics-Aware Adaptive Hyperbolic Density Peak Clustering PA-AHDPC\footnote{\url{https://github.com/llx-xupt/PA-AHDPC}}). First, PA-AHDPC employs dual-prior probabilistic screening to rapidly anchor high-confidence candidate regions. Second, PA-AHDPC exploits the exponential expansion property of the hyperbolic Poincaré ball relative to its radius
to address the local feature distribution. PA-AHDPC isolates weak targets from strongly structured clutter at the topological level on a negatively curved manifold. Third, a strict Local Gradient Difference (Strict-LGD) soft-decision mechanism incorporating energy spillover tolerance filters out pseudo-targets lacking continuous energy diffusion characteristics, such as high-brightness sensor dead pixels. Experiments show that PA-AHDPC outperforms the second-best SOTA algorithms by achieving an approximate 5 to 10 dB improvement in signal-to-clutter ratio gain (SCRG). To address the feature crowding effect where heavy-tailed background clutter overlaps with weak targets within flat Euclidean spaces, Zhu et al. \cite{Miscellaneous-202} propose a Hyperbolic Wasserstein Manifold Detector (HWM-Det\footnote{https://github.com/llx-xupt/hyperbolic-Wasserstein-manifold-detector}). First, a spatial
topological offset physically isolates target radiation. A cascade screening reduces
computational complexity to linear time. Second, local patches guided by Hessian structural tensors are mapped onto a Symmetric Positive Definite (SPD) manifold to keep the target's optical point spread function (PSF) prior. Third, the high-dimensional features are embedded into a Poincaré ball. An extreme value theory (EVT)-driven Bayesian framework generates a maximum a posteriori (MAP) recovery bound. Experiments show that HWM-Det outperforms RLCM, TLLCM and PSTNN \cite{SFD-RPCATensor-5}.
\end{itemize}

In summary, clustering methods present several interesting abilities for infrared small target detection especially with recent approaches based on density Peak clustering and hyperbolic density clustering published in 2025 and 2026, respectively.

\subsubsection{Discriminative Methods}
\label{Discriminative}
In 2025, Luo et al. \cite{SFD-Discriminative-1} propose a multi-frame small target detection method based on frequency data sphering (FDS) and independent component analysis (FICA-TD) in order to capture the characteristics of different components. First,  a  spatial-temporal tensor with its adjacent frames is constructed from the videos. Second, the low-frequency information is suppressed with a frequency high-pass filtering operator and a FDS operator. Third, an Independent Component Analysis (ICA) method allows to determine the number of signal sources and the basis of independent component space followed by projecting the frequency data-sphered tensor into the independent component space. A Sparsity Information Filter (SIF) is then employed to filter out the residual low-response information of the middle frontal slice. Finally, the target detection result sequence are obtained. Experiments conducted on five real IR sequences show that FICA-TD is superior to seven state-of-the-art competitive methods in terms of target detectability, background suppressibility, and overall performance. 

\subsection{Consistency of Backgrounds}
\label{Consistency}
Methods based on the background spatial consistency assume that the unknown background can be predicted based on the correlation between pixels while the occurrence of the target breaks the correlation and makes the prediction fail. In this way, the target and background separation is achieved by subtracting the predicted image from the original one. 
These filter based methods can be classified as follows: Wavelet/Contourlet filters \cite{SFD-Consistency-1,SFD-Consistency-2,SFD-Consistency-3,SFD-Consistency-4} ,Top-Hat filters \cite{SFD-TH-3,SFD-TH-4,SFD-TH-5,SFD-TH-6,SFD-TH-10,SFD-TH-12,SFD-TH-14}, Least Mean Square filters \cite{SFD-TDLMS-1,SFD-TDLMS-2,SFD-TDLMS-3,SFD-TDLMS-4,SFD-TDLMS-5} and Max-Min filters \cite{SFD-MaxM-1}.

\subsubsection{Wavelet/Contourlet based Methods}
In 1994, Gregoris et al. \cite{SFD-Consistency-1} propose a multi-scale wavelet transform algorithm in forward looking infrared (FLIR). In 2012, Zhao et al. \cite{SFD-Consistency-3}  design a detection algorithm based on wavelet transform and image fusion. The original image is decomposed and reconstructed by wavelet and it is separated into a low frequency image and some high frequency images. Then, the low frequency image is removed and the high frequency images are fused into a new image. The fused image is segmented with a threshold to obtain the target. In an other work, Zhao et al. \cite{SFD-Consistency-4} develop a detection method based on the nonsubsampled contourlet transform. The nonsubsampled contourlet transform allows to decompose the source image. Considering the different energy distribution characteristics of the target, background and noise, the energy image is obtained by energy cross correlation operation. Then, a few of candidate targets were segmented through adaptive threshold. Finally, the target was confirmed by using the continuity and consistency of target energy intensity and target movement in image sequence.  

\subsubsection{Top-Hat Filters}
Morphological detection is the first method employed infrared small target detection. Along with the development of morphological research, morphological image processing  has gradually become a new trend in image processing field and a suitable tool in small target detection. The final performance of the morphological filter strongly depends on the selection of the structuring element including the shape and value of it once given the morphological operation regulation. There will be different geometric structure information to analyze and process in the operation according to different structuring element being selected. In this context, most of the researchers have adopted the Top-Hat operations for infrared small target detection. More specifically, the morphological Top-Hat operation \cite{SFD-TH-3-1} is a high-pass filtering operation which includes opening Top-Hat operation and closing one. The opening Top-Hat operation can detect wave crest of an image and the closing Top-Hat operation can detect wave hollow. In 2002, Bin and Peng \cite{SFD-TH-3-2}  indicate that the Top-Hat operation is similar to a non-spread operation and the opening Top-Hat operation is always positive. Therefore, Top-Hat filters provide good results only in some corresponding image model. However, usually the image signals are extremely complicated and change all the time, so the selected structuring element should be adaptive for realizing optimization. But, classical top-hat transformation employs two same structuring elements in the operations of opening or closing, and does not use this difference information in the structuring element appropriately. This fact decreases the efficiency of classical top-hat transformation to differentiate the real target region and noise or clutter regions. Moreover, the noise or clutter may affect the performance of classical top-hat transformation as follows: 1) The noises may lead to target losing or increase the number of false alarms, and 2) the classical top-hat transformation cannot differentiate the heavy clutter and real target region. In literature, two methods for optimizing structuring elements have emerged \cite{SFD-TH-3}: morphological neural networks and morphological genetic algorithm. In 2006, Zeng et al. \cite{SFD-TH-3} propose two methods to optimize Top-Hat morphological filtering parameters. The first is based on neural network. Its structuring element is a two-layer feed-forward network which is trained by a mass of sample nets. This method considers Top-Hat operation as a whole and one layer, and defines the node of the output layer as the maximum gray-scale image vector after Top-Hat operation. The second method is based on genetic algorithm. This method adopts the interval discretization code and  primary and secondary mood crossover and mutation. In 2010, Bai and  Zhou \cite{SFD-TH-5} propose the definition, properties, multi-scale operations of top-hat transformation. The definition of this top-hat transformation employs two different but correlated structuring elements to reorganize the classical top-hat transformation, and takes into account of the difference information between the target and surrounding regions. Experiments show that this top-hat transformation can detect infrared  small target more efficiently than the previous top-hat transformations. In  further work, Bai and  Zhou \cite{SFD-TH-6} propose an other modified top-hat transformation improving the contrast between bright and dark regions and using an iterative thresholding method to automatically detect the target region. Nevertheless, these methods have problem in suppressing the fixed position sharp noise and poor real-time performance. To solve this problem, Wang et al. \cite{SFD-TH-7} design a three-dimensional dynamic queue pipeline and employs  a space-domain algorithm based on morphology Top-Hat operator in order to ensure the continuity of the target and the pipeline. However, the structure elements are desired to be adjusted adaptively. To this end, Deng et al. \cite{SFD-TH-4} propose an adaptive structure elements optimization method based on Quantum Genetic Algorithm (QGA). Experiments show that QGA achieve more stable performance compared with the traditional top-hat filter. \\

Despite its simplicity and the rise of LRSD models and deep neural networks, several researchers still develop improvements. In 2022, Yang et al. \cite{SFD-TH-13} propose a multi-structural element top-hat operator to obtain the suspected target area through adaptive threshold segmentation. This method employs image feature algorithms, such as central pixel contrast, regional gradient, and directional gradient, to extract the feature information of the suspected target at multiple scales, and the fuzzy decision method is employed for multi-feature fusion to achieve the final target detection. In 2023, Yavari et al. \cite{SFD-TH-11} propose a multi-stage morphological operator. The algorithm consists . of three stages. First, the noise is eliminated by means of Gaussian filtering to improve the SNR. Second, the morphological-based operation is applied but unlike the NTHF which uses the largest value of the neighborhood of a point, the average of $n$-largest values are employed. This strategy addresses the shortcoming of the NTHF when clutter is close to target. The residual noise and clutter is then reduced in the final stage using another morphological operation. In 2024,  Li et al. \cite{SFD-TH-12}  propose a Weighted Adaptive Ring Top-Hat transformation (WARTH). First, an adaptive ring-shaped structural element  and a target awareness indicator are employed to measure local and global feature information to detect small targets while minimizing false alarms. Second, a Difference-of-Structure Tensors (DoST) is proposed and the positive smallest eigenvalue of DoST (PSEDoST) is computed to obtain the adaptive ring-shaped that captures local feature information. Third, the image is converted into the Fourier phase spectrum, and a two-stage sliding window filtering technique is applied to obtain the target awareness indicator that perceives global feature information. At the last stage,  WARTH is defined by fusing the above two measurements which eliminate false alarms and improve the robustness of target detection.

\subsubsection{Least Mean Square Filters}
In 1988, Hadhoud and Thomas \cite{SFD-TDLMS-1} propose a general-purpose two-dimensional adaptive filter, which extends the one-dimensional Least Mean Square (LMS) filter and was later adopted for infrared small target detection. This method attempts to minimize the mean square error between the predicted response and the output signal.  Thus, the TDLMS filter predicts the pixel intensity based on the weighted intensity of nearby pixels. The prediction error is large near the target and is close to zero in the background. This is because the intensity of the pixels near the target changes rapidly and is hard to predict accurately. In the opposite, the intensity of the background pixels is homogeneous and easy to predict.  Thus, the TDLMS filter enhances small targets and suppresses backgrounds. In 1994, Azimi et al.  \cite{SFD-TDLMS-1-1} propose the two-dimensional BDLMS (Block Diagonal Least Mean Square) method. The weight matrix is updated once per block rather than once per pixel. The diagonal scanning is choosen to avoid the problems inherent in the one-dimensional (1D) standard scanning schemes. In 2009,  Bae et al. \cite{SFD-TDLMS-1-2} develop a mask-weighted TDLMS method that places the predicted pixel in the center of the masked prediction matrix. The method can obtain more neighborhood information than the conventional TDLMS. Nevertheless, the methods present problems related to the step-size adjustment of TDLMS  \cite{SFD-TDLMS-1-3,SFD-TDLMS-1-4,SFD-TDLMS-1-5}. First, Cao et al. \cite{SFD-TDLMS-3} employ five prediction matrices to obtain the neighborhood information in order to compute the step size. The step size is a function of the variance of the four prediction values in the four directions. The step size is long near the changeable area such as the cloud edges and short in the homogeneous backgrounds. In the same idea, Bae et al. \cite{SFD-TDLMS-4} employ a eight prediction matrices to obtain the neighborhood information improving the TDLMS algorithms by fully using the neighborhood information. This allow to determine whether the pixel is near the edge of the cloud or near the target. Thus, the step size near the edge of the cloud is increased and the step size near the target is reduced according to the neighborhood information.  In a further work, Bae et al. \cite{SFD-TDLMS-5} employ an edge directional 2D LMS filter to take advantage of edge information of prediction pixels corresponding to surrounding blocks around current filter window. To adjust adaptively the step size in the background and small target region, the adaptive region-dependent nonlinear step size is obtained with the variance of the prediction pixels of the surrounding blocks. This prediction structure and adaptive step size is then applied to the background region including objects such as cloud edge and small target region differently. Nevertheless, these above TDLMS filters can not detect small targets accurately. Moreover, their algorithms are time-consuming and do not perform well in real-time applications. To address these disadvantages, Zhao et al. \cite{SFD-TDLMS-2} develop a bilateral filtering method. This method filters the images rightwards and leftwards and then fuses the prediction error images. Thus, the cloud edges and noise spots are eliminated sharply and the targets are significantly enhanced using the addition of the leftward filtering and fusing processes. Zhao et al. \cite{SFD-TDLMS-2} also employ a selected, fixed step size and simplified filter structure to improve the computing speed. Experiments show that this algorithm provide better detection performance as well as competitive execution efficiency than the previous TDLMS algorithms. \\

In summary, these methods are based on morphological filtering to detect targets by suppressing the background. These methods show good detection performance in the case of simple and uniform backgrounds. 

\subsection{Saliency of Targets}
\label{Saliency}
Human Visual System (HVS) methods assume that the target is the most salient object. These HVS based methods can be classified as follows: Local contrast methods \cite{SFD-LCM-1,SFD-LCM-2,SFD-LCM-3,SFD-LCM-4,SFD-LCM-5,SFD-LCM-6} and global contrast methods \cite{SFD-GCM-1,SFD-GCM-2,SFD-GCM-3,SFD-GCM-4,SFD-GCM-5}. Their progress over time can be summarized as follows.

\subsubsection{Local Contrast Methods}
Conventional HVS algorithms commonly employ the ratio of gray values between the target and the surrounding background to define the local contrast, as a means to obtain enhancement of the target and suppression of the background. In this idea, Chen et al. \cite{SFD-LCM-4} propose in 2014 the Local Contrast Measure (LCM). In a further work, Han et al. \cite{SFD-LCM-5} propose an Improved LCM (ILCM) detection algorithm based on HVS to achieve good performance in detection rate, and to reduce false alarm rate and speed simultaneously. First, an HVS size-adaptation process is employed, and the IR image after pre-processing is divided into sub-blocks to improve detection speed. Based on HVS contrast mechanism, the improved local contrast measure is employed to obtain the saliency map. Finally, a threshold operation along with a traversal mechanism based on HVS attention shift mechanism is employed to get the target sub-blocks. This algorithm based on HVS (i.e difference of Gaussians (DoG) filters) is sensitive to not only real small targets but also background edges, which results in a high false alarm rate. To address this problem, Han et al. \cite{SFD-LCM-11} propose an improved DOG (IDoGb) filter. IDoGb is an extension of DoG but is sensitive to orientations and can better suppress the complex background edges. IDoGb achieves a lower false alarm rate and multi-scale detection can be also achieved. Experiments show that the IDoGb filter produces fewer false alarms than ILCM at the same detection rate, while consuming only about $0.1$ s for a single frame. However, these methods use the absolute local contrast, which may not be effective in eliminating noise and high-brightness clutter. To address this issue, Han et al. \cite{SFD-LCM-12} propose a Relative Local Contrast Measure (RLCM) to handle small targets of different sizes offering better effectiveness and robustness in complex background. In an other work, Wei et al. \cite{SFD-LCM-13} incorporate the biological vision mechanism and design an algorithm called Multiscale Patch-based Contrast Measure (MPCM). The algorithm allows the simultaneous detection of bright and dark targets in infrared images. However, it exhibits relatively poor robustness in complex backgrounds with strong noise. In 2016, Deng et al. \cite{SFD-LCM-14} design a Weighted Local Difference Measure (WLDM)-based scheme for the detection of small targets in various complex cloudy-sky backgrounds. First, a WLDM map allows to simultaneously enhance targets and suppress background clutters and noise. Thus, the true targets can be easily separated from jamming objects. Second, a simple adaptive threshold is used to segment the targets. In 2018, Gao et al. \cite{SFD-LCM-15} employ multiscale gray and variance difference measures. A multiscale adaptive gray difference measure is first computed to enhance small targets and improve detection accuracy. Second, a multiscale variance difference measure allows to alleviate the impact of background fluctuation. By integrating these two approaches, targets can be extracted more accurately using a threshold-adaptive segmentation. In 2019, Qin et al.  \cite{SFD-LCM-15} propose a method based on Facet Kernel and Random Walker. First, local order-statistic and mean filtering are applied to remove the pixel-sized noises with high brightness and smooth the infrared images. The infrared image is filtered by the Facet kernel to enhance the target pixels. Candidate target pixels are then extracted by an adaptive threshold. Then, Qin et al. \cite{SFD-LCM-16} define a Novel Local Contrast Descriptor (NLCD) to achieve clutter suppression and target enhancement. The candidate target pixels are selected as central pixels o obtain the local regions. The NLCD map of all local regions is weighted by the filtered map of Facet kernel to further enhance target. Finally, the target is detected by a thresholding operation on the weighted map. 

All these local contrast based methods adopt a sliding window with a two-layer structure, which restricts the computation of contrast features to a single scale at a time. However, several algorithms use multiscale windows to calculate the local contrast at various scales. The maximum value obtained from these calculations is selected as the final local contrast. For example, to achieve the best detection effect for targets of different sizes, the LCM algorithm uses four scales of sliding windows to traverse the image, and the sizes of the sub- windows are set to $3 \times 3$, $5 \times 5$, $7 \times 7$ and $9 \times 9$, respectively. However, this increases the computational complexity and processing time. Furthermore, recent works published in 2026 are based on ring local contrast measure either with   spatial–temporal prior  \cite{SFD-LCM-91}, with edge suppression \cite{SFD-LCM-92}, with structure-aware weighted information  \cite{SFD-LCM-113}, or feature-guided discriminative mechanisms \cite{SFD-LCM-116}.

\subsubsection{Global Contrast Methods}  
A local contrast method can work well only when the target is locally prominent, which is not always the case when the background is very complex. In 2023, Tang et al. \cite{SFD-GCM-1} design a fast infrared small target detection algorithm based on global contrast with a low computational time consumption. First, the original image is processed by dilate operation with a ring structuring element. Second the detection result is obtained with a Global Contrast Measure (GCM) between the original image and the processed image according to the ratio form and the difference form. But this method presents
limitations in background suppression and multi-target detection capabilities. To address these limitations, Tang et al. \cite{SFD-GCM-2} develop a Multiscale Global Contrast Measure (MSGCM). Experiments show that MSGCM provides better performance than GCM with an average increase of $2.95$ times in background suppression factor, $6.17$ times in signal-to-clutter ratio gain for single-target detection, and $10.49$ times for multi-target detection. To improve the multi-target detection capabilities, Tang et al. \cite{SFD-GCM-3} employ a method based on Facet Global Contrast Measure (FGCM). In an other work, Xie et al. \cite{SFD-GCM-4} propose an Improved Global Contrast Measure (IGCM). The original image is processed by dilate operation with a ring structuring element and mean operator with a square structuring element, respectively. Second, the improved global contrast measure is obtained by the two results above. IGCM provides an higher signal-to-clutter ratio and background suppression factor than the previous global contrast methods. In 2025, Han et al.\cite{SFD-GCM-5} develop a True Global Contrast Measure (TGCM) in which its baseline is obtained by global decomposition. This method can handle situations where the target is not locally salient. A Gaussian filtering operation is employed to suppress noises, and a local max dilation is used on the separated baseline to maintain as much edge/corner information as possible. The residual information during decomposition are employed as a weighting function to further suppress clutters. In a further work, Han et al. \cite{SFD-GCM-6} design a contrast method framework named Hybrid Contrast Measure (HCM). This framework consists of two main modules: a Relative Global Contrast Measure (RGCM), and a small patch local contrast weighting function. First, the sparse and low rank decomposition method is adopted to get the global background of a raw image as benchmark, and a Local Max Dilation (LMD) operation is applied on the global background to recover edge/corner information. Then, a Gaussian matched filtering operation is applied on the raw image to suppress noise. The RGCM is computed between the filtered image and the benchmark to enhance true small dim target and suppress flat background area simultaneously. Second, the Difference of Gaussians (DoG) filtering is employed and improved as the weighting function. Experiments show that RGCM presents robustness in presence of  interference of high brightness background, complex edges/corners and random noises.

\subsubsection{Transform Domain Methods}
\label{TransformDomain}
Transform domain methods often use a special transform such as Wavelet transform, Fourier transform, phase spectrum of quaternion Fourier transform (PQFT) and so on. Detection in complex backgrounds performs better in transform domain than in the spatial domain in spite of higher computation complexity \cite{MFD-11}. These transform domain methods define a frequency difference between the background and the target in the transform domain, and obtain the detection by magnifying the measured difference. Because more details in an image imply higher phase information values in the transform domain, better detection results can be obtained by employing the phase information \cite{MFD-11}. First, Qi et al. \cite{MFD-11-2} employ Phase spectrum of Fourier transform (PFT) to enhance targets. Then, Qi et al. \cite{MFD-11-1} present a method using a phase spectrum of quaternion Fourier transform (PQFT). In this model, the facet model allow to get four directional features which are employed to perform PQFT. The amplitude spectrum values are set to a constant value for each PQFT feature so that the saliency map consists of the inverse transform of the phase spectrum. Nevertheless, these two methods only employ the phase spectrum to recover signals, which will cause some information loss. It also lead to missed detection in the results. In an other approach, Yang et al. \cite{MFD-11-3} design a  method based on another direction feature. The filter is obtained by Gauss kernel function. Comparing with Fourier transform, cosine transform only performs on real even function and contains only real cosine terms. It is a real number of domain transformation. The actual signals are mostly in the real number domain, which conducts the Fourier transform very redundant. Moreover cosine transform allow to reduce computational complexity. According to these facts, Zhang et al. \cite{MFD-11} develop a method  based on spatial-temporal enhancement using quaternion Discrete Cosine Transform (QDCT). This method can support four channels, offering more information for DCT, whereas traditional DCT can only use gray information. Based on the fact that a small target often makes HVS have a higher response, the infrared small target detection is then formulated as a salient region detection problem in the transform domain.

In summary, HVS methods in the spatial or frequency domain achieve better detection performance in complex scenes than filters-based methods. It is because they address the contrast between the background and the targets.

\subsection{Multi-Frame Detection}

\subsubsection{Multi-Frame Cumulation Algorithms}
In 2007, Lu \cite{PSTD-5-1} designs a method based on higher-order cumulant. In 2016, Tang et al. \cite{PSTD-5-2} propose a similar approach based on morphology and high-order statistics. In 2017, Fan et al. \cite{MFD-10} employ a self-adaptive caliber temporal-spatial filtering. The accumulated image obtained by these three algorithms \cite{PSTD-5-1,PSTD-5-2,MFD-10} increases the number of accumulated frames. The target signal is enhanced but the streak noise is also enhanced. Because these three algorithms did not use the non-uniformity correction algorithm in the early stage, the background image still has more streak noise.  The algorithm \cite{PSTD-5-1} only employs the information of pixel points in time domain to estimate the background generating more noise in the difference image. The algorithm \cite{MFD-10} only employs the information of the pixel points in the spatial domain for background estimation. Thus, it generates a difference map with partial noise. However, the algorithm \cite{PSTD-5-2} combines the spatial and temporal information of pixel points to estimate the background. It makes full use of the spatial and temporal information of the pixel and makes the target signal in the difference image preserved.  \\

To address these limitations, Fan et al. \cite{PSTD-5} design in 2019 a non-uniform correction method of ring perturbation in sequence images. This method removes the non-uniformity of images. Then, a multi-frame cumulation technique enhances the dim targets. A Poisson distribution is employed to obtain candidate target images and a high-order cumulant allows to extract effective targets. Thus, the motion estimation is achieved on the dim target to make accumulation along the estimated direction. Experiments demonstrate that this ring perturbation correction method enhances the local signal-to-noise ratio (SNR) of dim small targets and provides better detection results than other algorithms \cite{PSTD-5-1,PSTD-5-2,MFD-10}, and can effectively detect dim small targets in strong clutter background \\

In 2017, Fan et al.  \cite{MFD-3} employs an Improved Anisotropy for Background Prediction (IABP) with a target enhancement based on High-Order Cumulates (HOC). A sequential image energy center detection algorithm integrates the neighborhood, continuity, area, and energy and other motion characteristics of the target. Experiments demonstrate that IABP provides a superior overall performance compared to previous background prediction methods increasing the signal-noise ratio of images. 

\subsubsection{Miscellaneous Algorithms}
In 2004, Peng et al.  \cite{MFD-4} design a strategy based on various nonlinear features that are representative of dim targets. An information fusion algorithm by Karhunen-Loeve (KL) transform is employed to obtain an integrated feature. In 2014, in the context of maritime scenes, Wang et al. \cite{MFD-5} propose a pipeline-filtering algorithm with self-adaptive anti-vibration abilities. First, the inter-frame vibration of sequence images is acquired in real time within the block matching method that combines the normalized cross-correlations coefficient with the normalized mutual information. Second, it is used to correct coordinates at the single-frame level Third, the corrected detection results allows to complete the pipeline-filtering. Experiments show that the algorithm can handle the challenge of inter-frame vibration of sequence images achieving accurate detection of small maritime targets.\\

In 2018, Liu et al.  \cite{MFD-6} develop a method based on flux density and direction diversity in gradient vector field. First, the infrared image is transformed to the Infrared Gradient Vector Field (IGVF). Second, a multiscale flux density (MFD) quantifies the extent of sink point character. In the MFD map, the small targets are enhanced and background clutters are suppressed. Third, a Gradient Direction Diversity (GDD) is employed. The residual noise due to the heavy clutter is suppressed by GDD. Finally, an adaptive threshold allows to separate the targets. Experiments show that IGVF outperforms previous SOTA methods in  complex background clutter. 

\begin{table*}
\begin{center}
\scalebox{0.79}{
\begin{tabular}{|l|l|}
\hline
\scriptsize{Category} &\scriptsize{Acronyms} 
\\
\hline
\hline
\scriptsize{Transform Domain Methods}  &\scriptsize{PFT (Qi et al.\cite{MFD-11-2}), PQFT (Qi et al.\cite{MFD-11-1}), QDCT (Zhang et al. \cite{MFD-11})} \\
\scriptsize{}  &\scriptsize{NSST-PQFT (Ren et al. \cite{MFD-11-4})} \\
\scriptsize{Classification Methods}  &\scriptsize{Bi et al. \cite{SFD-Classification-1}, Wu et al. \cite{SFD-Classification-3}} \\
\scriptsize{} &\scriptsize{DensityPeaksIR (Huang et al. \cite{SFD-Classification-5}), HSDW (Liu et al. \cite{SFD-Classification-4}), Sample Data (Fan et al. \cite{SFD-Classification-7})} \\
\scriptsize{Clustering Methods} &\scriptsize{Monte Carlo Linear Clustering  (Li et al. \cite{PSTD-37})} \\
\scriptsize{} &\scriptsize{Hyperbolic Density Clustering: HDC (Zhu et al. \cite{Miscellaneous-200}), PA-AHDPC (Zhu et al. \cite{Miscellaneous-201}), HWM-Det (Zhu et al. \cite{Miscellaneous-202}) } \\
\scriptsize{}  &\scriptsize{LDMGGC (Zhu et al. \cite{Applications-S-32}), IFCM-MF (Chen and Lin \cite{SFD-Classification-6})} \\
\hline
\scriptsize{Statistical Non Parametric Methods}  &\scriptsize{Kernel-based (Gu et al. \cite{SFD-Miscellaneous-7}), Neyman-Pearson Criteria (Askar et al. \cite{SFD-Miscellaneous-10}), Fast Algorithm (Zhang et al. \cite{SFD-Miscellaneous-11})} \\
\scriptsize{}  &\scriptsize{High-Order Statistics (Jiao and Wu \cite{PSTD-17})} \\
\scriptsize{}  &\scriptsize{Collinearity (Zhang et al. \cite{SFD-Miscellaneous-12})} \\
\hline
\scriptsize{Discriminative Methods}  &\scriptsize{Independent Component Analysis: FICA-TD (Luo et al. \cite{SFD-Discriminative-1}) } \\
\hline
\scriptsize{Miscellaneous Methods} &\scriptsize{Weighted Image Entropy (Deng et al. \cite{PSTD-1})} \\
\scriptsize{} &\scriptsize{Multiscale Kurtosis Map Fusion  (Xin et al. \cite{PSTD-36})} \\
\scriptsize{}  &\scriptsize{Multiscale Fuzzy Metric (Deng et al. \cite{PSTD-2})} \\
\scriptsize{}  &\scriptsize{Variational Mode Decomposition:(NVMD) [Band-Limited Subsignals] (Wang et al. \cite{PSTD-3})} \\
\scriptsize{} &\scriptsize{Fast Adaptive Scaling Algorithm (FAMSIS) (Chen et al. \cite{PSTD-12})} \\
\scriptsize{} &\scriptsize{Real-Time Implementations: SRD-1BVFM  (Zhou and Wang \cite{RT-6})}\\
\scriptsize{} &\scriptsize{FPGA Implementations: FAST (Wu et al. \cite{PSTD-10})}\\
\hline
\end{tabular}}
\caption{Conventional Unsupervised Methods (Part 2): Overview of Single-Frame Detection (28 papers).}
\label{OverviewSFD-2}
\end{center}
\end{table*}

\begin{table*}[t!]
\begin{center}
\scalebox{0.79}{
\begin{tabular}{|l|l|l|}
\hline
\scriptsize{Category} &\scriptsize{Acronyms} 
\\
\hline
\hline
\scriptsize{Multi-Frame Cumulation Algorithms} &\scriptsize{Higher-Order Cumulant (Lu \cite{PSTD-5-1}), Morphology and High-order Statistics (Tang et al. \cite{PSTD-5-2})}\\
\scriptsize{}  &\scriptsize{Self-Adaptive Caliber Temporal-Spatial Filtering (Fan et al. \cite{MFD-10})}  \\
\scriptsize{}  &\scriptsize{Ring Perturbation non-uniformity Correction (Fan et al.  \cite{PSTD-5} )} \\
\hline
\scriptsize{} &\scriptsize{Local Energy Center: IABP (Fan et al.  \cite{MFD-3})} \\
\scriptsize{Miscellaneous Algorithms} &\scriptsize{Information Fusion:IF-KL (Peng et al.  \cite{MFD-4} )} \\
\scriptsize{} &\scriptsize{Antivibration Pipeline-Filtering: APF (Wang et al.  \cite{MFD-5} )} \\
\scriptsize{} &\scriptsize{Flux Density and Direction Diversity: IGVF (Liu et al.  \cite{MFD-6} )} \\
\hline
\end{tabular}}
\caption{Conventional Unsupervised Methods: Overview of Multi-Frame Detection (8 papers).}
\label{OverviewMFD}
\end{center}
\end{table*}

\begin{table*}[p]
\centering
\scriptsize
\setlength{\tabcolsep}{3pt}
\renewcommand{\arraystretch}{1.05}
\begin{tabular}{|>{\raggedright\arraybackslash}p{2.4cm}|>{\raggedright\arraybackslash}p{3.3cm}|>{\raggedright\arraybackslash}p{11.4cm}|}
\hline
Category & Methods & Acronyms \\
\hline
\hline
Consistency of Backgrounds & Top-Hat Filters & Top-Hat (Jackway \cite{SFD-TH-3-1}) \\
 & & Study (Soni et al. \cite{SFD-Miscellaneous-1}), Optimizing the structuring element: Morphological Neural Networks (Zeng et al. \cite{SFD-TH-3}), Morphological Genetic Algorithm (Zeng et al. \cite{SFD-TH-3}) \\
 & & WTH (Bai et al. \cite{SFD-TH-5}), BTH (Bai et al. \cite{SFD-TH-5}) \\
 & & MWTH (Bai et al. \cite{SFD-TH-5-1}), MBTH (Bai et al. \cite{SFD-TH-5-1}), TEO (Bai et al. \cite{SFD-TH-6}) \\
 & & Dynamic Queue (Wang et al. \cite{SFD-TH-7}), QGA (Deng et al. \cite{SFD-TH-4}) \\
 & & Multi-stage Morphological (NTFH) (Yavari \cite{SFD-TH-11}), Weighted Adaptive Ring (WARTH) (Li et al. \cite{SFD-TH-12}), Balanced Ring Top-Hat (Zhu et al. \cite{SFD-TH-14}) \\
 & & PM (Lu et al. \cite{SFD-TH-10}) \\
 & & Mexican-Hat (Zhang et al. \cite{SFD-TH-15}) \\
 & & Multi-Feature Fuzzy Decision-Making (Yang et al. \cite{SFD-TH-13}) \\
 & Gaussian Filter & DoG (Huang et al. \cite{PSTD-13}), Improved DoG (Ye et al. \cite{PSTD-14}), ASC (Ruitao et al. \cite{PSTD-16}), GWM (Zhang and Li \cite{PSTD-33}) \\
 & & ODG-CTRS (Jia et al. \cite{SFD-LCM-117}) \\
 & Least Mean Square Filters & TDLMS (Hadhoud and Thomas \cite{SFD-TDLMS-1}), BDLMS (Azimi et al. \cite{SFD-TDLMS-1-1}), MW-TDLMS (Bae et al. \cite{SFD-TDLMS-1-2}) \\
 & & VS-LMS (Kwong and Johnston \cite{SFD-TDLMS-1-3}), Neighborhood-TDLMS (Cao et al. \cite{SFD-TDLMS-3}) \\
 & & Block Statistics TDLMS (Bae et al. \cite{SFD-TDLMS-4}), Edge Directional 2D-TDLMS (Bae et al. \cite{SFD-TDLMS-5}) \\
 & & Bilateral TDLMS (Zhao et al. \cite{SFD-TDLMS-2}) \\
 & Wiener Filters & Median-Modified Wiener Filter (Liu et al. \cite{PSTD-4}) \\
 & Particle Filters & MPF (Li and Ji \cite{PSTD-18}) \\
 & Bilateral Filters & IBF (Meng et al. \cite{PSTD-11}) \\
 & Directional Second-Order Directional Derivative & DSODD (Yang et al. \cite{PSTD-7}) \\
 & Miscellaneous Filters & Max-Min Filter (Deshpande et al. \cite{SFD-MaxM-1}), MILL (Kim et al. \cite{SFD-Miscellaneous-16}) \\
 & & Morphological Filter (Tom et al. \cite{SFD-TH-1}) (Fortin \cite{SFD-TH-2}) (Li et al. \cite{SFD-Miscellaneous-2}) (Yao et al. \cite{SFD-Miscellaneous-5}), Adaptive Morphological Filter (Bai et al. \cite{SFD-Miscellaneous-6}), ASRG (Bai et al. \cite{SFD-Miscellaneous-15}), MMP (Zhao et al. \cite{PSTD-30}) \\
 & & Neural Network Filter (Shirvaikar and Trivedi \cite{SFD-Miscellaneous-3}), 2-D AL (Ffrench et al. \cite{SFD-Miscellaneous-4}) \\
 & & Spatio-Temporal Filter (Fan et al. \cite{SFD-Miscellaneous-13}), HSWF (Cui et al. \cite{SFD-Miscellaneous-14}) \\
 & & Closest-Mean (Han et al. \cite{PSTD-19}), LCI Filter (Cao et al. \cite{PSTD-32}), MDMSHB (Peng et al. \cite{PSTD-34}) \\
\hline
Saliency of Targets & Local Contrast Measurements & LCM (Chen et al. \cite{SFD-LCM-4}), ALCM (Yi et al. \cite{PSTD-6}), ILCM (Han et al. \cite{SFD-LCM-5}), IDoGb (Han et al. \cite{SFD-LCM-11}) \\
 & & RLCM (Han et al. \cite{SFD-LCM-12}), MPCM (Wei et al. \cite{SFD-LCM-13}), WLDM (Deng et al. \cite{SFD-LCM-14}), STLDM (Du and Hamdulla \cite{MFD-12}) \\
 & & FGDRLCM (Zhao et al. \cite{SFD-LCM-116}) \\
 & & Facet Kernel (Qin et al. \cite{SFD-LCM-15}), NLCD (Qin et al. \cite{SFD-LCM-16}), STSM (Li et al. \cite{SFD-LCM-17}) \\
 & & WSLCM (Han et al. \cite{SFD-LCM-18}), TLLCM (Han et al. \cite{SFD-LCM-19}), ASPCM (Qiu et al. \cite{SFD-LCM-21}) \\
 & & MLCM-LEF (Xia et al. \cite{SFD-LCM-30}), ELCM (Guan et al. \cite{SFD-LCM-32}), NWLCM-CDC (Jia et al. \cite{Applications-S-34}) \\
 & & MRF-LCE (Li et al. \cite{SFD-LCM-101}), LGFFCM (Xiong et al. \cite{SFD-LCM-102}), MFLCM (Wu and Ren \cite{SFD-LCM-103}) \\
 & & LGDI-LGW (Dong et al. \cite{Applications-S-61}), TLSD (Li et al. \cite{SFD-Miscellaneous-17}) \\
 & & Derivative Scheme: MDWCM (Lu et al. \cite{SFD-LCM-33}), DECM (Bai and Bi \cite{SFD-LCM-34}), LCWMD (Xu et al. \cite{SFD-LCM-71}), FODD (Zheng and Zong \cite{SFD-LCM-75}) \\
 & & MRF-LCE (Li et al. \cite{SFD-LCM-35}), (Qiu et al. \cite{SFD-LCM-41}), MLGDR (He et al. \cite{PSTD-15}) \\
 & & CSWCM (Ye et al. \cite{SFD-LCM-100}), LCM-GR (Jiang et al. \cite{SFD-LCM-107}), HTCM (Guan et al. \cite{SFD-LCM-111}) \\
 & & TCBS (Zhou and Wang \cite{PSTD-60}) \\
 & & Entropy Scheme: EW-MLC (Wei et al. \cite{SFD-LCM-42}), EVWLCM (Xi et al. \cite{SFD-LCM-94}), MDLDE (Yao et al. \cite{SFD-LCM-97}) \\
 & & Layer Scheme: DL-LCM (Ren et al. \cite{SFD-LCM-43}), DLPC (Liu et al. \cite{SFD-LCM-67}) \\
 & & TrLCM (Luo et al. \cite{SFD-LCM-44}), SDWTLLCM (Chen et al. \cite{SFD-LCM-45}), WTLLCM (He et al. \cite{SFD-LCM-72}), STLCM (Yu et al. \cite{SFD-LCM-73}) \\
 & & MLCF-STC (Wang et al. \cite{SFD-LCM-60}), Improved TLLCM (Wang et al. \cite{SFD-LCM-99}) \\
 & & Relative Scheme: LRICM (Xie et al. \cite{SFD-LCM-95}) \\
 & & PFT (Shao et al. \cite{SFD-LCM-62}), MLLCM (Sun et al. \cite{SFD-LCM-63}), DLCM (Wu et al. \cite{SFD-LCM-43-1}), RDLCM (Lu et al. \cite{SFD-LCM-64}), LIG (Zhang \cite{SFD-LCM-43-2}) \\
 & & CSS-LCM (Hou et al. \cite{SFD-LCM-65}), STLECS (MDIMG) (Xi et al. \cite{SFD-LCM-66}) \\
 & & Weighted Scheme: TAWGIF (Wang et al. \cite{SFD-LCM-90}), WSTRLCM (Hu et al. \cite{SFD-LCM-91}) \\
 & & WIDLCM (Wang et al. \cite{SFD-LCM-96}), ODSWA (Li et al. \cite{SFD-LCM-98}) \\
 & & WIRLCM (Zhao et al. \cite{SFD-LCM-113}), SF-LIWGM (Xu et al. \cite{SFD-LCM-114}), WDDSM (Cui et al. \cite{SFD-LCM-115}) \\
 & & LOG Scheme: MSLCM-LOG (Chen et al. \cite{SFD-LCM-69}), ELC-LOG (Ma et al. \cite{SFD-LCM-70}) \\
 & & FWLCM (Jiang et al. \cite{SFD-LCM-74}), ESRLCM (Hu et al. \cite{SFD-LCM-92}) \\
 & & Shrinkage Problem (Xu et al. \cite{SFD-LCM-112}) \\
 & & Suppression Scheme: ARMS (Kong et al. \cite{DL-206}) \\
 & & Spatiotemporal Saliency (Pang et al. \cite{PSTD-31}) \\
 & & Multifactor Detectability Framework (Zhao et al. \cite{DL-334}) \\
 & Global Contrast Measurements & GCM (Tang et al. \cite{SFD-GCM-1}), MSGCM (Tang et al. \cite{SFD-GCM-2}), FGCM (Tang et al. \cite{SFD-GCM-3}) \\
 & & IGCM (Xie et al. \cite{SFD-GCM-4}), TGCM (Han et al. \cite{SFD-GCM-5}), RGCM (Han et al. \cite{SFD-GCM-6}) \\
& & FGCM (Han et al. \cite{SFD-GCM-7}) \\
\hline
\end{tabular}
\caption{Conventional Unsupervised Methods (Part 1): Overview of Single-Frame Detection (123 papers).}
\label{OverviewSFD-1}
\end{table*}

\section{Low-rank and Sparse Decomposition Unsupervised Models}
\label{LRSD}
Low-rank plus Sparse matrix Decomposition (LRSD) models are a dominant model-based approach in infrared small target detection. The LRSD model is also called Robust Principal Component Analysis (RPCA) introduced in 2009 by Candès et al \cite{MOD-Theory-1} to handle robustly sparse outliers in observed data. The original decomposition was extended to three terms decomposition (i.e low-rank matrix, sparse matrix and noise matrix) in 2010 by Zhou et al. \cite{MOD-Theory-2}. This version of RPCA is also known as stable RPCA in literature. Furthermore, the decomposition has been also extended in the tensor domain \cite{MOD-Theory-10} to take into account explicitly the spatio-temporal constraints. However, the LRSD model has also proven this efficiency in the close field of moving objects detection for video-surveillance \cite{MOD-Survey-4-1} by offering the possibility of spatio-temporal algorithms \cite{MOD-Method-10,MOD-Method-11,MOD-Method-12}, incremental algorithms \cite{MOD-Method-13,MOD-Method-14,MOD-Method-15} crucial for real-time applications, and deep unrolled RPCA methods \cite{MOD-Method-30, MOD-Method-31,MOD-Method-32}.  The rise and success of the LRSD models for infrared small target detection are based on the fact that it offers a rigorous mathematical foundation by jointly optimizing the nuclear norm (background regularization) and $l_1$-norm in contrast to filtering methods for background suppression or human visual-inspired contrast enhancement techniques. Moreover, the convex optimization framework demonstrates robustness when handling complex dynamic infrared backgrounds, and variable-sized targets across heterogeneous environments. The progress of LRSD methods in infrared small target detection follows a similar scheme that the one developed in moving object detection but using designed algorithms for the challenges met in IRSTD. The reader can refer to surveys \cite{MOD-Survey-100,MOD-Survey-101,MOD-Survey-102,MOD-Survey-300} and the DLAM website\footnote{https://sites.google.com/site/robustdlam/} for a list of representative fundamental papers about RPCA. \\

\subsection{Mathematical Preliminaries}
Candès et al. \cite{MOD-Theory-1} proposed a convex optimization to address the robust PCA problem also called LRSD. The observation matrix $A$ is assumed represented as: 
\begin{equation}
A = L + S
\label{EquationPCP1}
\end{equation}
where $L$ is a low-rank matrix and $S$ must be sparse matrix with a small fraction of nonzero entries. The straightforward formulation is to use $l_0$-norm to minimize the energy function:
\begin{equation}
\underset{L,S}{\text{min}} ~~ rank(L) + \lambda ||S||_{l_{0}}  ~~~ \text{subj} ~~~ A-L-S=0
\label{EquationPCP2}
\end{equation}
where $\lambda>0$ is an arbitrary balanced parameter.
But this problem is $NP$-hard, typical solution might involve a search with combinatorial complexity. This research seeks to solve for $L$ with the following optimization problem:
\begin{equation}
\underset{L,S}{\text{min}} ~~ ||L||_* + \lambda ||S||_{l_{1}}  ~~~ \text{subj} ~~~ A-L-S=0
\label{EquationPCP3}
\end{equation}
where $||.||_*$  and $||.||_{l_{1}}$ are the nuclear norm (which is the $l_1$-norm of singular value) and $l_1$-norm, respectively, and
$\lambda>0$ is an arbitrary balanced parameter. Usually, $\lambda=\frac{1}{\sqrt{max(m,n)}}$. Under these minimal assumptions, this approach called Principal Component Pursuit (PCP) solution perfectly recovers the low-rank and the sparse matrices. \\

\begin{figure*}[!t]
\begin{center}
\includegraphics[width=12cm]{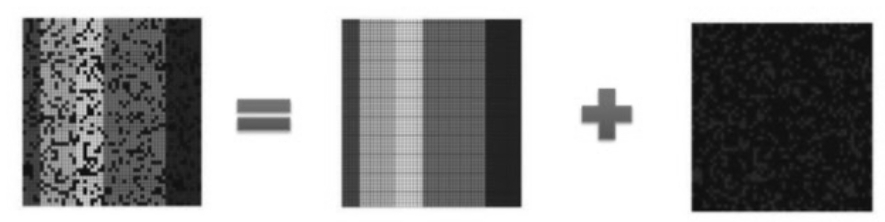}
\caption{Illustration of the decomposition of an observed matrix $A$ into a low rank matrix $L$ and a sparse matrix $S$.} 
\label{RPCAMatrix}
\end{center}
\end{figure*}

\indent  The low-rank minimization concerning $L$ offers a suitable framework for background due to the correlation between frames. So, minimizing $L$ and $S$ implies that the background is approximated by a low-rank subspace that can gradually change over time, while
the targets  constitute the correlated sparse outliers which are contained in $S$. To obtain the target mask, $S$ needs to be thresholded. The threshold is determined experimentally. $rank(L)$ influences the number of modes of the background that can be represented by $L$: If $rank(L)$ is to high, the model will incorporate the targets in its representation; if the $rank(L)$ is to low, the model tends to be uni-modal and then the multi-modality which appears in complex backgrounds will be not captured. The quality of  target detection is directly related to the assumption of the low-rank and sparsity of the background and targets, respectively.
In this case, the best separation is then obtained only when the optimization algorithm has converged. Essentially, the nuclear-norm term corresponds to low-frequency components along the temporal while the $l_1$ norm describes the high-frequency components.  Figure \ref{RPCAMatrix} shows an illustration of the decomposition of an observed matrix $A$ into a low rank matrix $L$ and a sparse matrix $S$. \\

The key characteristics of the LRSD problem formulations are the following ones:
\begin{itemize}
\item \textbf{Decomposition:} The form of the decomposition can be implicit or explicit. Furthermore, this decomposition can be made in two \cite{MOD-Theory-1} or three \cite{MOD-Theory-2} terms. Thus, all the decompositions can be written in a general formulation as follows:
\begin{equation}
A= \sum_{k=1}^K M_k = M_1+ M_2+M_3= L+S+E 
\label{EquationDecomposition1}
\end{equation}
with $K \in \left\{1,2,3\right\}$. The matrix $A$ can be corrupted by outliers where matrix $M_1=L$ is a low-rank matrix, the second matrix $M_2$ is an unconstrained (residual) matrix. $M_2$ is a sparse matrix $S$in explicit decomposition. The third matrix $M_3$ is generally the noise matrix $E$. The noise can be modeled by a Gaussian, a mixture of Gaussians (MoG) or a Laplacian distribution.
\item \textbf{Minimization problem:} The problem of the decomposition seeks to a minimization problem written in its original form or its Lagrangian form. Practically, the minimization problem is viewed as an optimization problem which can be convex or not. More specifically, the minimization problem seeks to the following optimization problem: $F(X)= f(x)+ g(x)$ where we have:
\begin{equation} 
f(x)=\frac{\lambda_0}{2}||A-L-S-E||_{norm_0}^{p_0} 
\end{equation}
\begin{equation}
g(x)=\lambda_1 ||L||_{norm_1}^{p_1} + \lambda_2 ||S||_{norm_2}^{p_2} + \lambda_3 ||E||_{norm_3}^{p_3}
\label{EquationMinimization}
\end{equation}
where we have: 
\begin{itemize}
\item $f(x)$ is smooth and convex function which has Lipschitz continuous gradients. $f(x)$ contains the loss function.  
\item $g(x)$ can be a nonsmooth and nonconvex function. $g(x)$ contains the low-rank regularizer, the sparse regularizer and the noise regularizer. \\

Equation \ref{EquationMinimization} can be written as follows:
\begin{equation}
\underset{L,S}{\text{min}} ~~  \lambda_1 ||L||_{norm_1}^{p_1} + \lambda_2 ||S||_{norm_2}^{p_2} + \lambda_3 ||E||_{norm_3}^{p_3} ~~~ \text{subj} ~~~ C
\label{EquationMinimizationSPCP}
\end{equation}
where the $\lambda_i$ are the regularization parameters. $C$ is a constraint on $A$, $L$, $S$ and $E$.
\end{itemize}
\item \textbf{Regularization Terms/Loss Functions:}  Loss functions are used for the minimization term, and the regularized functions are used to enforce the low-rank, sparse and noise constraints on $L$, $S$ and $E$, respectively. They are needed to solve the minimization problem. In literature, most of the time, the authors do not distinguish "loss functions" and "regularized functions", and they use them indifferently. In this paper, the regularized functions are respectively noted $f_{low}()$, $f_{sparse}()$ and $f_{noise}()$, and we refer to them as "regularization terms", the term "loss functions" being kept for deep learning models (Part 2). Thus, several loss functions can be used to enforce the constraints on each matrix which composes the decomposition. Most of the time, proxy loss functions are used as surrogate of the original loss functions (rank() and $l_0$-norm) to obtain a solvable problem. The surrogate lost functions present the main advantage that they allow to reach to a solvable convex problem but their use can present main advantages/disadvantages as follows: 
\begin{itemize}
\item The nuclear loss function is the surrogate proposed by Candès et al. \cite{MOD-Theory-1}.The Nuclear norm ($\|\cdot\|_{*}$) is defined as follows:
\begin{equation}
  \|\mathbf{L}\|_{*} = \sum_{i} \sigma_i(\mathbf{L}).
\end{equation}
where $\mathbf{L}$ is the low-rank (background) matrix and $\sigma_i$ 
denotes its $i$-th singular value.The tightest convex surrogate of the rank function over the unit spectral norm ball \cite{SFD-RPCAMatrix-1}. It is convex, but it causes target over-shrinkage because all singular values are penalised equally, which over-shrinks the background structure. Weighted \cite{SFD-RPCATensor-3} and partial-sum tensor \cite{SFD-RPCATensor-5} variants address this limitation. \\
The Frobenius norm ($\|\cdot\|_{F}$) is defined as follows:
\begin{equation}
  \|\mathbf{L}\|_{F} = \sqrt{\sum_{i,j} l_{ij}^2},
\end{equation}
where $l_{ij}$ denotes the $(i,j)$-th element of $\mathbf{L}$.  
A valid proxy for the nuclear norm \cite{SFD-RPCAMatrix-1, PE-4}, but fails to recover the low-rank matrix without a rank estimate and exhibits poor suppression of strong background edges \cite{SFD-RPCAMatrix-90}. \\

\item  The $l_1$-loss function also known as least absolute deviations (LAD) or least absolute errors (LAE) is robust to outliers but it is not stable and do not reach a unique solution. In addition, it may be suboptimal, since the $l_1$-norm is a loose approximation of the $l_0$-norm and often leads to an over-penalized problem. Furthermore, $l_1$-loss function cannot handle collinearity. The $\ell_1$-norm is defined as follows:
\begin{equation}
  \|\mathbf{S}\|_1 = \sum_{i,j} |s_{ij}|,
\end{equation}
where $s_{ij}$ denotes the $(i,j)$-th element of $\mathbf{S}$. 
It over-penalises sparse components as a loose approximation of $\|\cdot\|_0$ 
\cite{SFD-RPCAMatrix-1}. A weighted variant better preserves 
target energy \cite{SFD-RPCAMatrix-10}. \\

\item The $\ell_2$-loss function is sensitive to outliers and missing data, but it is stable and always reaches a unique solution. Its column-wise variant, the $\ell_{2,1}$-norm, is defined as follows:
\begin{equation}
  \|\mathbf{E}\|_{2,1} = \sum_{j}\sqrt{\sum_{i} e_{ij}^2},
\end{equation}
where $\mathbf{E}$ denotes the structured clutter component. 
More robust than $\ell_1$ to correlated background edges and 
clutter residuals \cite{SFD-RPCAMatrix-10}. \\

\item The $l_p$-loss function is more general than the $l_2$-loss function. The $\ell_p$-norm ($0 < p < 1$) is defined as follows.
\begin{equation}
  \|\mathbf{S}\|_p = \Bigl(\sum_{i,j}|s_{ij}|^p\Bigr)^{1/p}.
\end{equation}
Non-convex but a tighter approximation of $\|\cdot\|_0$ than 
$\ell_1$, recovering sparser solutions with fewer false detections 
\cite{SFD-RPCAMatrix-11}.
\item The Least Squares (LS) loss function is known to be very sensitive to outliers too  but the least squares loss function is a suitable solution in applications in which it is needed to take into account any and all outliers. The Least Squares (LS) is defined as follows:
\begin{equation}
  \mathcal{L}_{\mathrm{LS}} = 
  \|\mathbf{A} - \mathbf{L} - \mathbf{S}\|_F^2,
\end{equation}
where $\mathbf{A}$ is the observed infrared image. Stable and always 
yielding a unique closed-form solution, but highly sensitive to 
outliers and missing data \cite{SFD-RPCAMatrix-1}.
\end{itemize}

\indent From the computational point of view, the $\ell_1$-norm is non-smooth, but its proximal operator has a closed form (soft-thresholding), as does the $\ell_2$-norm; both problems can thus be solved efficiently. $l_1$-norm solutions can take into account the sparsity properties which allows it to be used for sparse constraints, which makes the computation more efficient than the $l_0$-norm \cite{SFD-RPCAMatrix-41}. \\

\item \textbf{Solvers:} Algorithms which are called solvers are then used to solve the optimization problem. Furthermore, instead of directly solving the original convex optimizations, some authors use their strongly convex approximations in order to design efficient algorithms. The key challenges related to the solvers are the following ones: 
 \begin{enumerate}
\item Choice of the solver to make the iterations as few as possible.
\item Choice of the SVD algorithm to make the iterations as efficient as possible. 
\end{enumerate}
\indent The solvers can be broadly classified into two categories as follows: 
\begin{itemize}
\item \textbf{Regularization based approaches:} The decomposition is formulated as regularized fitting, where the
regularizers are convex surrogates for rank and sparsity. Examples of algorithms in this category include the following solvers: Singular Value Thresholding (SVT), the Accelerated Proximal Gradient (APG) , and the Augmented Lagrange Multiplier (ALM).  
\item \textbf{Statistical inference based approaches:} Hierarchical statistical models are used to model the data generation process and prior distributions are selected to capture the low-rank and sparse properties of the respective terms. The joint distribution involving the observations, unknown variables and hyperparameters can be determined from the priors and conditional distributions. Posterior distributions of the unknown variables are approximated using Bayesian inference approaches. 
\end{itemize}  	 
\end{itemize}

\subsubsection{Definition and Classifications of Outliers}
The aim of RPCA is to be robust to outliers in the context of big data. Thus, the notion of outliers needs to be defined. Outliers can be defined as arbitrarily large valued measurements which do not characterize the true data samples. Practically, outliers can be viewed as observations or values that are considerably different from the majority of the data and usually follow heavy-tailed distribution by assuming that the unobservable noise have a fast-decaying Gaussian probability
distribution. As the observed examples are stacked as rows (or columns) in a data matrix, either some of the entries may be affected by \textit{1)} additive outliers (positive or negative values) or missing data, or \textit{2)} entire rows (or columns) are corrupted. Thus, outliers can be classified by considering their location in the observation matrix $A$ as follows: 
\begin{enumerate}
\item Element-wise outliers/noise. 
\item Row-wise or column-wise outliers/noise.
\item Missing data: They can be due to an acquisition problem.
\end{enumerate}
The shape of outliers varies according the observed data \cite{MOD-Survey-4-2}. In infrared small target detection, the outliers are the values corresponding to the target and the values corresponding to the background are the inliers. The outliers corresponding to the targets are element-wise outliers because of their size.

\subsubsection{Application to Infrared Small Target Detection} 
\label{subsubsec:ABFS} 
The adequacy of low-rank and sparse decomposition with the infrared small target detection relies on specific constraints not present in the basic assumptions of the first work published in 2011. These constraints are its robustness to noise, the  spatial and temporal constraints, the need of real-time implementations, and the ability to deal with the challenges met in infrared imaging sequences. In practice, researchers have proposed adaptations low-rank and sparse decomposition which attempt to take into account these specific constraints by acting on its key characteristics (i.e decomposition, formulation of the minimization problem, loss functions solvers). In this context, the minimization problem expressed in Equation \ref{EquationMinimizationSPCP} can be extended for small target detection in the following general formulation:
\begin{equation}
\begin{array}{ll}
\underset{L,S,E}{\text{min}} ~~  \underbrace{\lambda_1 f_{low}(L) + \lambda_2 f_{sparse}(S) + \lambda_3  f_{noise}(E)}_{Decomposition} \\
+ \underbrace{\lambda_4 f_{back}(L) + \lambda_5 f_{tar}(S)}_{Application} 
 ~~~ \text{subj} ~~~ C
\end{array}
\label{EquationMinimization2}
\end{equation}
where $\lambda_1$ up to $\lambda_5$ are regularization parameters. $C$ is the contraint which is often as $||A-L-S-E||_{norm_0}^{p_0}=0$. $f_{back}(L)$ and $f_{tar}(S)$ are regularization functions that allow the minimization to take into account the characteristics of the background and the target, respectively. $f_{back}(L)$ can be a mixed norm. $f_{tar}(S)$ can be the gradient, a Total Variation (TV), etc. Note that \textit{the first part} of Equation \ref{EquationMinimization2} with  $f_{low}(L)$, $f_{sparse}(S)$ and $f_{noise}(E)$ concerns mainly the decomposition into low-rank plus additive matrices, and \textit{the second part} with  $f_{back}(L)$ and $f_{tar}(S)$ concerns mainly the application to infrared small target detection. \\

Thus, the various LRSD based methods for infrared small target detection follow the same scheme and their changes are related to the decomposition, the minimization problem, the loss functions and the solvers. These changes have improved over time the suitability of the LRSD framework for infrared small target detection by addressing its challenges as possible.  Table \ref{Overview-1}, Table \ref{Overview-2} and Table \ref{Overview-3} provide an overview comparison of the diverse matrix decomposition  methods according to these criteria while Table \ref{OverviewT-1} and Table \ref{OverviewT-2} gives an overview of various tensor decomposition methods.
\\

Section \ref{LRSD-SFD} reviews the single-frame detection using matrix representations highlighting their progress over time. Section \ref{LRSD-SR} reviews sparse representation models. Section \ref{LRSD-MFD} surveys the multiple-frame detection using tensor representations. Note that even if data are either stacked in a matrix or tensor manner, the spatial and temporal information are kept in both case. However, it is only how the data are used that determines if the method is a single-frame or multi-frame one. In addition, the matrix format exploits either spatial (patch-image) or temporal (frame-stacking) correlations, whereas the tensor format allows both to be exploited jointly.  \\

\subsection{Single-Frame Detection}
\label{LRSD-SFD}
Single-frame detection using LRSD method are the methods that employ the matrix representation. In the IPI-based models, each column of the observed matrix $A$ is a vectorized patch extracted by a sliding window over a single image, so that only spatial (non-local) correlations are exploited.\\

In 2013, Gao et al. \cite{SFD-RPCAMatrix-1} propose the Infrared Patch-Image (IPI) model using the local self-correlation property of the global background patches. First, patch-images are constructed by using a window sliding from left and top to right and bottom in an image. Second, the target-background separation problem is reformulated as an optimization problem of recovering low-rank and sparse matrices. The most advantage of the IPI model is that it fits the reality well. Indeed, the low-rank assumption for the background patch-image is also commonly valid because the property of nonlocal self-correlation exists universally in infrared background images. Moreover, since the target size is small enough comparing with the original image size, the sparse assumption for the target patch-image is valid for almost all scenes. In practice, the IPI model present several weaknesses addressed by further works regarding the decomposition, the loss functions, the solver and the computation time.  These improvements can be categorized as done in the following sub-sections and Table \ref{OverviewMatrix} shows an overview classification of the diverse existing LRSD matrix methods applied to infrared small target detection.

\begin{table*}
\begin{center}
\scalebox{0.79}{
\begin{tabular}{|l|l|}
\hline
\scriptsize{Category}  &\scriptsize{Acronyms} 
\\
\hline
\hline
\scriptsize{RPCA Matrix Decomposition} &\scriptsize{IPI [Single Subspace] (Gao et al. \cite{SFD-RPCAMatrix-1}), NOP [Non-Overlapping Patch](Yang et al. \cite{SFD-RPCAMatrix-90})} \\ 
\hline
\scriptsize{Nuclear-norm Defects} &\scriptsize{TV-PCP  [Single Subspace](Wang et al. \cite{SFD-RPCAMatrix-15})} \\ 
\hline
\scriptsize{} &\scriptsize{SBL (Guo et al. \cite{SFD-RPCAMatrix-60}) } \\
\hline
\scriptsize{$l_1$-norm Defects} &\scriptsize{GRLA (Zhou et al. \cite{SFD-RPCAMatrix-13}), DGRLA (Zhou et al. \cite{SFD-RPCAMatrix-14}), $l_1$–$l_2$ norm-TV (Shao et al. \cite{SFD-RPCAMatrix-41})} \\
\scriptsize{} &\scriptsize{LMRPCA (Zhou et al. \cite{SFD-RPCAMatrix-39}), TV-IIPI (Wan et al. \cite{SFD-RPCAMatrix-35}), PUOT  (Bertrand et al. \cite{SFD-RPCAMatrix-33}) } \\ 
\scriptsize{} &\scriptsize{HQ-RPCA (Liu et al. \cite{SFD-RPCAMatrix-91})} \\
\hline 
\scriptsize{Nuclear Norm/$l_1$-norm Defects} &\scriptsize{WIPI (CWRPCA) (Dai et al. \cite{SFD-RPCATensor-1}), G-GRLA (Zhou et al. \cite{SFD-RPCAMatrix-13}), TNLRS (Zhu et al. \cite{SFD-RPCAMatrix-30})} \\ 
\scriptsize{} &\scriptsize{NIPPS [Single Subspace] (Dai et al. \cite{SFD-RPCAMatrix-5}), NRAM [Single Subspace] (Zhang et al. \cite{SFD-RPCAMatrix-10})} \\
\scriptsize{} &\scriptsize{NOLC [Single Subspace] (Zhang et al. \cite{SFD-RPCAMatrix-11}), R$S_{1/2}$NIPI (Zhou et al. \cite{SFD-RPCAMatrix-16})}\\
\scriptsize{} &\scriptsize{KRPCA (Yan et al. \cite{SFD-RPCAMatrix-37}), WIPI-WNNM (Chen et al. \cite{SFD-RPCAMatrix-49})}\\
\scriptsize{} &\scriptsize{Bi-NNM (Rawat et al. \cite{SFD-RPCAMatrix-42})} \\
\scriptsize{} &\scriptsize{RCTVW (Liu et al. \cite{SFD-RPCAMatrix-40}), NRAM-MRW (Wang et al. \cite{SFD-RPCAMatrix-43}), DW-SSD (Dan et al. \cite{SFD-RPCAMatrix-48}) } \\
\scriptsize{} &\scriptsize{HiLV-LRSD  (Liu et al. \cite{SFD-RPCAMatrix-38}), SRLCM (Li et al. \cite{SFD-RPCAMatrix-77}), PGTSLr (Chen et al. \cite{SFD-RPCAMatrix-36}), ITPM-HiLV-LRSD (Liu et al. \cite{SFD-RPCAMatrix-73}) } \\
\scriptsize{} &\scriptsize{NARIRM (Deng et al. \cite{PSTD-35}), GJL-MDW (Li et al. \cite{SFD-RPCAMatrix-101})} \\
\scriptsize{} &\scriptsize{EWBI (Zhu et al. \cite{SFD-RPCAMatrix-103}), Structured RPCA (Zhang et al. \cite{SFD-RPCAMatrix-104})} \\ 
\scriptsize{} &\scriptsize{HG-HDR (Wu and Li \cite{SFD-RPCAMatrix-106}), TLRD (Yao et al. \cite{SFD-RPCAMatrix-107}), Improved WNNM (Deng et al. \cite{SFD-RPCAMatrix-110}) } \\
\scriptsize{} &\scriptsize{ETPTV-HQF (Liu et al. \cite{SFD-RPCAMatrix-115}) } \\
\hline
\scriptsize{Heterogeneous Backgrounds} &\scriptsize{LRR (Li et al. \cite{SFD-RPCAMatrix-2-0}), GreB (Pang et al. \cite{SFD-RPCAMatrix-18}) } \\ 
\scriptsize{} &\scriptsize{MW-IPI (Li et al. \cite{SFD-RPCAMatrix-21}), GPCP (Li et al. \cite{SFD-RPCAMatrix-44})} \\
\scriptsize{} &\scriptsize{LER (Cao and Xue \cite{SFD-RPCAMatrix-32}), SCPLBMA (Li et al. \cite{SFD-RPCAMatrix-12}), MR-IRSTD (Wu et al. \cite{SFD-RPCAMatrix-93})} \\
\hline
\scriptsize{Adaptivity} &\scriptsize{WNM-IPI (Wang et al. \cite{SFD-RPCAMatrix-19}) } \\ 
\hline
\scriptsize{Temporal Constraints} &\scriptsize{FIRST-Net (Hu et al. \cite{SFD-RPCAMatrix-94}), MC-RPCA (Wang et al. \cite{SFD-RPCAMatrix-114})} \\
\hline
\scriptsize{Robustness to Noise} &\scriptsize{TV Regularized (DoLP images) (Hu et al. \cite{SFD-RPCAMatrix-68})} \\ 
\hline 
\scriptsize{Computation Time} &\scriptsize{Sobolev Norm (Zheng and Li \cite{SFD-RPCAMatrix-1-1}), Compressive (Li et al. \cite{SFD-RPCAMatrix-1-2})} \\ 
\hline
\scriptsize{Deep Unfold LRSD Networks} &\scriptsize{Deep-LSP-Net (Zhou et al. \cite{SFD-RPCAMatrix-45}), RPCANet (Wu et al. \cite{SFD-RPCAMatrix-46}), BEmST (Deng et al. \cite{SFD-RPCAMatrix-47})} \\
\scriptsize{} &\scriptsize{RPCANet++ (Wu et al. \cite{SFD-RPCAMatrix-65}), LSDSSMs  (Lu et al. \cite{SFD-RPCAMatrix-66}), (Yin et al. \cite{SFD-RPCAMatrix-92}) } \\
\scriptsize{} &\scriptsize{LST-RPCANet (Wang et al. \cite{SFD-RPCAMatrix-95}), DURD (Hao et al. \cite{SFD-RPCAMatrix-96}), ASD (Hao et al. \cite{SFD-RPCAMatrix-105})} \\
\scriptsize{} &\scriptsize{FBDM-Net (Fan et al. \cite{SFD-RPCAMatrix-111}), SCDE-Net (Zhang et al. \cite{SFD-RPCAMatrix-112}), BAMFNet (Pan et al. \cite{SFD-RPCAMatrix-113})} \\
\scriptsize{} &\scriptsize{Patch Limitation: LRRNet (Zhang et al. \cite{SFD-RPCAMatrix-63}) } \\
\scriptsize{} &\scriptsize{Global/Local Information: DUSRNet (Deng et al. \cite{SFD-RPCAMatrix-61}), ISTDFN (Ma et al. \cite{SFD-RPCAMatrix-97})} \\
\scriptsize{} &\scriptsize{Dynamic Adaptivity: DRPCA-Net (Xiong et al.\cite{SFD-RPCAMatrix-64}), Aw-DuNet (Yang et al. \cite{SFD-RPCAMatrix-100})} \\ 
\scriptsize{} &\scriptsize{Temporal Dimension: ESTNet (Zhu and Feng \cite{SFD-RPCAMatrix-71})}\\
\scriptsize{} &\scriptsize{Lightweight Algorithm: L-RPCANet (Liu et al. \cite{SFD-RPCAMatrix-67}), LSNet (Li et al. \cite{SFD-RPCAMatrix-70})} \\
\scriptsize{} &\scriptsize{State Space Model: RPCASSM (Liu et al. \cite{SFD-RPCAMatrix-102})} \\
\hline
\end{tabular}}
\caption{Single-Frame Detection Unsupervised Matrix Decomposition Methods : An Overview (69 papers).}
\label{OverviewMatrix}
\end{center}
\end{table*}

\subsubsection{Computation Time}
The computation time limitation can be addressed by using surrogate loss function requiring less iteration in the optimization problem. In view of time-consuming problem in nuclear-norm minimization using singular value decomposition, Zheng and Li \cite{SFD-RPCAMatrix-1-1} employ the Sobolev norm instead of  the nuclear-norm, and thus a small target detection method based on harmonic and sparse matrix decomposition was proposed reducing the computation time. In an other way, Li et al. \cite{SFD-RPCAMatrix-1-2} develop a compressive low-rank and sparse decomposition method in which the constant $r$ and $K$ denote the low-rank of background and sparse of target, respectively. Then, the Walsh Hadamard matrix was chosen as the measurement matrix.

\subsubsection{l1-norm Defects} 
Because of the defect of $l_1$-norm based sparsity measure, the IPI model falls in a dilemma where either over-shrinks the dim targets or preserves the strong edges in the target image.  As the $l_1$-norm is applied on the sparse component related to the target, changing this norm  and adding constraints can increase the accuracy of the target detection. To suppress the strong edges while preserving the dim targets, Dai et al. \cite{SFD-RPCATensor-1}  propose a Weighted Infrared Patch-Image (WIPI) model, incorporating structural prior information into the separation process. The WIPI model is formulated as a Column-wise Weighted RPCA (CWRPCA) problem. In addition, a target unlikelihood coefficient is designed based on the steering kernel, serving as the adaptive weight for each column. To solve the CWRPCA problem, Dai et al. \cite{SFD-RPCATensor-1} design an Alternating Direction Method (ADM). Nevertheless, calculating the structural prior brings additional computing time. \\

In 2019, Zhou et al. \cite{SFD-RPCAMatrix-13} propose a Graph-Regularized Laplace Low-rank Approximation model (GRLA) to reduce high false or missing alarm when encountering the background with intricate interferences. GRLA is extended to a generalized framework name G-GRLA by replacing different non-convex rank functions. Employing an ADMM algorithm, the solution is obtained by an iterative optimization scheme. But GLRA exploit only the spatial information and  ignore the structural priors across feature  space.
To address this limitation, Zhou et al. \cite{SFD-RPCAMatrix-14} propose a LRSD model combining the spatial-feature graph regularization and $l_{1/2}$-norm sparse constraint. The spatial and feature regularizations are imposed on the sparse component in the form of graph Laplacians where the sparse component is enforced as the eigenvectors of their graph Laplacian matrices. This method allows to take into account the geometric information in both data and feature space simultaneously. The $l_{1/2}$-norm acts as a surrogate of $l_1$-norm to constrain the sparse component reducing the false positive detection. Zhou et al. \cite{SFD-RPCAMatrix-14} employ a Linearized Alternating Direction Method with Adaptive Penalty (LADMAP) for optimization algorithm. This method based on Dual-Graph Regularization  Low-rank Approximation (DGRLA) is more robust in the presence of complex scenes with cloudy clutter, sea glitter, and man-made interference than IPI \cite{SFD-RPCAMatrix-1} and GRLA \cite{SFD-RPCAMatrix-13}. \\

In 2022,  Wan et al. \cite{SFD-RPCAMatrix-35} propose to use the interframe similarity of target appearance in order to impose
a temporal consistency constraint on the target matrix. A TV regularization term is also employed to alleviate the false
detection generated by noise. To solve the problem, an alternating optimization algorithm using singular value decomposition (SVD) and accelerated proximal gradient (APG) is employed. Experiments show that TV-IIPI outperforms IPI \cite{SFD-RPCAMatrix-1}, ECA-STT \cite{SFD-RPCATensor-16} and PSTNN \cite{SFD-RPCATensor-5} in the case of sea and ground scenes. In 2023, Shao et al. \cite{SFD-RPCAMatrix-41} employ the $l_1$ and $l_2$ ($l_1$–$l_2$) norm and total variation (TV) to constraint the sparse component. The $l_1$–$l_2$ norm allows to separate the target from the complex backgrounds while the total variation norm  allows to suppress the sparse clutters. Experiments show that the $l_1$–$l_2$ norm-TV outperforms IPI \cite{SFD-RPCAMatrix-1}, NOLC \cite{SFD-RPCAMatrix-11}, NRAM \cite{SFD-RPCAMatrix-10}. Zhou et al. \cite{SFD-RPCAMatrix-39} propose a Learned Masked Robust Principal Component Analysis (LMRPCA) algorithm for single-frame infrared small target detection. First, the original images are constructed into patch images, which are separated into low-rank and sparse components corresponding to the backgrounds and foreground masks. The optimization function is solved by an DMM method which is mapped to trainable convolutional layers. More specifically, LMRPCA employs elements of convolutional sparse coding to improve representation learning for foreground masks and side information in the auxiliary transform domain. Thus, Zhou et al. \cite{SFD-RPCAMatrix-39} assign learnable weights to different feature maps with a reweighted $l_1-l_1$ minimization. Experiments show that LMRPCA outperforms PSTNN \cite{SFD-RPCATensor-5} and DNA-Net \cite{DL-12}.

\subsubsection{Nuclear-norm Defects}  
As the nuclear norm is applied on the low rank component related to the background, changing this norm  and adding constraints can increase the robustness in the presence complex and noisy backgrounds. In practice, the IPI model presents weaknesses when facing extremely complex backgrounds. This lies in the mismatching of model’s implicit assumption of a large number of observations with the reality of deficient observations of strong edge. To address this problem, Dai et al. \cite{SFD-RPCAMatrix-5} employ the partial sum of singular values instead of the nuclear norm to constrain the low-rank background patch-image. It provides a more accurate background estimation and almost eliminates all the salient residuals in the decomposed arget image. Second, considering the fact that the infrared small target is always brighter than its adjacent background, Dai et al. \cite{SFD-RPCAMatrix-5} use an additional non-negative constraint to the sparse target patch image. It reduces off more undesirable components but also accelerates the convergence rate. An inexact augmented Lagrange multiplier is designed for the formulation problem named Non-negative Infrared Patch-image Partial Sum (NIPPS).  In 2017, Wang et al. \cite{SFD-RPCAMatrix-15} propose to use a total variation (TV) regularization term on the low-rank matrix to describe the inner smooth and crisp edges of background, which is of great importance in non-smooth and nonuniform background. The total variation is based on a local gradient operator which weights the low rank component to reduce clutters in detection results,as well as to provide a good estimation of background region. The proposed method named TV-PCP is solved using augmented Lagrangian function. TV-PCP outperforms the IPI model \cite{SFD-RPCAMatrix-1}. \\

In 2018, Zhang et al. \cite{SFD-RPCAMatrix-10}  design a non-convex, tighter rank surrogate and weighted $l_1$ norm which can suppress the background better while preserving the target efficiently. More specifically, a structured $l_{2,1}$ norm is employed to wipe out the strong residuals. To solve this non-convex model, Zhang et al. \cite{SFD-RPCAMatrix-10}  design an non-convex optimization algorithm based on Alternating Direction Method of Multipliers (ADMM) plus Difference of Convex (DC) programming.  This method is called Non-convex Rank Approximation Minimization Joint $l_{2,1}$ norm (NRAM) method. In a further work, Zhang et al. \cite{SFD-RPCAMatrix-11} propose a LRSD method based on Non-convex Optimization with Schatten $q$-norm and $L_p$-norm Constraint (NOLC) requiring a non-convex optimization algorithm. The NOLC model can  detect more the targets but also suppress better the background area than IPI \cite{SFD-RPCAMatrix-1}, RIPT and NRAM \cite{SFD-RPCAMatrix-10}. In summary, the NRAM method is based on the structural noise and uses the $l_{2,1}$-norm constraint that emphasizes that the row of the block image is sparse. To reach this effect, the size of the structured noise must be smaller than the size of the sliding window to obtain the effect of row sparseness. However, for structured noise, its size often cannot meet the requirements (such as bridges and houses), which makes the NRAM using the $l_{2,1}$-norm constrained noise term sometimes less robust. The NOLC method is based on the difference between the target and the background sparsity from the perspective of the target, and directly employs the stricter $L_p$-norm to constrain the sparse item. This method allows to describe the target more directly and accurately than the IPI model and the NRAM method providing good detection effects under various complex backgrounds. Furthermore, NOLC always restores sparse targets in the noisy infrared small target image. The NOLC method improves the convergence strategy  making the convergence speed better than IPI \cite{SFD-RPCAMatrix-1} and NRAM \cite{SFD-RPCAMatrix-10} methods. \\

In 2019, Zhou et al. \cite{SFD-RPCAMatrix-16} propose the integration of Schatten 1/2 quasi-norm regularization and reweighted sparse enhancement in the LRSD framework. This method provides better approximation for the sparse-regularized low-rank function, exactly recovering the background patch-image. Second, an adaptive weight is applied to suppress the salient sparse outliers. A softening half-thresholding operator, instead of the original half-thresholding operator, is used to solve the Schatten 1/2 minimization subproblem. To deal with  interference of heavy clutter and strong edge, Yan et al. \cite{SFD-RPCAMatrix-37} adopt a Kernel RPCA (KRPCA) method to estimate the background component.  More specifically, a total variation regularization is employed to smooth the background while the $l_p$-norm is choosen to recover the sparse component and the graph Laplacian regularization is utilized as a sparse constraint to constrain the homogeneous target region. Experiments show that KRPCA outperforms IPI \cite{SFD-RPCAMatrix-1}, RIPT, TV-PCP and NRAM \cite{SFD-RPCAMatrix-10}. \\

In 2020, Zhu et al. \cite{SFD-RPCAMatrix-30}
design a Target-aware method based on a Non-local Low-Rank model and Saliency (TNLRS) filter regularization enabling joint target saliency learning in a lower dimensional discriminative manifold. First, a non-local patch construction is employed for the target-aware low-rank model. By combining similar patches, Zhu et al. \cite{SFD-RPCAMatrix-30} reconstruct them together to obtain a better generalization of non-local spatial sparsity constraints. To improve target saliency learning, the saliency filtering regularization term based on entropy is restricted to lie between the background and foreground. Thus, the regularization of the saliency filtering locally takes into account the contexts from the target and surrounding areas and limits the deviated approximation of the low-rank matrix. \\

In 2023, Liu et al. \cite{SFD-RPCAMatrix-38} design a High Local Variance LRSD (HiLV-LRSD) method. HiLV-LRSD allows to distinguish the sparse points with HLS, the non-low rank as the targets and the remaining regions as the background. First, a local variance allows
represent local salience. Thus, Liu et al. \cite{SFD-RPCAMatrix-38} employ an $LV^*$ norm to constrain the background’s low-rank and local variance. Second an adaptively reweighted ($l_v$,$l_1$)-norm allows to constrain the
sparsity of the target and enhance the influence of local variance.  Experiments show that HiLV-LRSD outperforms IPI \cite{SFD-RPCAMatrix-1}, ReWIPI, NIPPS \cite{SFD-RPCAMatrix-5}, NRAM \cite{SFD-RPCAMatrix-10} and PSTNN \cite{SFD-RPCATensor-5}. In 2023, Liu et al. \cite{SFD-RPCAMatrix-40} (RCTVW\footnote{\url{https://github.com/LiuTing20a/RCTVW}}) employ a total variational to constraint Representative Coefficient (RC) matrix instead of data matrix to describe local smooth prior to remove noise and to reduce computational complexity. Thus, Liu et al.  \cite{SFD-RPCAMatrix-40} control the number of columns in the representative coefficient matrix to maintain the low-rank characteristics of background to
avoid SVD calculation and to improve the detection efficiency.
The RCTV regularization simultaneously describe
local smooth prior and low-rank prior. Second, Liu et al. \cite{SFD-RPCAMatrix-40} employ a log-sum function to adaptively assign weights to targets in order to better increase the sparsity of targets and distinguish sparse non target points. Experiments show that RTCTW outperforms ASTTV \cite{SFD-RPCATensor-32} and NPSTT  \cite{SFD-RPCATensor-31}. \\

In  2024, Dan et al. \cite{SFD-RPCAMatrix-48} propose a Dynamic Weight-guided Smooth-Sparse decomposition (DW-SSD) method. First, a surface-fitting perspective allows to reconstruct the smooth vignetting using the B-spline as the basis. Second, Total Variation (TV) allows to regularize the coefficients of the B-spline basis. This term allows a flexible description for different background local regions with low computation cost dealing with the high brightness vignetting and improving the emergence of the target. Third, the dynamic weight based on spatial–temporal prior characteristic is employed for the target component, where the differences in spatial contrast and temporal profile are jointly used to guide the separation of the target from dense noise. Experiments show that DW-SSD outperforms PSTNN \cite{SFD-RPCATensor-5}.In 2024, Chen et al. \cite{SFD-RPCAMatrix-49} propose to use Weighted Nuclear Norm Minimization (WNNM) optimization with  sparse representation based on the local IPI model. Experiments show that the proposed method outperforms IPI and ReWIPI. 

\subsubsection{Adaptivity} 
The adaptivity is generally addressed by a matrix that weighted either the low-rank component or the sparse component by using local prior or computed information. First, Wang and Qin \cite{SFD-RPCAMatrix-3} propose an  adaptive weighting parameter LRSD method. More specifically, a multi-scale transform and patch transform are used to generate an image patch. Thus, the weighting parameter of RPCA no longer has global uniform value, but has an appropriate value according to the complexity of each patch. In addition, the image processing at the patch level allows parallel computing in multi-core systems to increase the speed of target-background separation even in real time. \\

In an other approach, Gao et al. \cite{SFD-RPCAMatrix-6} employ a weighting matrix $W$ to obtain the sparse matrix. The nonlinear weights are constructed based on a logistic sigmoid function, and offer more advantages than weights of singular values in suppressing background and keeping small targets. More specifically, the entry-wise weights are designed based on the local weighted entropy, and allow to extract the targets and to suppress background clutters. An Accelerated Proximal Gradient (APG)  is used to solve the corresponding optimization problem. At the end, the location correlation of the dense and sparse reconstruction maps and tracklet association of the location correlation maps for successive frames are employed to suppress false alarms and confirm suspected targets. Experiments show the performance this co-detection model in the airbone case. In 2017, Wang et al \cite{SFD-RPCAMatrix-19} propose a weight nuclear norm minimization based IPI model. This model consider each singular value differently instead of equally in the IPI model offering more flexibly to describe complex background. In 2018, Yuan and Dong \cite{SFD-RPCAMatrix-20} design an adaptive scheme based on the ratio minimization of singular values. More specifically, a multi-scale fusion and parameter regulation are employed. They are both based on the non-negative target assumption and the ratio between partial and total sum of background singular values. The parameters are thus regulated adaptively using the image entropy and the standard deviation of image patches. This method shows robustness in extremely variant and complex backgrounds, and is more accurate with low false alarm rate than IPI \cite{SFD-RPCAMatrix-1}. In 2021, Bertrand et al. \cite{SFD-RPCAMatrix-33} employ an Unbalanced Optimal Transport (UOT) method as a regularizer into the IPI model to take into account the temporal correlation
between frames. \\

In an other work, Cao and Xue \cite{SFD-RPCAMatrix-32} study the influence of various blocking methods under different image backgrounds. Thus, a selection method based on the larger value of the minimum local entropy of image segmentation is employed. Experiments show that the best results are obtained by taking the larger
value of the minimum local entropy as the reference.

\subsubsection{Heterogeneous Backgrounds} 
To deal with the highly heterogeneous background, Li et al. \cite{SFD-RPCAMatrix-2-0} employs a Low-Rank Representation (LRR) to describe the background patch-image, assuming that all background subspace clusters. But, this LRR based method does not perform as well as the IPI model in suppressing the background clutter and noise.\\

In 2020, Pang et al. \cite{SFD-RPCAMatrix-18} design a LRSD method based on greedy bilateral factorization inspired by by Zhou
and Tao \cite{MOD-Method-180}. First, the target is considered as an independent sparse motion structure by analyzing the complex structure information of the IR image sequences. Second, the greedy bilateral factorization (GreB) strategy is adopted to approximate the low-rank part of the algorithm by substituting $L$ by its bilateral factorization. This scheme also accelerates the efficiency of the algorithm. Experiments show that GreB outperforms IPI and NRAM \cite{SFD-RPCAMatrix-10}. In 2019, Li et al. \cite{SFD-RPCAMatrix-21} propose to exploit both local and nonlocal priors in a dual-window local contrast method (DW-LCM) and multiscale-window IR patch-image (MW-IPI). \\

In 2024, Li et al. \cite{SFD-RPCAMatrix-44} propose a customized  Group low-rank regularized Principal Component Pursuit model (GPCP). Experiments on ground backgrounds, sea backgrounds and sky backgrounds show that GPCP outperforms IPI \cite{SFD-RPCAMatrix-1} and NRAM \cite{SFD-RPCAMatrix-10}. Instead of equalizing the background area as in the original IPI \cite{SFD-RPCAMatrix-1}, Li et al. \cite{SFD-RPCAMatrix-12} a Statistical Clustering Partitioning Low-rank Background Modeling Algorithm (SCPLBMA). It separates the image into areas according to their statistics so that each area is consistent statistical distribution. Thus, SCPLBMA can preserve the natural structure of the image.\\

In 2026, Wu et al. \cite{SFD-RPCAMatrix-93} develop a Manifold Geometric Prior and Structural Rectification for Infrared Small Target Detection (MR-IRSTD) method. First, a spatiotemporal patch matrix is built. Second, a KD-tree-accelerated Local Linear Embedding (LLE)  extracts local geometric correlations to form a data-driven Manifold Geometric Prior, providing an optimal linear prediction for complex backgrounds. By shifting the nuclear norm constraint from the raw background to the manifold residual, the structural rectification mechanism absorbs the ghosting artifacts into the low-rank component them to avoid the contamination by the sparse targets. Third, the sparse targets and the rectified background are separated through an adaptive Alternating Direction Method of Multipliers (ADMM) algorithm. Experiments show that MR-IRSTD method offers superior robustness in complex scenarios with strong edges and noise than ASTTV and PSTNN.

\subsubsection{Robustness to noise} 
In order to better separate the target from noise, He et al. \cite{SFD-RPCAMatrix-2} propose a Low-Rank and Sparse Representation (LRSR) model, employing sparse representation for the small target. In 2018, Gao et al. \cite{SFD-RPCAMatrix-17} address the sparse noise component by adopting a Mixture of Gaussians (MoG) with Markov random field (MRF). The MRF guided MoG noise model under the Bayesian framework model the small target detection problem. Using a variational Bayesian, the small target component is separated from complex background noise. A simple adaptive segmentation method is used to extract small target.

\subsubsection{Temporal Constraints} 
Temporal constraints are addressed by FIRST-Net \cite{SFD-RPCAMatrix-94}, reviewed below with deep unfolding networks, and by MC-RPCA \cite{SFD-RPCAMatrix-114}. \\

In summary, the IPI model is the original method, and its principle and solution method are directly inherited from the proposed LRSD scheme of Candes et al. \cite{MOD-Theory-1}. The changes about the nuclear norm, the low-rank part and its adaptivity aim to improve the reconstruction of the background whilst the changes about the $l_1$-norm, the sparse part and its adaptivity aim to increase the accuracy of the target detection. The changes in the optimization algorithm concern both the solution of the problem formulation and the computation time.  \\

\subsubsection{Deep Unfold LRSD Networks}
Since the rise of deep learning, several authors propose deep low-rank and sparse neural networks  (also called deep unfold RPCA networks), namely Deep-LSP-Net \cite{SFD-RPCAMatrix-45}, RPCANet\footnote{https://github.com/fengyiwu98/RPCANet}\cite{SFD-RPCAMatrix-46}, DUSRNet\footnote{https://github.com/Awais-k/DUSRNet}\cite{SFD-RPCAMatrix-61}, LRRNet\footnote{https://github.com/HaLongbao/LRRNet/} \cite{SFD-RPCAMatrix-63}, DRPCA-Net\footnote{https://github.com/GrokCV/DRPCA-Net}\cite{SFD-RPCAMatrix-64}, RPCANet++ \footnote{https://fengyiwu98.github.io/rpcanetx/}\cite{SFD-RPCAMatrix-65}, LSDSSMs \cite{SFD-RPCAMatrix-66}, L-RPCANet\footnote{https://github.com/xianchaoxiu/L-RPCANet}\cite{SFD-RPCAMatrix-67}, LSNet \cite{SFD-RPCAMatrix-70}, DURD\footnote{https://github.com/haofan72/DURD}\cite{SFD-RPCAMatrix-96} and ESTNet \cite{SFD-RPCAMatrix-71}.  \\

More specifically, algorithmic unfolding has been introduced to map the iterative optimization process of conventional LRSD models into standard neural network architectures \cite{MOD-Theory-3}. This technique allows for the integration of domain knowledge into the network design, providing  more interpretable structures. RPCA-based unfolded networks (also called deep unfolding networks (DUNs)) have demonstrated superior detection performance with fewer training parameters compared to conventional residual neural networks in the field of moving object detection \cite{MOD-Theory-4,MOD-Theory-5}. \\ 

First, in 2023,  Zhou et al. \cite{SFD-RPCAMatrix-45} unroll the iterative optimization algorithm of ADMMs into deep neural networks, where a generalized sparsifying transform and a singular value thresholding (SVT) operator are learned by the Convolutional Neural Networks (CNNs) to avoid tedious parameter tuning and improve the interpretability of the neural networks. In 2024, Wu et al. \cite{SFD-RPCAMatrix-46} formulates the infrared small target detection task as  low-rank background estimation, sparse target extraction, and image reconstruction in a relaxed RPCA model. By unfolding the iterative optimization updating steps into a deep-learning framework, time-consuming and complex matrix calculations are replaced by theory-guided neural networks. This learnable deep network architecture derived from a relaxed RPCA model is named RPCANet. RPCANet detects targets with clear interpretability and preserves the intrinsic image feature, instead of directly transforming the detection task into a matrix decomposition problem. \\

In a further work, Wu et al. \cite{SFD-RPCAMatrix-65} propose RPCANet++ as a sparse object segmentation framework that fuses the interpretability of RPCA with efficient deep architectures and incorporates a memory-augmented module to preserve background features during background extraction and a deep contrastive prior module to accelerate target extraction during target approximation. More specifically, RPCANet++ unfolds a relaxed RPCA model into a structured network comprising a Background Approximation Module (BAM), an Object Extraction Module (OEM), and an Image Restoration Module (IRM). To mitigate inter-stage transmission loss in the BAM, Wu et al. \cite{SFD-RPCAMatrix-65} employ a Memory-Augmented Module (MAM) to enhance background feature preservation, while a Deep Contrast Prior Module (DCPM) exploits saliency cues to achieve object extraction. Experiments show that RPCANet++ outperforms NRAM \cite{SFD-RPCAMatrix-10} and RPCANet \cite{SFD-RPCAMatrix-46}. \\
In practice, RPCANet \cite{SFD-RPCAMatrix-46} presents several weaknesses addressed by further works and these improvements can be categorized as follows.
\begin{itemize}
\item \textbf{Few-shot Learning:} In 2026, Hu et al. \cite{SFD-RPCAMatrix-94} design a Few frame-guided spatio-temporal deep unfolding Network (FIRST-Net). It is a physics-driven architecture featuring hierarchical deep unfolding layers that explicitly encode target sparsity and background low-rank priors. Based on frame-level ablation studies, 5-frame configuration are selected as optimal achieving $74.6\%$ detection accuracy on high-velocity targets while maintaining real-time processing. Experiments demonstrate SOTA performance with $76.8\%$ IoU and $7.8 \times 10^{-5}$ false alarm suppression. FIRST-Net bridges the gap between single-frame efficiency and multi-frame robustness. \\
\item \textbf{Patch Limitation:} RPCANet \cite{SFD-RPCAMatrix-46}  attempts to unfold RPCA into a trainable network but imposes low-rank constraints on patch-images rather than on the entire image which results in suboptimal detection performance and significant computational overhead, especially on high-resolution infrared data. To address these limitations, Zhang et al. \cite{SFD-RPCAMatrix-63} design a patch-free Low-Rank Representation Network (LRRNet) that integrates low-rank constraints directly in the image domain avoiding the need for patch-image construction. This design is based on a key insight: the low-rank nature of infrared images implies an inherent compressibility—when the rank is sufficiently small, the image information can be sparsely encoded while being nearly “losslessly” preserved.
\item \textbf{Global/Local Information:} The global low-rank characteristic of infrared image is emphasized in LRSD decomposition and ignore local structural feature. To address this issue, Deng et al. \cite{SFD-RPCAMatrix-61} develop a Deep Unfolding Sparse Regularized Network (DUSRNet). DUSRNet combines the  feature extraction capability of deep learning with the fine structure description capability of sparse regularization. To adaptively describe the low-rank and sparse characteristics, the sparse-regularized model is embedded into the deep neural network in a end-to-end manner. Deng et al. \cite{SFD-RPCAMatrix-61} design an adaptive background estimation module, sparse target extraction module and infrared image reconstruction module. Experiments show that  DUSRNet outperforms RPCANet \cite{SFD-RPCAMatrix-46} on the NUDT-SIRST, SIRST and IRSTD-1k datasets in extremely high scene complexity and variability.
\item \textbf{Dynamic Adaptivity:} In 2025, Xiong et al.\cite{SFD-RPCAMatrix-64} propose a dynamic RPCA network (DRPCA-Net), a novel deep unfolding network (DUN) that integrates the sparsity-aware prior into a learnable architecture. Instead of conventional deep unfolding methods that rely on static, globally learned parameters, DRPCA-Net employs a dynamic unfolding mechanism via a lightweight hypernetwork. It enables the model to adaptively generate iteration-wise parameters conditioned on the input scene, thereby enhancing its robustness and generalization across diverse backgrounds. Moreover,  Xiong et al. \cite{SFD-RPCAMatrix-64} employ a Dynamic Residual Group (DRG) module to better capture contextual variations within the background achieving more accurate low-rank estimation and improved separation of small targets. 
\item \textbf{Lightweight Algorithm:} In 2025, Liu et al. \cite{SFD-RPCAMatrix-67} design a lightweight framework based on RPCA called L-RPCANet. A hierarchical bottleneck structure is provided to reduce and increase the channel dimension in the single-channel input infrared image to achieve channel-wise feature refinement, with bottleneck layers designed in each module to extract features. This strategy reduces the number of channels in feature extraction and improves the lightweightness of network parameters. In addition, a noise reduction module is embedded to improve the robustness against complex noise. Squeeze-andexcitation networks (SENets) are exploited as a channel attention mechanism to focus on the varying importance of different features across channels, thereby improving performance while maintaining both lightweightness and robustness. Experiments show L-RPCANet that outperforms or achieves similar results than RPCANet \cite{SFD-RPCAMatrix-46}, DRPCANet \cite{SFD-RPCAMatrix-64} and RPCANet++ \cite{SFD-RPCAMatrix-65}. In summary, L-RPCANet not only improves the detection performance and noise robustness, but also requires fewer parameters, achieves faster GPU inference time, and is more applicable to real-time detection.
\item \textbf{Temporal Dimension:} However, all these unfolding networks suffer from the challenges of slow inference and inability to learn temporal dimensional features. In addition, They focus solely on spatial structural features, neglecting the motion characteristics of small targets in the temporal dimension. To address these limitations, Zhu and Feng \cite{SFD-RPCAMatrix-71} design a bilevel-optimization RPCA guided spatio-temporal network (ESTNet). ESTNet is the first unfolding network that integrates CUR decomposition. More specifically, Zhu and Feng \cite{SFD-RPCAMatrix-71} formulate the infrared small target detection as a bilevel-optimized RPCA problem and solve it through algorithmic unfolding. To substantially increase the inference efficiency, ESTNet incorporates learnable CUR decompositions at each unfolding layer to replace the computationally intensive SVT operator. Moreover, Convolutional Long Short-Term Memory (CLSTM) modules are embedded within each unfolding layer to capture the multi-frame spatiotemporal motion characteristics of small targets. Experiments show that ESTNet outperforms  IPI  \cite{SFD-RPCAMatrix-1} and RPCANet \cite{SFD-RPCAMatrix-46}.
\end{itemize}

Even if these methods balance interpretability and learning capability, they suffer from two main weaknesses: the cascading accumulation of background estimation errors and the insufficient adaptability of fixed handcrafted sparse priors. To address these limitations in holistic manner, Wang et al. \cite{SFD-RPCAMatrix-95} design a deep unfolding robust PCA with Learnable Sparse Transform (LST-RPCANet). First, a Multi-Channel Attention Supervised Transmission-Enhancement Module (MCASTM) block error diffusion and purify cross-stage background features through an attention selection mechanism. Second, a Learnable Sparse Transformation Target Extraction Module (LSTEM) adaptively learns optimal sparse representations in the transform domain for target separation. Third, a Dynamic Attention-Guided Feature Enhancement Module (DAGFEM) enhances the image reconstruction fidelity. Experiments on SIRST V1, NUDT-SIRST, and IRSTD-1K datasets show LST-RPCANet outperforms data-based methods (DNANet \cite{DL-12}, UIU-Net \cite{DL-220}, L2SKNet \cite{DL-370}, SCTransNet \cite{DL-36}, MLCL-Net \cite{DL-599-77}, MSDA-Net \cite{DL-361}) and previous deep unfolding methods (RPCANet \cite{SFD-RPCAMatrix-46}, DRPCANet \cite{SFD-RPCAMatrix-64}) with lightweight parameters.  \\
Nevertheless, unfolded networks that retain an explicit singular value thresholding require an SVD of high-dimensional data at each layer, which significantly slows inference.

\subsection{Sparse Representation}
\label{LRSD-SR}
In 2014, Li et al. \cite{PSTD-45-1} design a sparse representation on a discriminative over-complete dictionary. First, an adaptive morphological over-complete dictionary is trained and built online according to the content of infrared image by K-singular value decomposition (K-SVD) algorithm. Second, an adaptive morphological over-complete dictionary is divided into a target over-complete dictionary describing target signals, and a background over-complete dictionary. This discriminative over-complete dictionary captures significant features of background clutter and dim targets better than a structural over-complete dictionary and also enhances the sparse feature difference between background and target more efficiently than a discriminative over-complete dictionary learned offline and classified manually. Third, the target and background clutter are sparsely decomposed over their corresponding over-complete dictionaries.  \\

In 2016, Wan et al. \cite{PSTD-45-2} propose a non-negativity constraint-based sparse representation. First, a frequency saliency detection based pre-processing extracts candidate regions that possibly contain the target. Second, a target over-complete dictionary is built by a varietal two-dimensional Gaussian model with an extent feature constraint and a background term. Third, a sparse representation model with a non-negativity constraint is employed for the candidate regions to obtain the corresponding coefficient vectors. The detection problem is converted to an $l_1$-regularized optimization using an Accelerated Proximal Gradient (APG) method. Finally, an evaluation index named Sparse Rate (SR) is employed to extract the real target by an adaptive segmentation directly. \\

In 2019, Lu et al. \cite{PSTD-45} propose a method based on sparse representation via an online-learned double sparse background dictionary. First, a double sparsity model is used to construct a background dictionary. The Kronecker product of two separable Discrete Cosine Transform (DCT) matrices allow to form the base dictionary. It is the first work based on a double sparsity model in the infrared small target detection problems. Second, a modified online learning algorithm is employed to train the dictionary. Thus, the sparse representation based decomposition process relies on an online-learned dictionary. The sparse representation model decompose the target image into background component, target component and noise component. Finally, an edge clutter suppression strategy improves the robustness of detection.  Experiments show that this sparse representation model outperforms Top-hat, ILCM \cite{SFD-LCM-5}, RLCM \cite{SFD-LCM-12}, LIGP \cite{SFD-LCM-43-2}, DECM  \cite{SFD-LCM-34} and SRvDOCD \cite{PSTD-45-1}. \\

\begin{table*} [ht!]
\begin{center}
\scalebox{0.75}{
\begin{tabular}{|l|l|l|l|l|l|l|}
\hline
\scriptsize{Methods}&\scriptsize{Decomposition} &\scriptsize{Minimization} &\scriptsize{Constraints}   &\scriptsize{Solver}  &\scriptsize{Convexity} 
&\scriptsize{Complexity}
\\
\hline
\hline
\scriptsize{\textbf{IPI (2013)}} 
&\scriptsize{$A=L+S$} 
&\scriptsize{$\underset{L,S}{\text{min}} ~~ ||L||_* + \lambda ||S||_{l_{1}}$}
&\scriptsize{$A-L-S=0$}
&\scriptsize{Inexact ALM \cite{SFD-RPCAMatrix-2}}
&\scriptsize{Yes}
&\scriptsize{$O_{iter}(mnr)$} \\ 
\scriptsize{Gao et al. \cite{SFD-RPCAMatrix-1}} 
&\scriptsize{} 
&\scriptsize{}
&\scriptsize{}
&\scriptsize{}
&\scriptsize{}
&\scriptsize{}\\
\hline
\scriptsize{\textbf{LRSR (2015)}} 
&\scriptsize{$A=AZ+DS+E$} 
&\scriptsize{$\underset{Z,S,E}{\text{min}} ~~ ||AZ||_* + \lambda ||E||_{l_{1}}+ \beta ||S||_{l_{1}}$}
&\scriptsize{$A-AZ-DS-E=0$}
&\scriptsize{Inexact ALM \cite{SFD-RPCAMatrix-2}}
&\scriptsize{Yes}
&\scriptsize{$O_{iter}(r^3 + nr^2 +mnr)$} \\
\scriptsize{He et al. \cite{SFD-RPCAMatrix-2}} 
&\scriptsize{} 
&\scriptsize{}
&\scriptsize{$D$ denotes the over-complete dictionary}
&\scriptsize{}
&\scriptsize{}
&\scriptsize{} \\
\hline 
\scriptsize{\textbf{T-BS (2015)}} 
&\scriptsize{$A=L+S+E$} 
&\scriptsize{$\underset{L,S,E}{\text{min}} ~~ ||L||_* + \lambda ||S+E||_{l_{1}}$}
&\scriptsize{$A-L-S-E=0$}
&\scriptsize{Landweber Algorithm \cite{SFD-RPCAMatrix-3}}
&\scriptsize{Yes}
&\scriptsize{$O_{iter}(r^3 + nr^2 +mnr)$} \\
\scriptsize{Wang and Qin \cite{SFD-RPCAMatrix-3}} 
&\scriptsize{} 
&\scriptsize{}
&\scriptsize{$S=S_{+}+S_{-}$, $E=E_{+}+E_{-}$}
&\scriptsize{}
&\scriptsize{}
&\scriptsize{} \\
\hline 
\scriptsize{\textbf{RST-CD (2017)}} 
&\scriptsize{$A=L+W \odot T$} 
&\scriptsize{$\underset{L,W}{\text{min}} ~~ ||L||_* + \lambda ||W \odot T||_{l_{1}}$}
&\scriptsize{$A-L-W \odot T=0$}
&\scriptsize{APG \cite{SFD-RPCAMatrix-6}}
&\scriptsize{Yes}
&\scriptsize{$O_{iter}(mn\min(m,n))$} \\
\scriptsize{Gao et al. \cite{SFD-RPCAMatrix-6}}
&\scriptsize{} 
&\scriptsize{}
&\scriptsize{$W$ is the weighting matrix.}
&\scriptsize{}
&\scriptsize{}
&\scriptsize{} \\
\scriptsize{}
&\scriptsize{} 
&\scriptsize{}
&\scriptsize{$T$ is the target matrix.}
&\scriptsize{}
&\scriptsize{}
&\scriptsize{} \\
\hline  
\scriptsize{\textbf{TV-PCP (2017)}} 
&\scriptsize{$A=L+S+E$} 
&\scriptsize{$\underset{L,S}{\text{min}} ~~ ||L||_{*} + \lambda_1 TV (L)+ ||S||_{l_{w,1}} +   \lambda_2 ||S||_{l_{1}}$}
&\scriptsize{$A-L-S-E=0$, $||E|| \leq \delta$}
&\scriptsize{Augmented Lagrangian \cite{SFD-RPCAMatrix-15}}
&\scriptsize{Yes}
&\scriptsize{$O_{iter}(mn\min(m,n))$} \\
\scriptsize{Wang et al. \cite{SFD-RPCAMatrix-15}} 
&\scriptsize{} 
&\scriptsize{$TV(\cdot)$ denotes the TV norm.}
&\scriptsize{}
&\scriptsize{}
&\scriptsize{}
&\scriptsize{} \\ 
\hline
\scriptsize{\textbf{WNM-IPI (2017)}} 
&\scriptsize{$A=L+S$} 
&\scriptsize{$\underset{L,S}{\text{min}} ~~ ||L||_{w,*} +  \lambda ||S||_{l_{1}}$}
&\scriptsize{$A-L-S=0$}
&\scriptsize{APG \cite{SFD-RPCAMatrix-19}}
&\scriptsize{No}
&\scriptsize{$O_{iter}(mn\min(m,n) + mn)$} \\
\scriptsize{Wang et al. \cite{SFD-RPCAMatrix-19}} 
&\scriptsize{} 
&\scriptsize{}
&\scriptsize{}
&\scriptsize{}
&\scriptsize{}
&\scriptsize{} \\ 
\hline
\scriptsize{\textbf{NIPPS (2017)}} 
&\scriptsize{$A=L+S$} 
&\scriptsize{$\underset{L,S}{\text{min}} ~~ ||L||_{*,\leq r} + \lambda ||S||_{l_1,\geq r}$}
&\scriptsize{$A-L-S=0$, $S>0$}
&\scriptsize{Augmented Lagrangian \cite{SFD-RPCAMatrix-5}}
&\scriptsize{Yes}
&\scriptsize{$O_{iter}(mn\min(m,n))$} \\
\scriptsize{Dai et al. \cite{SFD-RPCAMatrix-5}} 
&\scriptsize{} 
&\scriptsize{}
&\scriptsize{}
&\scriptsize{}
&\scriptsize{}
&\scriptsize{} \\
\hline
\scriptsize{\textbf{RPCA (2017)}} 
&\scriptsize{$A=L+S$} 
&\scriptsize{$\underset{L,S}{\text{min}} ~~ ||L||_{*} + \lambda ||S||_{l_1}$}
&\scriptsize{$A-L-S=0$}
&\scriptsize{Inexact Augmented Lagrangian \cite{SFD-RPCAMatrix-4}}
&\scriptsize{Yes}
&\scriptsize{$O_{iter}(mn\min(m,n) + mn)$} \\
\scriptsize{Fan et al. \cite{SFD-RPCAMatrix-4}} 
&\scriptsize{} 
&\scriptsize{}
&\scriptsize{}
&\scriptsize{}
&\scriptsize{}
&\scriptsize{} \\
\hline
\scriptsize{\textbf{NRAM (2018)}} 
&\scriptsize{$A=L+S+E$} 
&\scriptsize{$\underset{L,S,E}{\text{min}} ~~ ||L||_{\gamma} + \lambda ||S||_{l_{w,1}} +  \beta ||E||_{l_{2,1}}$}
&\scriptsize{$A-L-S-E=0$}
&\scriptsize{ADMM-DC \cite{SFD-RPCAMatrix-10}}
&\scriptsize{No}
&\scriptsize{$O_{iter}(mn^2)$} \\
\scriptsize{Zhang et al. \cite{SFD-RPCAMatrix-10}} 
&\scriptsize{} 
&\scriptsize{}
&\scriptsize{}
&\scriptsize{}
&\scriptsize{}
&\scriptsize{} \\ 
\hline
\scriptsize{\textbf{NOLC (2018)}} 
&\scriptsize{$A=L+S$} 
&\scriptsize{$\underset{L,S}{\text{min}} ~~ ||L||_{Sq} + \lambda ||S||_{l_{q}} $}
&\scriptsize{$||A-L-S||_F \leq \delta$}
&\scriptsize{ADMM \cite{SFD-RPCAMatrix-11}}
&\scriptsize{No}
&\scriptsize{$O_{iter}(mn\min(m,n))$} \\
\scriptsize{Zhang et al. \cite{SFD-RPCAMatrix-11}} 
&\scriptsize{} 
&\scriptsize{}
&\scriptsize{}
&\scriptsize{}
&\scriptsize{}
&\scriptsize{} \\
\hline
\scriptsize{\textbf{RMSV (2018)}} 
&\scriptsize{$A=L+S$} 
&\scriptsize{$\underset{L,S}{\text{min}} ~~ ||L||_{*,p} + \lambda ||S||_{l_{1},\geq 0} + tr(Y^T (A-L-S)) $}
&\scriptsize{$A-L-S=0$, $S \geq 0$}
&\scriptsize{IALM \cite{SFD-RPCAMatrix-20}}
&\scriptsize{No}
&\scriptsize{$O_{iter}(mn\min(m,n) + mn)$} \\
\scriptsize{Yuan and Dong \cite{SFD-RPCAMatrix-20}} 
&\scriptsize{} 
&\scriptsize{}
&\scriptsize{}
&\scriptsize{}
&\scriptsize{}
&\scriptsize{} \\
\hline
\scriptsize{\textbf{GRLA (2019)}} 
&\scriptsize{$A=L+S$} 
&\scriptsize{$\underset{L,S}{\text{min}} ~~ ||L||_{*} + \lambda_1 ||W_S \odot S||_{l_{1}} +\lambda_2 Tr(ZMZ^T)$}
&\scriptsize{$A-L-S=0$, $L=Z$}
&\scriptsize{ADMM \cite{SFD-RPCAMatrix-13}}
&\scriptsize{No}
&\scriptsize{$O_{iter}(mn^2)$} \\
\scriptsize{Zhou et al. \cite{SFD-RPCAMatrix-13}}
&\scriptsize{} 
&\scriptsize{$W_S$ is a weighted matrix.}
&\scriptsize{}
&\scriptsize{}
&\scriptsize{}
&\scriptsize{} \\
\scriptsize{}
&\scriptsize{} 
&\scriptsize{$M$ is the graph Laplacian matrix.}
&\scriptsize{}
&\scriptsize{}
&\scriptsize{}
&\scriptsize{} \\
\hline
\scriptsize{\textbf{G-GRLA (2019)}} 
&\scriptsize{$A=L+S$} 
&\scriptsize{$\underset{L,S}{\text{min}} ~~ ||L||_{NC\text{-}Norm} + \lambda_1 ||W_S \odot S||_{l_{1}} +\lambda_2 Tr(ZMZ^T)$}
&\scriptsize{$A-L-S=0$, $L=Z$}
&\scriptsize{ADMM \cite{SFD-RPCAMatrix-13}}
&\scriptsize{No}
&\scriptsize{$O_{iter}(mn^2)$} \\
\scriptsize{Zhou et al. \cite{SFD-RPCAMatrix-13}}
&\scriptsize{} 
&\scriptsize{$W_S$ is a weighted matrix.}
&\scriptsize{}
&\scriptsize{}
&\scriptsize{}
&\scriptsize{} \\
\scriptsize{}
&\scriptsize{} 
&\scriptsize{$M$ is the graph Laplacian matrix. NC-Norm is a non-convex norm.}
&\scriptsize{}
&\scriptsize{}
&\scriptsize{}
&\scriptsize{} \\
\hline
\scriptsize{\textbf{DGRLA (2020)}} 
&\scriptsize{$A=L+S$} 
&\scriptsize{$\underset{L,S}{\text{min}} ~~ ||L||_{*} + \lambda ||S||_{l_{1/2>0}}^{1/2} +\gamma_1 Tr(ST_PS^T) + \gamma_2 Tr(ST_FS^T)$}
&\scriptsize{$||A-L-S||_F=0$, $S \geq 0$}
&\scriptsize{LADMAP \cite{SFD-RPCAMatrix-14}}
&\scriptsize{No}
&\scriptsize{$O_{iter}(mn\max(m,n))$} \\
\scriptsize{Zhou et al. \cite{SFD-RPCAMatrix-14}}
&\scriptsize{} 
&\scriptsize{$T_P$ Patch graph Laplacian matrix.}
&\scriptsize{}
&\scriptsize{}
&\scriptsize{}
&\scriptsize{} \\
\scriptsize{}
&\scriptsize{} 
&\scriptsize{$T_F$ Feature graph Laplacian matrix.}
&\scriptsize{}
&\scriptsize{}
&\scriptsize{}
&\scriptsize{} \\
\hline
\end{tabular}}
\caption{Low-rank plus Sparse Matrix Decomposition for Infrared Small Target Detection: An homogeneous overview (Part 1). $m \times n$
is the size of the observation matrix, and $r$ is the rank.}
\label{Overview-1}
\end{center}
\end{table*}

\begin{table*}
\begin{center}
\scalebox{0.75}{
\begin{tabular}{|l|l|l|l|l|l|l|}
\hline
\scriptsize{Methods}&\scriptsize{Decomposition} &\scriptsize{Minimization} &\scriptsize{Constraints}   &\scriptsize{Solver}  &\scriptsize{Convexity} 
&\scriptsize{Complexity}
\\
\hline
\hline
\scriptsize{\textbf{MoG-MRF (2018)}} 
&\scriptsize{$A=L+S$} 
&\scriptsize{$\underset{L,S}{\text{min}} ~~ ||L||_* + \lambda ||S||_{l_{1}}$}
&\scriptsize{$A-L-S=0$}
&\scriptsize{Variational Bayesian \cite{MOD-Method-170}}
&\scriptsize{No}
&\scriptsize{$O_{iter}((m+n)r^3+kmnr)$} \\ 
\scriptsize{Gao et al. \cite{SFD-RPCAMatrix-17}} 
&\scriptsize{} 
&\scriptsize{}
&\scriptsize{}
&\scriptsize{}
&\scriptsize{}
&\scriptsize{}\\
\hline
\scriptsize{\textbf{R$S_{1/2}$NIPI (2019)}} 
&\scriptsize{$A=L+S+E$} 
&\scriptsize{$\underset{L,S}{\text{min}} ~~ ||L||_{S_{1/2}}^{1/2} + \lambda ||S||_{l_{1,W_E}}$}
&\scriptsize{$A-L-S-E=0$, $||E|| \leq \eta$}
&\scriptsize{ADMM \cite{SFD-RPCAMatrix-16}}
&\scriptsize{No}
&\scriptsize{$O_{iter}(mn + mn\min(m,n) + mnr + mr)$} \\
\scriptsize{Zhou et al. \cite{SFD-RPCAMatrix-16}}
&\scriptsize{} 
&\scriptsize{$W_E=\{w_{E,ij}\}$: Weights for every entry in the target patch-image matrix}
&\scriptsize{}
&\scriptsize{}
&\scriptsize{}
&\scriptsize{} \\
\hline
\scriptsize{\textbf{MW-IPI (2019)}} 
&\scriptsize{$A=L+S$} 
&\scriptsize{$\underset{L,S}{\text{min}} ~~ ||L||_{*} + \lambda ||S||_{l_{1}}$}
&\scriptsize{$A-L-S=0$}
&\scriptsize{ADMM \cite{SFD-RPCAMatrix-21}}
&\scriptsize{No}
&\scriptsize{$O_{iter}(mn + mn\min(m,n) + mnr + mr)$} \\
\scriptsize{Li et al. \cite{SFD-RPCAMatrix-21}}
&\scriptsize{} 
&\scriptsize{}
&\scriptsize{}
&\scriptsize{}
&\scriptsize{}
&\scriptsize{} \\
\hline
\scriptsize{\textbf{GreB (2020)}} 
&\scriptsize{$A=L+S+E$} 
&\scriptsize{$\underset{X,Y,S}{\text{min}} ~~ ||A-XY-S||_* + \lambda ||S||_{l_{1}}$}
&\scriptsize{$rank(X)=rank(Y) \leq r$}
&\scriptsize{GreB \cite{MOD-Method-180}}
&\scriptsize{No}
&\scriptsize{$O_{iter}(mnr^2+3mnr+mr)$} \\ 
\scriptsize{Pang et al. \cite{SFD-RPCAMatrix-18}} 
&\scriptsize{} 
&\scriptsize{}
&\scriptsize{}
&\scriptsize{}
&\scriptsize{}
&\scriptsize{}\\
\hline
\scriptsize{\textbf{TNLRS (2020)}} 
&\scriptsize{$A=L+S+E$}
&\scriptsize{$\underset{L,S,\{Z_i\}}{\text{min}} ~~ \eta\sum_i \big(||\tilde{R}_iL - Z_i||_F^2 + \lambda\log\det((Z_iZ_i^T)^{1/2}+\mu I)\big)$}
&\scriptsize{$||A-L-S||_F^2 \leq \zeta^2$}
&\scriptsize{ADMM \cite{SFD-RPCAMatrix-30}}
&\scriptsize{No}
&\scriptsize{$O_{iter}(Kpq\min(p,q) + Kpq + mn)$} \\ 
\scriptsize{Zhu et al. \cite{SFD-RPCAMatrix-30}}  
&\scriptsize{} 
&\scriptsize{$+ \alpha||S\odot\log S||_{l_{1}} + \beta||A-L-S||_F^2$}
&\scriptsize{}
&\scriptsize{}
&\scriptsize{}
&\scriptsize{}\\
\scriptsize{}  
&\scriptsize{} 
&\scriptsize{$K$ non-local patch groups of size $p \times q$.}
&\scriptsize{}
&\scriptsize{}
&\scriptsize{}
&\scriptsize{}\\
\hline
\scriptsize{\textbf{SCPLBMA (2021)}} 
&\scriptsize{$A=L+S$} 
&\scriptsize{$\underset{L,S}{\text{min}} ~~ ||L||_* + \lambda ||S||_{l_{1}}$}
&\scriptsize{$A-L-S=0$}
&\scriptsize{ALM \cite{SFD-RPCAMatrix-12}}
&\scriptsize{Yes}
&\scriptsize{$O_{iter}(mn\min(m,n) + mn)$} \\
\scriptsize{Li et al. \cite{SFD-RPCAMatrix-12}}  
&\scriptsize{} 
&\scriptsize{}
&\scriptsize{}
&\scriptsize{}
&\scriptsize{}
&\scriptsize{}\\
\hline
\scriptsize{\textbf{IPI with PUOT (2021)}} 
&\scriptsize{$A=L+S$} 
&\scriptsize{$\underset{L,S}{\text{min}} ~~ \frac{1}{2}||A-L-S||_F^2 + \lambda_1 ||S||_{l_{1}} + \lambda_2 ||L||_*$}
&\scriptsize{$L \geq 0$, $S \geq 0$}
&\scriptsize{ADMM \cite{SFD-RPCAMatrix-33}}
&\scriptsize{Yes}
&\scriptsize{$O_{iter}(mn\min(m,n) + mn)$}\\
\scriptsize{Bertrand et al. \cite{SFD-RPCAMatrix-33}} 
&\scriptsize{} 
&\scriptsize{}
&\scriptsize{}
&\scriptsize{}
&\scriptsize{}
&\scriptsize{}\\
\hline
\scriptsize{\textbf{TV-IIPI (2022)}} 
&\scriptsize{$A=L+S+E$} 
&\scriptsize{$\underset{L,S}{\text{min}} ~~ \frac{1}{2}||A-L-S||_F^2 + \lambda ||L||_* + \alpha ||S||_{l_{1}}$}
&\scriptsize{None}
&\scriptsize{APG \cite{SFD-RPCAMatrix-35}}
&\scriptsize{Yes}
&\scriptsize{$O_{iter}(mn\min(m,n) + mn)$} \\
\scriptsize{Wan et al. \cite{SFD-RPCAMatrix-35}}
&\scriptsize{} 
&\scriptsize{$+ \beta ||S-S_0||_F^2 + \gamma \, TV(S)$}
&\scriptsize{}
&\scriptsize{}
&\scriptsize{}
&\scriptsize{} \\
\scriptsize{}
&\scriptsize{} 
&\scriptsize{$S_0$ is the target matrix of the previous frame.}
&\scriptsize{}
&\scriptsize{}
&\scriptsize{}
&\scriptsize{} \\
\hline
\scriptsize{\textbf{KRPCA (2022)}} 
&\scriptsize{$A=L+S$} 
&\scriptsize{$\underset{L,S}{\text{min}} ~~ \psi(L) + \lambda_1 \Uppsi(S)+ \beta_1 f(L) + \lambda_2 f(S)$}
&\scriptsize{$A-L-S=0$}
&\scriptsize{ADMM \cite{SFD-RPCAMatrix-37}}
&\scriptsize{No}
&\scriptsize{$O_{iter}(m^2n + mnk + mn)$} \\
\scriptsize{Yan et al. \cite{SFD-RPCAMatrix-37}}
&\scriptsize{} 
&\scriptsize{}
&\scriptsize{}
&\scriptsize{}
&\scriptsize{}
&\scriptsize{} \\
\hline
\scriptsize{\textbf{SRLCM (2022)}} 
&\scriptsize{$A=L+S$} 
&\scriptsize{$\underset{L,S}{\text{min}} ~~ ||L||_{\Phi,\sigma} + \lambda ||S||_{\Phi,l_{0}}$}
&\scriptsize{$A-L-S=0$} 
&\scriptsize{ADMM \cite{SFD-RPCAMatrix-77}}
&\scriptsize{No}
&\scriptsize{$O_{iter}(mn\min(m,n) + mn)$} \\
\scriptsize{Li et al. \cite{SFD-RPCAMatrix-77}}
&\scriptsize{} 
&\scriptsize{}
&\scriptsize{}
&\scriptsize{}
&\scriptsize{}
&\scriptsize{} \\
\hline
\end{tabular}}
\caption{Low-rank plus Sparse Matrix Decomposition for Infrared Small Target Detection: An homogeneous overview (Part 2). $m \times n$ is the size of the observation matrix, and $r$ is the rank.}
\label{Overview-2}
\end{center}
\end{table*}

\begin{table*}
\begin{center}
\scalebox{0.75}{
\begin{tabular}{|l|l|l|l|l|l|l|}
\hline
\scriptsize{Methods}&\scriptsize{Decomposition} &\scriptsize{Minimization} &\scriptsize{Constraints}   &\scriptsize{Solver}  &\scriptsize{Convexity} 
&\scriptsize{Complexity}
\\
\hline
\hline
\scriptsize{\textbf{HiLV-LRSD (2023)}} 
&\scriptsize{$A=L+S$} 
&\scriptsize{$\underset{L,S}{\text{min}} ~~ ||L||_{LV*} + \lambda_1 ||S||_{l_v,l_1}$} 
&\scriptsize{$A-L-S=0$} 
&\scriptsize{PI-ADMM \cite{SFD-RPCAMatrix-38}}
&\scriptsize{No}
&\scriptsize{$O_{iter}(mn\min(m,n) + mnr + mn + \min(m,n))$}\\
\scriptsize{Liu et al. \cite{SFD-RPCAMatrix-38}}
&\scriptsize{} 
&\scriptsize{}
&\scriptsize{}
&\scriptsize{}
&\scriptsize{}
&\scriptsize{} \\
\hline
\scriptsize{\textbf{LMRPCA (2023)}} 
&\scriptsize{$A=L+S$} 
&\scriptsize{$\underset{L,S}{\text{min}} ~~ ||L||_{*} + \lambda_1 \Sigma_i g_i ||S_i - \tilde{S_i}||_{l_1} + \lambda_2 \Sigma_i g_i ||S_i||_{l_1}$} 
&\scriptsize{$(1-S) \odot (A-L)=0$} 
&\scriptsize{ADMM \cite{SFD-RPCAMatrix-39}}
&\scriptsize{No}
&\scriptsize{$O_{iter}(mn\min(m,n) + mn)$}  \\
\scriptsize{Zhou et al. \cite{SFD-RPCAMatrix-39}}
&\scriptsize{} 
&\scriptsize{}
&\scriptsize{}
&\scriptsize{}
&\scriptsize{}
&\scriptsize{} \\
\hline
\scriptsize{\textbf{RCTVW (2023)}} 
&\scriptsize{$A=L+S$} 
&\scriptsize{$\underset{L,S}{\text{min}} ~~ ||L||_{RCTV} + \lambda_1 ||S||_{w,l_{1}}$}
&\scriptsize{$A-L-S=0$} 
&\scriptsize{ADMM \cite{SFD-RPCAMatrix-40}}
&\scriptsize{No}
&\scriptsize{$O_{iter}(mr + mn + nr\min(n,r))$} \\
\scriptsize{Liu et al. \cite{SFD-RPCAMatrix-40}}
&\scriptsize{} 
&\scriptsize{}
&\scriptsize{}
&\scriptsize{}
&\scriptsize{}
&\scriptsize{} \\
\hline
\scriptsize{\textbf{$l_1$--$l_2$ norm-TV (2023)}} 
&\scriptsize{$A=L+S$} 
&\scriptsize{$\underset{L,S}{\text{min}} ~~ ||L||_* + \lambda_1 ||S||_{l_{1},l_{2}} + \lambda_2 TV(F)$}
&\scriptsize{$A-L-S=0$, $S=F$}
&\scriptsize{ADMM \cite{SFD-RPCAMatrix-41}}
&\scriptsize{No}
&\scriptsize{$O_{iter}(mnr^2)$} \\
\scriptsize{Shao et al. \cite{SFD-RPCAMatrix-41}}
&\scriptsize{} 
&\scriptsize{}
&\scriptsize{}
&\scriptsize{}
&\scriptsize{}
&\scriptsize{} \\
\hline
\scriptsize{\textbf{Bi-NNM (2023)}} 
&\scriptsize{$A=L+S$} 
&\scriptsize{$\underset{L,S}{\text{min}} ~~ ||L||_{BiN} + \lambda_1 ||S||_{w,l_{1}}$}
&\scriptsize{$A-L-S=0$}
&\scriptsize{ADMM \cite{SFD-RPCAMatrix-42}}
&\scriptsize{No}
&\scriptsize{$O_{iter}(mnr + mr^2 + nr^2 + mn)$} \\
\scriptsize{Rawat et al. \cite{SFD-RPCAMatrix-42}}
&\scriptsize{} 
&\scriptsize{}
&\scriptsize{}
&\scriptsize{}
&\scriptsize{}
&\scriptsize{} \\
\hline
\scriptsize{\textbf{NRAM-MRW (2023)}} 
&\scriptsize{$A=L+S+E$} 
&\scriptsize{$\underset{L,S,E}{\text{min}} ~~ ||L||_\gamma + \lambda ||S||_{w,l_{2}} + \beta ||E||_{l_{2,1}}$}
&\scriptsize{$A-L-S-E=0$}
&\scriptsize{ADMM \cite{SFD-RPCAMatrix-43}}
&\scriptsize{No}
&\scriptsize{$O_{iter}(mn\min(m,n) + mn)$} \\
\scriptsize{Wang et al. \cite{SFD-RPCAMatrix-43}}
&\scriptsize{} 
&\scriptsize{}
&\scriptsize{}
&\scriptsize{}
&\scriptsize{}
&\scriptsize{} \\
\hline
\scriptsize{\textbf{GPCP (2024)}} 
&\scriptsize{$A=L+S+E$} 
&\scriptsize{$\underset{L,S}{\text{min}} ~~ \Sigma_k \mu_k ||L_k||_* + \lambda_1 ||S||_{l_{1}}$}
&\scriptsize{$A-L-S-E=0$}
&\scriptsize{ADMM \cite{SFD-RPCAMatrix-44}}
&\scriptsize{Yes}
&\scriptsize{$O_{iter}(mnr^3)$} \\ 
\scriptsize{Li et al. \cite{SFD-RPCAMatrix-44}}
&\scriptsize{} 
&\scriptsize{$L_k$: $k$-th group of the background.}
&\scriptsize{}
&\scriptsize{}
&\scriptsize{}
&\scriptsize{} \\
\hline
\scriptsize{\textbf{DW-SSD (2024)}} 
&\scriptsize{$A=L+S+E$} 
&\scriptsize{$\underset{\Theta,S,E}{\text{min}} ~~ ||\Theta||_{TV} + \lambda_1 ||W \odot S||_{sl_{0}} + \beta ||E||_{l_{2,1}}$}
&\scriptsize{$A-L-S-E=0$} 
&\scriptsize{ADMM \cite{SFD-RPCAMatrix-48}}
&\scriptsize{No}
&\scriptsize{$O_{iter}(mnr + mr^2 + nr^2 + r^3 + mn)$} \\
\scriptsize{Dan et al. \cite{SFD-RPCAMatrix-48}}
&\scriptsize{} 
&\scriptsize{$L=S_H\Theta S_V^T$, $||\cdot||_{sl_{0}}$: smoothed $l_0$-norm.}
&\scriptsize{}
&\scriptsize{}
&\scriptsize{}
&\scriptsize{} \\
\hline
\scriptsize{\textbf{WIPI-WNNM (2024)}} 
&\scriptsize{$A=L+S+E$} 
&\scriptsize{$\underset{L,S}{\text{min}} ~~ ||L||_{w,*} + \lambda_1 ||S||_{l_{1}}$} 
&\scriptsize{$A-L-S-E=0$}
&\scriptsize{ADMM \cite{SFD-RPCAMatrix-49}}
&\scriptsize{No}
&\scriptsize{$O_{iter}(mn\min(m,n) + mn)$} \\
\scriptsize{Chen et al. \cite{SFD-RPCAMatrix-49}}
&\scriptsize{} 
&\scriptsize{$||\cdot||_{w,*}$: weighted nuclear norm.}
&\scriptsize{}
&\scriptsize{}
&\scriptsize{}
&\scriptsize{} \\
\hline
\scriptsize{\textbf{SBL (2025)}} 
&\scriptsize{$Y=\Phi(L+S)+N$} 
&\scriptsize{$\underset{L,S}{\text{min}} ~~ \Sigma_k \frac{1}{2\lambda}||Y_k-\Phi(S_k+L)||_F^2 + tr(S_k^H\Gamma_k^{-1}S_k) + tr(L^H\Gamma_L^{-1}L)$}
&\scriptsize{None}
&\scriptsize{VB-EM \cite{SFD-RPCAMatrix-60}}
&\scriptsize{No}
&\scriptsize{$O_{iter}(Kr(m+r)(r+n))$} \\
\scriptsize{Guo et al. \cite{SFD-RPCAMatrix-60}}
&\scriptsize{} 
&\scriptsize{$\Phi$: measurement operator, $N$: noise, $K$: number of frames.}
&\scriptsize{}
&\scriptsize{}
&\scriptsize{}
&\scriptsize{} \\
\hline
\scriptsize{\textbf{TV Regularized (DoLP images) (2025)}} 
&\scriptsize{$A=L+S+E$} 
&\scriptsize{$\underset{L,S,E}{\text{min}} ~~ ||L||_{*} + \lambda_1 ||S+E||_{l_{1}} + \lambda_2 ||E||_{l_{1}} + \lambda_3 ||\Omega(S)||_{TV}$}
&\scriptsize{$A-L-S-E=0$} 
&\scriptsize{ADMM \cite{SFD-RPCAMatrix-68}}
&\scriptsize{No}
&\scriptsize{$O_{iter}(mn\min(m,n) + mn)$} \\
\scriptsize{Hu et al. \cite{SFD-RPCAMatrix-68}}
&\scriptsize{} 
&\scriptsize{$||\cdot||_{TV}$: TV regularization.}
&\scriptsize{}
&\scriptsize{}
&\scriptsize{}
&\scriptsize{} \\
\hline
\scriptsize{\textbf{ITPM-HiLV-LRSD (2026)}} 
&\scriptsize{$A=L+S+E$} 
&\scriptsize{$\underset{L,S}{\text{min}} ~~ ||L||_{LV*} + \lambda_1 ||S||_{w,l_{1}}$}
&\scriptsize{$||A-L-S||_F \leq \delta$} 
&\scriptsize{ADMM \cite{SFD-RPCAMatrix-73}}
&\scriptsize{No}
&\scriptsize{$O_{iter}(mn\min(m,n) + mn)$} \\
\scriptsize{Liu et al. \cite{SFD-RPCAMatrix-73}}
&\scriptsize{} 
&\scriptsize{}
&\scriptsize{}
&\scriptsize{}
&\scriptsize{}
&\scriptsize{} \\
\hline
\end{tabular}}
\caption{Low-rank plus Sparse Matrix Decomposition for Infrared Small Target Detection: An homogeneous overview (Part 3). $m \times n$
is the size of the observation matrix, and $r$ is the rank.}
\label{Overview-3}
\end{center}
\end{table*}

\subsection{Multi-Frame Detection}
\label{LRSD-MFD}
Even if single-frame methods are relatively efficient, they do not take into account the temporal information that are suitable to detect small target features. Different spatiotemporal detection methods have been developed in literature, with spatiotemporal tensor optimization models being particularly notable. The reader can refer to the DLAT website\footnote{https://sites.google.com/site/robustdlat/} for a list of representative fundamental papers about diverse robust tensor decompositions.

\subsubsection{Mathematical Preliminaries}
\begin{figure*}[!t]
\begin{center}
\includegraphics[width=12cm]{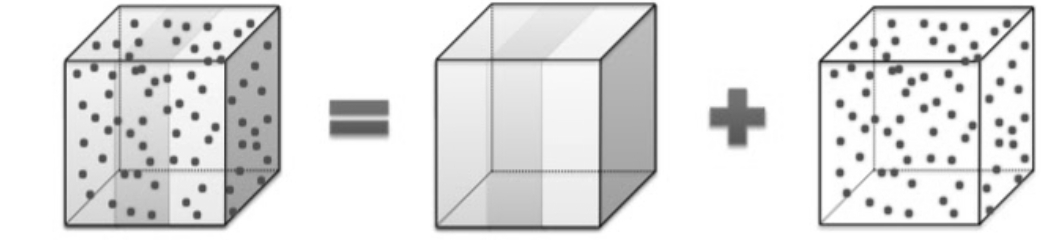}
\caption{Illustration of the decomposition of an observed tensor $\mathcal{A}$ into a low rank tensor $\mathcal{L}$ and a sparse tensor $\mathcal{S}$.} 
\label{RPCATensorF}
\end{center}
\end{figure*}

The  robust PCA problem formulated in a matrix manner \cite{MOD-Theory-1} is expressed in a tensor manner as follows \cite{MOD-Survey-4-10}: 
\begin{equation}
\mathcal{A} = \mathcal{L} + \mathcal{S}
\label{EquationPCP10}
\end{equation}
where $\mathcal{L}$ is a low-rank tensor and $\mathcal{S}$ must be sparse tensor with a small fraction of nonzero entries. Note that the decomposition problem is more harder than in the matrix case are there are several ways to manage tensor as CP decomposition \cite{MOD-Survey-4-11}, Tücker decomposition \cite{MOD-Survey-4-12}, tubal decomposition \cite{MOD-Survey-4-13}, CUR decomposition \cite{MOD-Survey-4-14}, BM decomposition \cite{MOD-Survey-4-15} and ring decomposition \cite{MOD-Survey-4-16}. Otherwise, the characteristics are the same than in the matrix case in terms of decomposition in two or three tensors, loss functions and solvers. Figure \ref{RPCATensorF} shows an illustration of the decomposition of an observed tensor $\mathcal{A}$ into a low rank tensor $\mathcal{L}$ and a sparse tensor $\mathcal{S}$.  \\

\subsubsection{RPCA Tensor based Decomposition}
\label{RPCATensor}
In 2019, Zhang et al. \cite{SFD-RPCATensor-9} propose an Image-Patch Tensor (IPT) model. Based on the structural relationship between pixels, the original single-frame image is stacked as a new image-patch tensor. Based on the correlation of the background image patch and the sparsity of the target image patch, the small target detection problem is formulated into an optimization problem of separating the low-rank part and the sparse part of the tensor. After post filtering, the target is separated adaptively. The IPT model is thus the transformation of the IPI model  \cite{SFD-RPCAMatrix-1} published in 2013 from matrix up to tensor six years later. In practice, the IPT model presents several weaknesses addressed by further works regarding the decomposition, the loss functions, the solver and the computation time.  These improvements can be categorized as done in the following items and Table \ref{OverviewTensor} shows an overview classification of the diverse existing LRSD tensor methods applied to infrared small target detection.

\begin{table*}
\begin{center}
\scalebox{0.70}{
\begin{tabular}{|l|l|l|}
\hline
\scriptsize{Category}&\scriptsize{Methods} &\scriptsize{Acronyms} 
\\
\hline
\hline
\scriptsize{RPCA Tensors Decomposition} &\scriptsize{} &\scriptsize{}  \\
\cline{2-3}
\scriptsize{} &\scriptsize{Background} &\scriptsize{Improved IPI (Deng et al. \cite{SFD-RPCAMatrix-69}), ASTTV (Liu et al. \cite{SFD-RPCATensor-32}), STFF [Tubal] (Lu et al. \cite{SFD-RPCATensor-69}) } \\
\cline{2-3}
\scriptsize{} &\scriptsize{Target} &\scriptsize{ReIPT (Dai and  Wu \cite{SFD-RPCATensor-3}), ReWIPI (Guo et al. \cite{SFD-RPCATensor-4}), WSNM-STIPT (Sun et al. \cite{SFD-RPCATensor-11})} \\
\scriptsize{} &\scriptsize{} &\scriptsize{Facet-LRSD (Dong et al. \cite{SFD-RPCATensor-12}),  FDMDEA-STT (Pang et al. \cite{SFD-RPCATensor-13}), STC-LRSD (Fan et al. \cite{SFD-RPCAMatrix-76})} \\
\scriptsize{} &\scriptsize{} &\scriptsize{STTM-SFR (Pang et al. \cite{SFD-RPCATensor-14}), ReIPT-LPM (Li et al. \cite{SFD-RPCATensor-15})} \\
\scriptsize{} &\scriptsize{} &\scriptsize{WE [Entropy] (Liu and  Wang \cite{SFD-RPCATensor-41}), STBMPT (Aliha et al.  \cite{SFD-RPCATensor-42}), Twist Tensor (Li et al.  \cite{SFD-RPCATensor-45}) } \\
\scriptsize{} &\scriptsize{} &\scriptsize{CEW (Zhao et al.  \cite{SFD-RPCATensor-60}), IWS  (Zhang et al.  \cite{SFD-RPCATensor-73}), LRTA-SP (Pang et al.  \cite{SFD-RPCATensor-74})}\\
\scriptsize{} &\scriptsize{} &\scriptsize{EI-LRSD (Zhu et al.  \cite{SFD-RPCATensor-90}), APSTD (Luo et al.  \cite{SFD-RPCATensor-112})}\\
\scriptsize{} &\scriptsize{} &\scriptsize{SDD  (Zhou et al. \cite{SFD-RPCATensor-117})} \\
\cline{2-3}
\scriptsize{} &\scriptsize{Background/Target} &\scriptsize{PSTNN [Single-Frame] (Zhang and Peng \cite{SFD-RPCATensor-5}), NTRS (Guan et al. \cite{SFD-RPCATensor-6}), WNRIPT (Sun et al. \cite{SFD-RPCATensor-7}), STTVWNIPT (Sun et al. \cite{SFD-RPCATensor-10})} \\
\scriptsize{} &\scriptsize{} &\scriptsize{ECA-STT [Multi-Frame] (Zhang et al. \cite{SFD-RPCATensor-16}), STTM (Liu et al. \cite{SFD-RPCATensor-17}), TR [Ring] (Deng et al. \cite{SFD-RPCATensor-18}), STT-TRNR [Ring] (Yi et al. \cite{SFD-RPCATensor-35}), 4D-TR [Ring] (Wu et al. \cite{SFD-RPCATensor-36}) } \\
\scriptsize{} &\scriptsize{} &\scriptsize{TCTD [Tucker] (Zhao et al. \cite{SFD-RPCATensor-19}), LogTFNN (NTFRA) [Single-Frame] (Kong et al. \cite{SFD-RPCATensor-20}), ITRM (Zhou et al. \cite{SFD-RPCATensor-21})} \\
\scriptsize{} &\scriptsize{} &\scriptsize{METTR [Single-frame] (Cao et al. \cite{SFD-RPCATensor-30}), NPSTT (Wang et al. \cite{SFD-RPCATensor-31}), IMNN-LWEC [Multi-Frame] (Luo et al. \cite{SFD-RPCATensor-34})} \\
\scriptsize{} &\scriptsize{} &\scriptsize{TSC $p$N (Yan et al. \cite{SFD-RPCATensor-37}), STRL-LBCM (Luo et al. \cite{SFD-RPCATensor-38}), TD-FP [Tucker] (Liu et al. \cite{SFD-RPCATensor-39})} \\
\scriptsize{} &\scriptsize{} &\scriptsize{MNSTLA (Wang et al. \cite{SFD-RPCATensor-44}), TLNMTQR (Zhao and Xu \cite{SFD-RPCATensor-62}), TTNN [Train] (Wu et al.  \cite{SFD-RPCATensor-63})} \\
\scriptsize{} &\scriptsize{} &\scriptsize{FST-FLNN (Luo et al.  \cite{SFD-RPCATensor-64}), 4DST-BTMD (Luo et al.  \cite{SFD-RPCATensor-65}), SEoST (Luo et al.  \cite{SFD-RPCATensor-67})} \\
\scriptsize{} &\scriptsize{} &\scriptsize{DNLCDRW (Sun et al.  \cite{SFD-RPCATensor-68}), ASTIT-WPTAR (Sun et al.  \cite{SFD-RPCATensor-70}), 5D-STFC (Luo et al.  \cite{SFD-RPCATensor-71})} \\
\scriptsize{} &\scriptsize{} &\scriptsize{HGLAPR (Luo et al.  \cite{SFD-RPCATensor-72}), FGLR (Liu et al.  \cite{SFD-RPCATensor-75}), Lp-MCTV (Chen et al.  \cite{SFD-RPCATensor-99})} \\
\scriptsize{} &\scriptsize{} &\scriptsize{FRR (Li and Chen  \cite{SFD-RPCATensor-100}, GC (Ge et al. \cite{SFD-RPCATensor-102}, EASTIT-NWTAR (Sun et al.  \cite{SFD-RPCATensor-103}) } \\
\scriptsize{} &\scriptsize{} &\scriptsize{
GLLS-LMAG (Li et al. \cite{SFD-RPCAMatrix-62}), MSR-FT (Li et al. \cite{SFD-RPCAMatrix-75}), DMS-OGSTV (Ge et al. \cite{SFD-RPCATensor-93}) } \\
\scriptsize{} &\scriptsize{} &\scriptsize{TP-STTA (Zhao et al. \cite{SFD-RPCATensor-106}), WTANN-$l_p$ (Yang et al. \cite{SFD-RPCATensor-113}), KTNN  (Ge et al. \cite{SFD-RPCATensor-114})} \\ 
\scriptsize{} &\scriptsize{} &\scriptsize{ILN-SSR (Liu et al. \cite{SFD-RPCAMatrix-99}), LSTT (Lu et al. \cite{SFD-RPCATensor-116}), GFT-TNN (Wu et al. \cite{SFD-RPCATensor-120})} \\
\scriptsize{} &\scriptsize{} &\scriptsize{SH-FSTT (Zhu et al. \cite{SFD-RPCATensor-131}), SGSC (Hong et al. \cite{SFD-RPCATensor-132}), SHFCTM (Zhu et al. \cite{SFD-RPCATensor-133})} \\
\scriptsize{} &\scriptsize{} &\scriptsize{HMRT (Wu et al. \cite{SFD-RPCATensor-134}) , ANTLS (Chen et al. \cite{SFD-RPCATensor-135}), INR (Liu et al. \cite{SFD-RPCATensor-136})} \\
\cline{2-3}
\scriptsize{} &\scriptsize{Heterogeneous Backgrounds}  &\scriptsize{Adaptivity: ANLPT (Zhang et al. \cite{SFD-RPCATensor-40}), ATN (Nie et al. \cite{SFD-RPCATensor-43})} \\
\scriptsize{} &\scriptsize{}  &\scriptsize{Clustering: CTSTC (Luo et al. \cite{PSTD-41})  PTCTV-KMC (Huang et al. \cite{PSTD-43}) } \\
\cline{2-3}
\scriptsize{} &\scriptsize{Computation Time}  &\scriptsize{
BTR-ISTD (Li \cite{SFD-RPCATensor-95}), IPCE (Zhang \cite{SFD-RPCATensor-110})  } \\
\cline{2-3}
\scriptsize{} &\scriptsize{Bayesian Decomposition} &\scriptsize{(Tan et al. \cite{SFD-RPCATensor-111})} \\
\cline{2-3}
\scriptsize{} &\scriptsize{Deep Unfold Tensor Decomposition}  
&\scriptsize{DNN-aided LRSD (Yin et al. \cite{SFD-RPCAMatrix-72}), TenRPCANet (Zhang et al. \cite{SFD-RPCATensor-94}), TTDUNet (Xiao and Hu \cite{SFD-RPCATensor-96}) } \\
\scriptsize{} &\scriptsize{}  &\scriptsize{
NeurSTT (Wu et al. \cite{SFD-RPCATensor-97}), HLTL-Net (Qi et al. \cite{SFD-RPCATensor-98}), DRTENet (Ma et al. \cite{SFD-RPCATensor-104})} \\
\scriptsize{} &\scriptsize{}  &\scriptsize{WSWTNN-PnP (Liu et al. \cite{SFD-RPCATensor-105})} \\
\scriptsize{} &\scriptsize{}  &\scriptsize{ST-UBRNet (Ma et al. \cite{SFD-RPCATensor-115})} \\
\scriptsize{} &\scriptsize{}  &\scriptsize{DGTU-Net (Zhou et al. \cite{SFD-RPCATensor-121})} \\
\scriptsize{} &\scriptsize{}  &\scriptsize{LCPNet (Zhang et al. \cite{SFD-RPCATensor-130})} \\
\hline
\scriptsize{Robust Tensors Completion} &\scriptsize{} &\scriptsize{ (Zhu et al. \cite{SFD-RTC-1}),(Deng et al. \cite{SFD-RTC-2}), NRLTC (Chen et al. \cite{SFD-RTC-3}) } \\ 
\scriptsize{} &\scriptsize{} &\scriptsize{SSTTC (Xia et al. \cite{SFD-RTC-4}),CEM-AD  (Wang et al. \cite{SFD-RPCATensor-101}) } \\ 
\hline
\end{tabular}}
\caption{Multi-Frame Detection Unsupervised Tensor Decomposition Methods: An Overview (90 papers).}
\label{OverviewTensor}
\end{center}
\end{table*}

\begin{enumerate}
\item  \textbf{Nuclear-norm and $l_1$-norm Defects}  As the nuclear norm is applied on the low rank component related to the background, changing this norm  and adding constraints can increase the robustness in the presence complex and noisy backgrounds. \\ 

\textbf{Weighted Norm:} In 2018, Sun et al. \cite{SFD-RPCATensor-7} propose to employ a weighted tensor nuclear norm. More specifically, a weight tensor is incorporated with tensor nuclear norm to preserve the background edges better. Then, the separation between target and background is formulated as a convex weighted tensor RPCA model. Experiments show that WNRIPT outperforms IPI and RIPT. In 2019, Sun et al. \cite{SFD-RPCATensor-11} propose a Spatial–Temporal Infrared Patch-Tensor (STIPT) structure. Considering the overshrinkage problem in low-rank component estimation field caused by vanilla nuclear norm and weighted nuclear norm, weighted Schatten p-norm is employed to improve the performance by considering the particular meaning of different singular values. An Alternating Direction Method of Multipliers (ADMM) is employed as solver and an adaptive threshold is used to separate the targets. Experiments show that STIPT-WNSM outperforms RIPT \cite{SFD-RPCATensor-3}, WNRIPT \cite{SFD-RPCATensor-7}, WSNMIPT and STIPT-WNNM. In 2019, Zhang and Peng \cite{SFD-RPCATensor-5} design a nonconvex low-rank constraint named Partial Sum of Tensor Nuclear Norm (PSTNN) joint weighted $l_1$-norm to suppress the background and preserve the target. An improved local prior map is also employed to simultaneously encode with target-related and background-related information into the model. A reweighted scheme for enhancing the sparsity and high-efficiency version of tensor singular value decomposition (t-SVD) allow to reduce the algorithm complexity and computation time. Experiments show that PSTNN provide better results than NRAM \cite{SFD-RPCAMatrix-10}, NIPPS, ReWIPI and RIPT. To avoid the problem of insufficient structural information extraction in PSTNN, Sun et al. \cite{SFD-RPCATensor-68} employ the local contrast difference between the target and the background designing a  double-neighborhood local contrast based on direction residual weighting method (DNLCDRW). It improves the PSTNN's ability to suppress bright and complex backgrounds by constraining the sparse terms through DNLCDRW, and better improves the ability of IR weak target detection. In 2020, Guan et al. \cite{SFD-RPCATensor-6} employ a non-convex tensor rank surrogate joint local contrast energy (NTRS). More specifically, a non-convex tensor rank surrogate merging Tensor Nuclear Norm (TNN) and the Laplace function are employed for low rank background patch-tensor constraint. It allows to adaptively allocate weight for every singular value and better approximate $l_0$-norm. Considering that the local prior map is equivalent to the saliency map, Guan et al. \cite{SFD-RPCATensor-6} design a local contrast energy feature into IPT detection framework to weight target tensor. To remove the structured edges, Guan et al. \cite{SFD-RPCATensor-6} employ an additional structured sparse regularization term using the $l_{1,1,2}$-norm of third-order tensor. Experiments show that NTRS outperforms IPI and RIPT. In 2025, Lu et al. \cite{SFD-RPCATensor-69} design a method based on a spatial–temporal feature fusion tensor model. The partial tubal nuclear norm is employed to constrain  the low-rank background. Experiments show that STFFT provides better results than Tophat, LCM \cite{SFD-LCM-4}, PSTNN and MSLSTIPT \cite{SFD-ML-2}. \\  

\textbf{Total Variation Norm:}  Traditional Total Variation (TV) regularization based method only considers the spatial information of a single frame, and its efficiency cannot meet the real-time processing requirement. To address this limitation, Sun et al. \cite{SFD-RPCATensor-10}  design an anisotropic spatial-temporal TV regularization to use both single-frame and inter-frame information. The spatial-temporal TV regularization and weighted IPT model (STTV-WNIPT) allows to separate the target and background. For the low-rank background component, the weighted tensor nuclear norm and spatial-temporal TV regularization are employed to describe the smoothness. For sparse target component and a noise component, the $l_1$-norm and Frobenius norm terms are used, respectively. Experiments show that STTV-WNIPT outperforms TopHat, IPI \cite{SFD-RPCAMatrix-1}, WNRIPT and TV-PCP.  In 2023, Wang et al. \cite{SFD-RPCATensor-44} employ a 3D Total variation (3D-TV) to regularize the foreground to take into account the spatial-temporal continuity of the target and to suppress the interference caused by dynamic background and target movement on the foreground extraction. A solver allows he minimization of the non-convex spatial-temporal tensor low-rank approximation algorithm (MNSTLA). Experiments show that MNSTLA provide most of the time better results than  NRAM, RIPT, and WSNMSTIPT. In 2023, Zhao and Xu \cite{SFD-RPCATensor-62} design a Tensor $l_{2,1}$-Norm Minimization via T-QR (TLNMTQR) method to isolate the noise and improve computational speed without compromising accuracy. By employing an Asymmetric Spatial-Temporal Total Variation regularization method (ASSTV), the flexibility and efficacy is improved to handle time series data. Experiments show that TLNMTQR outperforms PSTNN \cite{SFD-RPCATensor-5}, MTFRA and MSLSTIPT \cite{SFD-ML-2}.  In 2025, Ge et al. \cite{SFD-RPCATensor-93} propose to use both spatial and temporal information. First, a multi-scale energy
movement saliency map (MESM) is employed for target motion characteristics. This map serves as a sparse prior and is incorporated into the spatio-temporal tensor model giving strong decomposition guidance even when thetarget contrast is weak or overlapped with strong edges. Second, a reweighted scheme integrating motion saliency is designed
based on the minimum temporal projection of the target component after each iteration to suppress background clutter and
enhancing target separation. Third, an overlapping group sparse total variation (OGSTV) regularization from 2D to 3D is used. Experiments show that 3D-OGSTV better distinguishes edges from flat regions than ASTTV-NTLA \cite{SFD-RPCATensor-32}, LogTFNN \cite{SFD-RPCATensor-20}, ECA-STT \cite{SFD-RPCATensor-16} and IMNN-LWEC \cite{SFD-RPCATensor-34}. In 2026, Chen et al. \cite{SFD-RPCATensor-99} design a structured regularization method named $l_p$-MCTV that integrates the $l_p$-norm and a Min-max-Concave Total Variation (MCTV) regularization to improve detection accuracy. To suppress the interference due to the complex backgrounds, Chen et al. \cite{SFD-RPCATensor-99} use a partial sum minimization of singular values strategy to estimate the background component. In addition, the $l_p$-norm regularization allows to extract the target. The MCTV regularization allows to model the noise component. \\

\textbf{Dynamic Backgrounds:} To deal with dynamic backgrounds, Sun et al. \cite{SFD-RPCATensor-70} design an Adaptive Spatial-Temporal Infrared Tensor (ASTIT) and Weighted Tensor Average Rank Approximation (WPTAR) method. The ASTIT
construction method uses the structural similarity index measure (SSIM) to dynamically adjust the temporal step to obtain physically accurate representation of background changes and maintaining its low
rank property. Transpose errors are solved by using the weighted Schatten $p$-norm and tensor average rank (WPTAR) norm  to increase the
physical fidelity of the data representation.  Sun et al. \cite{SFD-RPCATensor-70} approximate the $l_0$ norm with the $p$-norm  reconstruction accuracy obtaining a more precise physical depiction of sparse targets.  Experiments show that ASTIT-WPTAR outperforms RCTVW \cite{SFD-RPCAMatrix-40}, 4DST-BTMD, IMNN-LWEC \cite{SFD-RPCATensor-34}, ASTTV-NTLA \cite{SFD-RPCATensor-32} and ECASTT.  In 2025, Luo et al. \cite{SFD-RPCATensor-71} propose a 5-D Spatial–Temporal Factor-based Completion model (5D-STFC). More specifically, a 5-D whitened spatial–temporal patch-tensor is built. Second, Luo et al. \cite{SFD-RPCATensor-71} employ a spatial–temporal factor-based low-rank background estimation norm and a Moreau envelope-derived sparsity estimation norm based on joint spatial–temporal knowledge. Experiments show that 5D-STFC provides better results than FAMSIS \cite{PSTD-12}, SSTTC \cite{SFD-RTC-4}, 4D-TR \cite{SFD-RPCATensor-36} and AGPCNet \cite{DL-21}. \\

\textbf{Facet Model:} In 2020, Dong et al. \cite{SFD-RPCATensor-12}
combines a first-order directional derivative of facet model with structural tensor. Four directional derivatives of the
original infrared image are computed from the Facet model. Following the directional derivative feature of the target, the structure tensor eigenvalue matrix allows to enhance the target and to suppress the background clutter. Four salient maps are thus obtained and a final fusion map is obtained by multiplying the salient maps of all directions. Experiments show that Facet-LRSD outperforms Top-hat and LCM. In 2021, Pang et al. \cite{SFD-RPCATensor-13} combines a Facet Derivative-based Multidirectional Edge Awareness with Spatial–Temporal Tensor (FDMDEA-STT) The problem is formulated with the prior information of the target and
background in the spatial–temporal domain. Based on the Facet derivative, a multidirectional edge awareness mapping is employed.  Pang et al. \cite{SFD-RPCATensor-13} fuse it into the model as sparse prior information. Experiments show that FDMDEA-STT provide better results than RIPT and ECA-SST. \\

\textbf{Graph Laplacian Regularization:} In a further work, Luo et al. \cite{SFD-RPCATensor-72} propose an Hypergraph Laplacian regularization and Asymmetric Penalty function-based Regularization (HGLAPR). More specifically, a hypergraph structure is built and a hypergraph Laplacian regularization is designed as well as a Laplace-based tensor nuclear norm for low-rank background estimation. Moreover, an asymmetric penalty function-based sparsity regularization is employed for more accurate target estimation. Experiments show that HGLAPR outperforms MSLSTIPT \cite{SFD-ML-2}, 4D-TR \cite{SFD-RPCATensor-36}, AGPCNet \cite{DL-21} and RPCANet \cite{SFD-RPCAMatrix-46}. In 2025, Liu et al. \cite{SFD-RPCATensor-75} also use a Fast Graph Laplacian Regularization (FGLR) instead of nuclear norm to avoid singular value decomposition computation and to reduce computational complexity. In addition, Liu et al. \cite{SFD-RPCATensor-75} employ Minimax Concave Penalty (MCP) function instead of $l_1$-norm to constrains the sparsity of targets. Experiments show that FGLR-MCP provides better results than Top-hat, SMSL, IMNN-LWEC \cite{SFD-RPCATensor-34} and 4D-TR \cite{SFD-RPCATensor-36}. In 2025, Ge et al. \cite{SFD-RPCATensor-102} propose a spectral residual method and design it as a convex regularizer significantly reducing computational costs. In addition, a bimodal slice-based graph Laplacian regularization is employed. Graphs are constructed with slices as vertices in both temporal and spatial dimensions in order to
capture local geometric features while suppressing the complexity of pixel-level graph construction. In addition, Ge et al. \cite{SFD-RPCATensor-102} 
combine single-frame priors and low-cost multi-frame motion posterior information to improve target extraction accuracy. Experiments show that  GC outperforms NFTDGSTV \cite{SFD-RPCATensor-92}, MSLSTIPT \cite{SFD-ML-2}, ASTTV-NTLA \cite{SFD-RPCATensor-32}, 4DST-BTMD \cite{SFD-RPCATensor-65} and FGLR-MCP. \\

\item  \textbf{Heterogeneous Backgrounds} When facing highly heterogeneous backgrounds, they would not perform very well because of  the existence of strong edges and other interfering components and ignoring the priors fully. In 2017, Dai and Wu \cite{SFD-RPCATensor-3} propose a Reweighted Infrared Patch-Tensor (RIPT) model with both non-local and local priors. More specifically, Dai and Wu \cite{SFD-RPCATensor-3} design an entry-wise local structure-adaptive and sparsity enhancing weight to replace the globally constant weighting parameter. The decomposition is then achieved via the element-wise reweighted higher-order RPCA with an additional convergence condition according to the practical situation of target detection. Experiments show that RIPT outperforms IPI \cite{SFD-RPCAMatrix-1}, WIPI, NIPPS, IPT and WIPT. In 2023, Zhang et al. \cite{SFD-RPCATensor-40} design a self-adaptive low-rank sparse tensor decomposition model. The entropy of image block is employed for fast matching of non-local similar blocks to obtain a better sparse tensor for small targets. The adaptive strategy of low-rank sparse tensor decomposition is employed for different background environments, which adaptively compute the weight coefficients. Experiments show that ANLPT outperforms RIPT, PSTNN \cite{SFD-RPCATensor-5} and NTFRA \cite{SFD-RPCATensor-20}.\\

\item  \textbf{Computation Time}
In 2025, Li \cite{SFD-RPCATensor-95} propose a four-dimensional infrared tensor using a Bidirectional Tensor Ring (BTR) decomposition to separate weak spatial correlations from strong temporal-patch correlations while simultaneously capturing interactions between these two components. This model is solved with a Proximal Alternating Minimization (PAM) framework. Experiments demonstrate that BTR-ISTD outperforms IPI \cite{SFD-RPCAMatrix-1}, PSTNN \cite{SFD-RPCATensor-5}, RIPTT, ECA-STT in terms of detection accuracy, dynamic backgrounds suppression capability, and computational speed.
\end{enumerate}

\begin{table*}
\begin{center}
\scalebox{0.70}{
\begin{tabular}{|l|l|l|l|l|l|l|}
\hline
\scriptsize{Methods}&\scriptsize{Decomposition} &\scriptsize{Minimization} &\scriptsize{Constraints}   &\scriptsize{Solver}  &\scriptsize{Convexity} 
&\scriptsize{Complexity}
\\
\hline
\hline
\scriptsize{\textbf{ReIPT (RIPT) (2017)}} 
&\scriptsize{$\mathcal{A}=\mathcal{L}+\mathcal{S}+\mathcal{E}$} 
&\scriptsize{$\underset{\mathcal{L},\mathcal{S}}{\text{min}} ~~ \Sigma_{i=1}^3||\mathcal{L}_i||_{*} + \lambda ||\mathcal{W}_{LS} \odot \mathcal{S}||_{l_{1}}$}
&\scriptsize{$||\mathcal{A}-\mathcal{L}-\mathcal{S}||_F \leq \delta$}
&\scriptsize{ADMM \cite{SFD-RPCATensor-3}}
&\scriptsize{No}
&\scriptsize{$O_{iter}(mn\min(m,n) + m^{3/2}n + mn)$} \\ 
\scriptsize{Dai and Wu \cite{SFD-RPCATensor-3}} 
&\scriptsize{} 
&\scriptsize{$\mathcal{W}_{LS}$ (sparsity enhancement weight tensor)}
&\scriptsize{}
&\scriptsize{}
&\scriptsize{}
&\scriptsize{}\\
\hline
\scriptsize{\textbf{WNRIPT (2018)}} 
&\scriptsize{$\mathcal{A}=\mathcal{L}+\mathcal{S}$} 
&\scriptsize{$\underset{\mathcal{L},\mathcal{S}}{\text{min}} ~~ ||\mathcal{L}||_{\mathcal{W}_L,*} + \lambda ||\mathcal{W}_{S} \odot \mathcal{S}||_{l_{1}}$}
&\scriptsize{$\mathcal{A}-\mathcal{L}-\mathcal{S}=0$}
&\scriptsize{ADMM/IALM \cite{SFD-RPCATensor-7}}
&\scriptsize{No}
&\scriptsize{$O_{iter}(m^{3/2}n + mn)$} \\ 
\scriptsize{Sun et al. \cite{SFD-RPCATensor-7}} 
&\scriptsize{} 
&\scriptsize{$\mathcal{W}_{L}$, $\mathcal{W}_{S}$ (weight tensors)}
&\scriptsize{}
&\scriptsize{}
&\scriptsize{}
&\scriptsize{}\\
\hline
\scriptsize{\textbf{PSTNN (2019)}} 
&\scriptsize{$\mathcal{A}=\mathcal{L}+\mathcal{S}$} 
&\scriptsize{$\underset{\mathcal{L},\mathcal{S}}{\text{min}} ~~ ||\mathcal{L}||_{PSTNN} + \lambda ||\mathcal{W} \odot \mathcal{S}||_{l_{1}}$}
&\scriptsize{$\mathcal{A}-\mathcal{L}-\mathcal{S}=0$}
&\scriptsize{ADMM \cite{SFD-RPCATensor-5}}
&\scriptsize{No}
&\scriptsize{$O_{iter}(m^{3/2}n + mn)$} \\ 
\scriptsize{Zhang and Peng \cite{SFD-RPCATensor-5}} 
&\scriptsize{} 
&\scriptsize{$\mathcal{W}$ (weight tensor)}
&\scriptsize{}
&\scriptsize{}
&\scriptsize{}
&\scriptsize{}\\
\hline
\scriptsize{\textbf{WSNM-STIPT (2019)}} 
&\scriptsize{$\mathcal{A}=\mathcal{L}+\mathcal{S}+\mathcal{E}$} 
&\scriptsize{$\underset{\mathcal{L},\mathcal{S}}{\text{min}} ~~ ||\mathcal{L}||_{W,S_p}^p + \lambda ||\mathcal{S}||_{l_{1}}$}
&\scriptsize{$||\mathcal{A}-\mathcal{L}-\mathcal{S}||_F \leq \delta$}
&\scriptsize{ADMM \cite{SFD-RPCATensor-11}}
&\scriptsize{No}
&\scriptsize{$O_{iter}(mnd\min(m,n) + mnd)$} \\ 
\scriptsize{Sun et al. \cite{SFD-RPCATensor-11}} 
&\scriptsize{} 
&\scriptsize{}
&\scriptsize{}
&\scriptsize{}
&\scriptsize{}
&\scriptsize{}\\
\hline
\scriptsize{\textbf{STTV-WNIPT (2019)}} 
&\scriptsize{$\mathcal{A}=\mathcal{L}+\mathcal{S}+\mathcal{E}$} 
&\scriptsize{$\underset{\mathcal{L},\mathcal{S},\mathcal{E}}{\text{min}} ~~ ||\mathcal{L}||_{\mathcal{W}_L,*} + \lambda_1 ||\mathcal{L}||_{SSTV} + \lambda_2 ||\mathcal{S}||_{l_{1}} + \lambda_3||\mathcal{E}||_{F}^2$}
&\scriptsize{$||\mathcal{A}-\mathcal{L}-\mathcal{S}||_F \leq \delta$}
&\scriptsize{ADMM \cite{SFD-RPCATensor-10}}
&\scriptsize{No}
&\scriptsize{$O_{iter}(mnd\min(m,n) + mnd)$} \\ 
\scriptsize{Sun et al. \cite{SFD-RPCATensor-10}} 
&\scriptsize{} 
&\scriptsize{}
&\scriptsize{}
&\scriptsize{}
&\scriptsize{}
&\scriptsize{}\\
\hline
\scriptsize{\textbf{NTRS (2020)}} 
&\scriptsize{$\mathcal{A}=\mathcal{L}+\mathcal{S}+\mathcal{E}$} 
&\scriptsize{$\underset{\mathcal{L},\mathcal{S},\mathcal{E}}{\text{min}} ~~ ||\mathcal{L}||_{\epsilon} + \lambda ||\mathcal{W} \odot \mathcal{S}||_{l_{0}} + \beta ||\mathcal{E}||_{l_{1,1,2}}$}
&\scriptsize{$\mathcal{A}-\mathcal{L}-\mathcal{S}-\mathcal{E}=0$}
&\scriptsize{ADMM \cite{SFD-RPCATensor-6}}
&\scriptsize{No}
&\scriptsize{$O_{iter}(m^{3/2}n + mn)$} \\ 
\scriptsize{Guan et al. \cite{SFD-RPCATensor-6}} 
&\scriptsize{} 
&\scriptsize{$\mathcal{W}$ (weight tensor)}
&\scriptsize{}
&\scriptsize{}
&\scriptsize{}
&\scriptsize{}\\
\hline
\scriptsize{\textbf{FDMDEA-STT (2021)}} 
&\scriptsize{$\mathcal{A}=\mathcal{L}+\mathcal{S}+\mathcal{E}$} 
&\scriptsize{$\underset{\mathcal{L},\mathcal{S},\mathcal{E}}{\text{min}} ~~ P_{ls}(\mathcal{B}) + \lambda \Pi_{i=1}^N P_{ls}^*(\mathcal{L}_i) + \beta ||\mathcal{W} \odot \mathcal{S}||_{l_{1}} + \frac{\gamma}{2} ||\mathcal{E}||_F^2$}
&\scriptsize{$||\mathcal{A}-\mathcal{L}-\mathcal{S}||_F \leq \delta$}
&\scriptsize{ADMM \cite{SFD-RPCATensor-13}}
&\scriptsize{No}
&\scriptsize{$O_{iter}(mnd(m+n+d) + m^3+n^3+d^3 + mnd)$} \\ 
\scriptsize{Pang et al. \cite{SFD-RPCATensor-13}} 
&\scriptsize{} 
&\scriptsize{$P_{ls}$, $P_{ls}^*$ (log-sum forms of the vector and matrix sparsities)} 
&\scriptsize{}
&\scriptsize{}
&\scriptsize{}
&\scriptsize{}\\
\hline
\scriptsize{\textbf{STBMPT (2023)}} 
&\scriptsize{$\mathcal{A}=\mathcal{L}+\mathcal{S}+\mathcal{E}$} 
&\scriptsize{$\underset{\mathcal{L},\mathcal{S}}{\text{min}} ~~ ||\mathcal{L}||_{LBPTNN} + \lambda ||\mathcal{W} \odot \mathcal{S}||_{l_{1}}$}
&\scriptsize{$\mathcal{A}-\mathcal{L}-\mathcal{S}-\mathcal{E}=0$, $||\mathcal{E}||_F \leq \delta$}
&\scriptsize{ADMM \cite{SFD-RPCATensor-42}}
&\scriptsize{No}
&\scriptsize{$O_{iter}(m^{3/2}n + mn)$} \\ 
\scriptsize{Aliha et al. \cite{SFD-RPCATensor-42}} 
&\scriptsize{} 
&\scriptsize{}
&\scriptsize{}
&\scriptsize{}
&\scriptsize{}
&\scriptsize{}\\
\hline
\scriptsize{\textbf{MNSTLA (2023)}} 
&\scriptsize{$\mathcal{A}=\mathcal{L}+\mathcal{S}+\mathcal{E}$} 
&\scriptsize{$\underset{\mathcal{L},\mathcal{S},\mathcal{E}}{\text{min}} ~~ ||\mathcal{L}||_{\gamma,*} + \lambda_1 ||V(\mathcal{L})||_{3D\text{-}TV} + \lambda_2 ||\mathcal{S}||_{l_1} + \lambda_3 ||\mathcal{E}||_{l_{2,1}}$}
&\scriptsize{$\mathcal{A}-\mathcal{L}-\mathcal{S}-\mathcal{E}=0$}
&\scriptsize{ADMM \cite{SFD-RPCATensor-44}}
&\scriptsize{No}
&\scriptsize{$O_{iter}(mnd\min(m,n) + mnd)$} \\ 
\scriptsize{Wang et al. \cite{SFD-RPCATensor-44}} 
&\scriptsize{} 
&\scriptsize{}
&\scriptsize{}
&\scriptsize{}
&\scriptsize{}
&\scriptsize{}\\
\hline
\scriptsize{\textbf{TLNMTQR-ASTTV (2023)}} 
&\scriptsize{$\mathcal{A}=\mathcal{L}+\mathcal{S}+\mathcal{E}$} 
&\scriptsize{$\underset{\mathcal{L},\mathcal{S},\mathcal{E}}{\text{min}} ~~ ||\mathcal{L}||_{l_{2,1}} + \lambda_{TV} ||\mathcal{L}||_{ASTTV} + \lambda_S ||\mathcal{S}||_{l_1} + \lambda_3 ||\mathcal{E}||_{F}^2$}
&\scriptsize{$\mathcal{A}-\mathcal{L}-\mathcal{S}-\mathcal{E}=0$}
&\scriptsize{ADMM \cite{SFD-RPCATensor-62}}
&\scriptsize{No}
&\scriptsize{$O_{iter}(mnrd + mr^2d + nr^2d + r^2d + mnd)$} \\ 
\scriptsize{Zhao and Xu \cite{SFD-RPCATensor-62}} 
&\scriptsize{} 
&\scriptsize{}
&\scriptsize{}
&\scriptsize{}
&\scriptsize{}
&\scriptsize{}\\
\hline
\scriptsize{\textbf{DNLCDRW (2024)}} 
&\scriptsize{$\mathcal{A}=\mathcal{L}+\mathcal{S}+\mathcal{E}$} 
&\scriptsize{$\underset{\mathcal{L},\mathcal{S}}{\text{min}} ~~ ||\mathcal{L}||_{PSTNN} + \lambda ||\mathcal{W} \odot \mathcal{S}||_{l_{1}}$}
&\scriptsize{$\mathcal{A}-\mathcal{L}-\mathcal{S}-\mathcal{E}=0$}
&\scriptsize{ADMM \cite{SFD-RPCATensor-68}}
&\scriptsize{No}
&\scriptsize{$O_{iter}(m^{3/2}n + mn)$} \\ 
\scriptsize{Sun et al. \cite{SFD-RPCATensor-68}} 
&\scriptsize{} 
&\scriptsize{}
&\scriptsize{}
&\scriptsize{}
&\scriptsize{}
&\scriptsize{}\\
\hline
\end{tabular}}
\caption{Low-rank plus Sparse Tensor Decomposition for Infrared Small Target Detection: An homogeneous overview (Part 1). $\mathcal{A} \in \mathbb{R}^{m \times n \times d}$ is the observation tensor, where $m \times n$ is the size of each frontal slice and $d$ is the number of slices (frames or patches), and $r$ is the tensor (tubal) rank.}
\label{OverviewT-1}
\end{center}
\end{table*}

\begin{table*}
\begin{center}
\scalebox{0.70}{
\begin{tabular}{|l|l|l|l|l|l|l|}
\hline
\scriptsize{Methods}&\scriptsize{Decomposition} &\scriptsize{Minimization} &\scriptsize{Constraints}   &\scriptsize{Solver}  &\scriptsize{Convexity} 
&\scriptsize{Complexity}
\\
\hline
\hline
\scriptsize{\textbf{STFFT (2025)}} 
&\scriptsize{$\mathcal{A}=\mathcal{L}+\mathcal{S}$} 
&\scriptsize{$\underset{\mathcal{L},\mathcal{S}}{\text{min}} ~~ ||\mathcal{L}||_{PTNN} + \lambda_1 ||\mathcal{W} \odot \mathcal{S}||_{l_1}$}
&\scriptsize{$\mathcal{A}-\mathcal{L}-\mathcal{S}=0$}
&\scriptsize{ADMM \cite{SFD-RPCATensor-69}}
&\scriptsize{No}
&\scriptsize{$O_{iter}(mnd\min(m,n) + mnd)$} \\ 
\scriptsize{Lu et al. \cite{SFD-RPCATensor-69}} 
&\scriptsize{} 
&\scriptsize{$\mathcal{W}$ (sparsity weight tensor)}
&\scriptsize{}
&\scriptsize{}
&\scriptsize{}
&\scriptsize{}\\
\hline
\scriptsize{\textbf{ASTIT-WPTAR (2025)}} 
&\scriptsize{$\mathcal{A}=\mathcal{L}+\mathcal{S}+\mathcal{E}$} 
&\scriptsize{$\underset{\mathcal{L},\mathcal{S}}{\text{min}} ~~ \Sigma_{k=1}^3 ||\mathcal{L}^{T_k}||_{wp,wa} + \lambda ||\mathcal{S}||_{l_p}^p$}
&\scriptsize{$||\mathcal{A}-\mathcal{L}-\mathcal{S}||_F \leq \delta$}
&\scriptsize{ADMM \cite{SFD-RPCATensor-70}}
&\scriptsize{No}
&\scriptsize{$O_{iter}(mnd\min(m,n) + mnd^2 + mnd)$} \\ 
\scriptsize{Sun et al. \cite{SFD-RPCATensor-70}} 
&\scriptsize{} 
&\scriptsize{$\mathcal{L}^{T_k}$: $k$-th mode transform of $\mathcal{L}$.}
&\scriptsize{}
&\scriptsize{}
&\scriptsize{}
&\scriptsize{}\\
\hline
\scriptsize{\textbf{5D-STFC (2025)}} 
&\scriptsize{$\mathcal{A}^{5D}=\mathcal{L}^{5D}+\mathcal{S}^{5D}+\mathcal{E}^{5D}$} 
&\scriptsize{$\underset{\mathcal{L}^{5D},\mathcal{S}^{5D},\mathcal{E}^{5D}}{\text{min}} ~~ ||\mathcal{L}^{5D}||_{STF\text{-}N} + \lambda_1 ||\mathcal{W}^{5D} \odot \mathcal{S}^{5D}||_{STK\text{-}S} + \lambda_2 ||\mathcal{E}^{5D}||_F^2$}
&\scriptsize{$\mathcal{A}^{5D}-\mathcal{L}^{5D}-\mathcal{S}^{5D}-\mathcal{E}^{5D}=0$}
&\scriptsize{Multi-Block ADMM \cite{SFD-RPCATensor-71}}
&\scriptsize{No}
&\scriptsize{$O_{iter}(mnd\min(m,n) + mnd)$} \\ 
\scriptsize{Luo et al. \cite{SFD-RPCATensor-71}} 
&\scriptsize{} 
&\scriptsize{$\mathcal{A}^{5D}$: 5-D tensor; $d$: product of non-spatial modes.}
&\scriptsize{}
&\scriptsize{}
&\scriptsize{}
&\scriptsize{}\\
\hline
\scriptsize{\textbf{HGLAPR (2025)}} 
&\scriptsize{$\mathcal{A}=\mathcal{L}+\mathcal{S}+\mathcal{E}$} 
&\scriptsize{$\underset{\mathcal{L},\mathcal{S},\mathcal{E}}{\text{min}} ~~ ||\mathcal{L}||_{*,\epsilon} + \lambda_1 g(\mathcal{W} \odot \mathcal{S}) + \lambda_2 ||\mathcal{E}||_F^2 + \lambda_3 tr(\mathcal{L}^T B\mathcal{L})$}
&\scriptsize{$\mathcal{A}-\mathcal{L}-\mathcal{S}-\mathcal{E}=0$}
&\scriptsize{ADMM \cite{SFD-RPCATensor-72}}
&\scriptsize{No}
&\scriptsize{$O_{iter}(mnd\min(m,n) + mnd^2 + d^3 + mnd)$} \\ 
\scriptsize{Luo et al. \cite{SFD-RPCATensor-72}} 
&\scriptsize{} 
&\scriptsize{$B$ (hypergraph Laplacian matrix).}
&\scriptsize{}
&\scriptsize{}
&\scriptsize{}
&\scriptsize{}\\
\hline
\scriptsize{\textbf{FGLR-MCP (2025)}} 
&\scriptsize{$\mathcal{A}=\mathcal{L}+\mathcal{S}+\mathcal{E}$} 
&\scriptsize{$\underset{\mathcal{L},\mathcal{S}}{\text{min}} ~~ tr(\mathcal{L}\mathcal{B}\mathcal{L}^T) + \lambda_1 \Psi_1(\mathcal{S})$}
&\scriptsize{$\mathcal{A}-\mathcal{L}-\mathcal{S}-\mathcal{E}=0$}
&\scriptsize{ADMM \cite{SFD-RPCATensor-75}}
&\scriptsize{No}
&\scriptsize{$O_{iter}(mnd^2 + d^3 + mnd)$} \\ 
\scriptsize{Liu et al. \cite{SFD-RPCATensor-75}} 
&\scriptsize{} 
&\scriptsize{$\Psi_1(\cdot)$ is the MCP function, $\mathcal{B}$ (graph Laplacian).} 
&\scriptsize{}
&\scriptsize{}
&\scriptsize{}
&\scriptsize{}\\
\hline
\scriptsize{\textbf{3D-OGSTV (DMS-OGSTV) (2025)}} 
&\scriptsize{$\mathcal{A}=\mathcal{L}+\mathcal{S}+\mathcal{E}$} 
&\scriptsize{$\underset{\mathcal{L},\mathcal{S},\mathcal{E}}{\text{min}} ~~ \Sigma_n \alpha_n ||\mathcal{U}_n||_{l_r^p} + \lambda_1 ||\mathcal{B}||_{l_1} + \lambda_2 ||\mathcal{W}_p \odot \mathcal{S}||_{W_d,l_1}$}
&\scriptsize{$\mathcal{A}-\mathcal{L}-\mathcal{S}-\mathcal{E}=0$}
&\scriptsize{ADMM \cite{SFD-RPCATensor-93}}
&\scriptsize{No}
&\scriptsize{$O_{iter}(mndr + (m+n+d)r^2 + mnd)$} \\ 
\scriptsize{Ge et al. \cite{SFD-RPCATensor-93}} 
&\scriptsize{} 
&\scriptsize{$+ \lambda_3 ||\mathcal{L}||_{3D\text{-}OGSTV} + \lambda_4 ||\mathcal{E}||_{F}^2$}
&\scriptsize{}
&\scriptsize{}
&\scriptsize{}
&\scriptsize{}\\
\scriptsize{} 
&\scriptsize{} 
&\scriptsize{$\mathcal{L}=\mathcal{B} \times_1 \mathcal{U}_1 \times_2 \mathcal{U}_2 \cdots \times_N \mathcal{U}_N$}
&\scriptsize{}
&\scriptsize{}
&\scriptsize{}
&\scriptsize{}\\
\hline
\scriptsize{\textbf{GLLS-LMAG (2025)}} 
&\scriptsize{$\mathcal{A}=\mathcal{L}+\mathcal{S}$} 
&\scriptsize{$\underset{\mathcal{L},\mathcal{S}}{\text{min}} ~~ ||\mathcal{L}||_{GLLS} + ||\mathcal{L}||_{LMAG} + \lambda ||\mathcal{S}||_{\omega,l_{1}}$}
&\scriptsize{$\mathcal{A}-\mathcal{L}-\mathcal{S}=0$} 
&\scriptsize{ADMM \cite{SFD-RPCAMatrix-62}}
&\scriptsize{No}
&\scriptsize{$O_{iter}(mnd\min(m,n) + mnd)$} \\
\scriptsize{Li et al. \cite{SFD-RPCAMatrix-62}} 
&\scriptsize{} 
&\scriptsize{}
&\scriptsize{}
&\scriptsize{}
&\scriptsize{}
&\scriptsize{} \\
\hline
\scriptsize{\textbf{Improved IPI (2025)}} 
&\scriptsize{$\mathcal{A}=\mathcal{L}+\mathcal{S}+\mathcal{E}$} 
&\scriptsize{$\underset{\mathcal{L},\mathcal{S},\mathcal{E}}{\text{min}} ~~ ||\mathcal{L}||_{*} + \lambda_1 TV(\mathcal{L}) + \lambda_2 ||\mathcal{S} \odot \mathcal{W}||_{l_1}$}
&\scriptsize{$\mathcal{A}-\mathcal{L}-\mathcal{S}-\mathcal{E}=0$, $||\mathcal{E}||_F \leq \delta$} 
&\scriptsize{ADMM \cite{SFD-RPCAMatrix-69}}
&\scriptsize{Yes}
&\scriptsize{$O_{iter}(mn\min(m,n) + m^2n + mn^2$} \\
\scriptsize{Deng et al. \cite{SFD-RPCAMatrix-69}}
&\scriptsize{} 
&\scriptsize{$\mathcal{W}$ (corner strength function).}
&\scriptsize{}
&\scriptsize{}
&\scriptsize{}
&\scriptsize{$+ m^3 + n^3 + mn)$} \\
\scriptsize{}
&\scriptsize{} 
&\scriptsize{$TV(\cdot)$ (bilateral filtering regularization).}
&\scriptsize{}
&\scriptsize{}
&\scriptsize{}
&\scriptsize{} \\
\hline
\scriptsize{\textbf{GC (2025)}} 
&\scriptsize{$\mathcal{A}=\mathcal{L}+\mathcal{S}+\mathcal{E}$} 
&\scriptsize{$\underset{\mathcal{L},\mathcal{S},\mathcal{E}}{\text{min}} ~~ \Sigma_{n=1}^q ||\mathcal{U}_n-\mathcal{U}_{cs,n}||_{F}^2$}
&\scriptsize{$\mathcal{A}-\mathcal{L}-\mathcal{S}-\mathcal{E}=0$} 
&\scriptsize{ADMM \cite{SFD-RPCATensor-102}}
&\scriptsize{Yes}
&\scriptsize{$O_{iter}(m^3d + m^2nd + mnd^2 + m^3 + d^3 + mnd)$} \\
\scriptsize{Ge et al. \cite{SFD-RPCATensor-102}}
&\scriptsize{}
&\scriptsize{$+\Sigma_{n=1}^2 \theta_n tr(\mathcal{L}_n^T \Psi_n \mathcal{L}_n) + \lambda_1 ||\mathcal{W}_p \odot \mathcal{S}||_{l_1} + \lambda_2 ||\mathcal{E}||_F^2$}
&\scriptsize{}
&\scriptsize{}
&\scriptsize{}
&\scriptsize{} \\
\scriptsize{}
&\scriptsize{}
&\scriptsize{$\mathcal{L}_n=\mathcal{E}_n\mathcal{U}_n$}
&\scriptsize{}
&\scriptsize{}
&\scriptsize{}
&\scriptsize{} \\
\hline
\scriptsize{\textbf{MSR-FT (2026)}} 
&\scriptsize{$\mathcal{A}=\mathcal{L}+\mathcal{S}+\mathcal{E}$} 
&\scriptsize{$\underset{\mathcal{L},\mathcal{S},\mathcal{E}}{\text{min}} ~~ ||\mathcal{L}||_{MSTNN} + \lambda_1 Prox_{p_\lambda}(\mathcal{W} \odot \mathcal{S}) + \lambda_2 ||\mathcal{E}||_{F}^2$}
&\scriptsize{$\mathcal{A}-\mathcal{L}-\mathcal{S}-\mathcal{E}=0$} 
&\scriptsize{ADMM \cite{SFD-RPCAMatrix-75}}
&\scriptsize{No}
&\scriptsize{$O_{iter}(mnd\min(m,n) + mnd)$} \\
\scriptsize{Li et al. \cite{SFD-RPCAMatrix-75}}
&\scriptsize{} 
&\scriptsize{$\mathcal{W}$ (weighted saliency tensor).}
&\scriptsize{}
&\scriptsize{}
&\scriptsize{}
&\scriptsize{} \\
\hline
\scriptsize{\textbf{STC-LRSD (2026)}} 
&\scriptsize{$\mathcal{A}^{4D}=\mathcal{L}^{4D}+\mathcal{S}^{4D}$} 
&\scriptsize{$\underset{\mathcal{L}^{4D},\mathcal{S}^{4D}}{\text{min}} ~~ \Sigma_{n=1}^{N-1} \beta_n ||\mathcal{L}^{4D}||_{*} + \alpha ||\mathcal{S}^{4D}||_{l_{1}}$}
&\scriptsize{$\mathcal{A}^{4D}-\mathcal{L}^{4D}-\mathcal{S}^{4D}=0$} 
&\scriptsize{ADMM \cite{SFD-RPCAMatrix-76}}
&\scriptsize{Yes}
&\scriptsize{$O_{iter}(mnd\min(m,n) + mnd)$} \\
\scriptsize{Fan et al. \cite{SFD-RPCAMatrix-76}}
&\scriptsize{} 
&\scriptsize{$\mathcal{A}^{4D}$: 4-D tensor; $d$: product of non-spatial modes.}
&\scriptsize{}
&\scriptsize{}
&\scriptsize{}
&\scriptsize{} \\
\hline
\scriptsize{\textbf{Lp-MCTV (2026)}} 
&\scriptsize{$\mathcal{A}=\mathcal{L}+\mathcal{S}+\mathcal{E}$} 
&\scriptsize{$\underset{\mathcal{L},\mathcal{S},\mathcal{E}}{\text{min}} ~~ ||\mathcal{L}||_{p=r} + \lambda_1 ||\mathcal{S}||_{l_p}^p + \lambda_2 ||\mathcal{E}||_{MCTV}$}
&\scriptsize{$\mathcal{A}-\mathcal{L}-\mathcal{S}-\mathcal{E}=0$} 
&\scriptsize{ADMM \cite{SFD-RPCATensor-99}}
&\scriptsize{No}
&\scriptsize{$O_{iter}(mnd\min(m,n) + mnd)$} \\
\scriptsize{Chen et al. \cite{SFD-RPCATensor-99}}
&\scriptsize{} 
&\scriptsize{}
&\scriptsize{}
&\scriptsize{}
&\scriptsize{}
&\scriptsize{} \\
\hline
\end{tabular}}
\caption{Low-rank plus Sparse Tensor Decomposition for Infrared Small Target Detection: An homogeneous overview (Part 2). $\mathcal{A} \in \mathbb{R}^{m \times n \times d}$ is the observation tensor, where $m \times n$ is the size of each frontal slice and $d$ is the number of slices (frames or patches), and $r$ is the tensor (tubal) rank.}
\label{OverviewT-2}
\end{center}
\end{table*}

\subsubsection{RTC Tensor based Decomposition}
\label{RTCTensor}
In 2020, Zhu et al. \cite{SFD-RTC-1} design a low-rank tensor completion with Top-Hat regularization in order to take both the structural prior knowledge of the target and the self-correlation of the background into account. First, a tensor model for the high-dimensional structural characteristics of multiframe infrared images is constructed. Second, a method based on low-rank tensor completion with top-hat regularization is developed. Third, a closed solution to the optimization algorithm is given to solve the formulation problem. In a further work, Deng et al. \cite{SFD-RTC-2} propose a generalized low-rank double-tensor nuclear norm completion framework. More specifically, this LRSD framework employ the double nuclear norm and ring-structural elements. Indeed, the double nuclear norm instead of the single nuclear norm as is employ in the relaxation of the rank function. It solves the problem that the suboptimal solution deviates from the original solution and better approaches the rank minimization. Moreover, weighted ring structural elements are used instead of traditional structural elements to make  better use of the target information and its surrounding background. In 2026, Chen et al. \cite{SFD-RTC-3} propose a Nonconvex Relaxed Low-Rank Tensor Completion (NRLTC) approximated by using a non convex function to the singular values of the tensor’s squared processing matrix. In addition, Chen et al. \cite{SFD-RTC-3} develop a reweighted multidirectional structural prior which adaptively integrates multidirectional edge responses to enforce true target structures while eliminating strong background edges and discrete noise. Thus, the prior preserves target details and avoid overshrinkage during detection. Experiments show that NRLTC outperforms IPI and NRAM.

\subsubsection{Multi-subspace Learning}
\label{MultisubspaceLearning}
Previous robust subspace methods seek for one subspace presenting weaknesses to consider the complex and heterogeneous backgrounds in infrared images. In 2017, Wang et al. \cite{SFD-ML-1} propose a stable multi-subspace learning approach to deal with heterogeneous scene.  \\

In 2021, Sun et al. \cite{SFD-ML-2} design a Multiple Subspace Learning and Spatial-Temporal Patch-Tensor (MSLSTIPT) framework. First, a tensor data structure is built to employ inner correlation in spatial and temporal domain. Second, MSLSTIPT promotes the multi-subspace property to tensor domain to separate the target and the background more accurately.  Third, an optimization algorithm based on Alternating Direction Method of Multipliers (ADMM) allows to solve the corresponding minimization problem.


\section{Resources and Codes}
\label{RDC}

\subsection{Target Detection WebSite}
\label{Resources}
This website currently lists 610 of the references reviewed here, links to 19 available datasets and to 161 codes. For each category, the list is regularly updated and classified following the mathematical, machine learning and signal processing models. An overview of the content of the Target Detection Website\footnote{https://sites.google.com/site/targetdetectionwebsite/} is given at the home page.

\subsection{Codes}
\label{Codes}
Several codes are available on GitHub and provided by the authors. Moreover, repositories groups a collection of SOTA algorithms and toolbox solutions are also available as follows:
\begin{itemize}
\item \textbf{Repositories:} First, Tianfang Zhang groups approximatively 67 links to the codes released on GitHub, 107 links to papers in the field, 18 links to datasets and three recommended surveys papers at his Awesome Infrared Small Targets GitHub page\footnote{https://github.com/Tianfang-Zhang/awesome-infrared-small-targets}. These codes are classified into the following categories: Background suppression-based methods (7 papers, 2 codes), human visual system-based methods (15 papers), optimization-based methods (30 papers, 18 codes), deep learning-based methods (51 papers, 42 codes), and deep unfolding-based methods (4 papers, 4 codes). Even if this repository is directly provided on GitHub and classified algorithms into categories, the Target Detection Website provides more links with 161 codes.
\item \textbf{Toolbox Solutions:} For the ICPR 2024 Resource-Limited Infrared Small Target Detection Challenge, a toolbox named BasicIRSTD was provided. More specifically, BasicIRSTD\footnote{https://github.com/XinyiYing/BasicIRSTD} is a PyTorch-based open-source and easy-to-use toolbox for infrared small target detection. This toolbox provides a simple pipeline to train/test methods, and builds a benchmark to evaluate the performance of existing methods. The BasicIRSTD helps researchers to get access to infrared small target detection quickly, and facilitates the development of novel methods. In 2025, Wu et al. \cite{PE-2} provide the TensorISTD toolbox\footnote{https://github.com/fengyiwu98/TensorISTD} \cite{PE-2} which is an open-source MATLAB toolbox designed for optimization-based ISTD research and evaluation. TensorISTD provides a unified, streamlined evaluation pipeline covering: Single-frame/Multi-frame detection, Matrix/ Tensor representations, and evaluation metrics (SCRG, BSF, CG, and 3D-ROC series). TensorISTD allows quick prototyping and verification of optimization-based detection methods. In summary, TensorISTD provides is a toolbox for LRSD methods while BasicIRSTD appears to be able to manage all the categories of methods.
\end{itemize}

\section{Datasets}
\label{Datasets}
The various datasets can be classified into two main categories: visual datasets and datasets that includes both image and text modalities. They  can be further sub-classified into small-scale datasets and large-scale datasets. In 2026, Zhang et al. \cite{DL-501} divided datasets into two categories: sparse-target datasets and dense-target datasets. \\

The first datasets have been designed for sparse-target scenarios, where each frame contains only a few isolated targets and the minimum inter-target distance is relatively large. These datasets are thus focus on sparse-target conditions and do not reflect scenarios where multiple small targets appear densely within a single frame. To address these limitations, researchers have increasingly shifted their attention toward dense infrared small target detection (dense ISTD) \cite{DL-501}.  

\subsection{Visual Modality Datasets}
Visual modality datasets are the first datasets available in the field and they progressively move from sparse-target datasets/small-size datasets up to dense-target datasets/large scale datasets. These datasets can be categorized as follows.

\subsubsection{General Environments}
This section lists both small and large-scale datasets for sparse and dense target detections. The datasets can be listed as follows.

\subsubsection{Small-Scale Datasets}
Small-scale datasets have been developed at the beginning of serious performance evaluation of infrared small target detection and they can be enumerated as follows:
\begin{itemize}
\item \textbf{SIRST Dataset (2019):} This dataset provided by Wang et al. \cite{DL-15} includes 10000 training images and 100 test images. Neverthless, many targets do not meet the definition of Society of Photo-Optical Instrumentation Engineers (SPIE) and present synthesized traces with illogical annotations. These problems may lead to the inapplicability toward SIRST detection \cite{DL-12}. 
\item \textbf{NUST-SIRST Dataset (2019):} The NUST-SIRST \footnote{https://github.com/wanghuanphd/MDvsFA\_cGAN} dataset introduced by Wang et al. \cite{DL-15} in 2019 contains approximately 10,000 training images and 100 test images, representing the first publicly available benchmark for deep learning based ISTD. However, targets are synthetically generated via Gaussian sampling, causing labeled masks to be systematically inflated beyond true target boundaries and violating the SPIE small-target definition. These annotation inconsistencies have led subsequent works to advise against its use as a primary benchmark \cite{Survey-10}, though it remains a standard historical baseline for longitudinal comparison. 
\item \textbf{DAUB Dataset (2019):} This dataset developed by Hui et al. \cite{Dataset-1} contains 21 image sequences, 30 trajectories and 16177 frames in relation to 16944 targets. Each target corresponds to a label location in the image, and each image sequence corresponds to a label file.
\item \textbf{NUAA-SIRST Dataset (2021):} The NUAA-SIRST dataset\footnote{https://github.com/YimianDai/sirst} is provided by Dai et al. \cite{DL-10}  with high-quality images and labels. However, the number of images is 427 (256 for training), which cannot well cover dramatic scene changes in infrared small target detection. Moreover, these real infrared data are all manually labeled with many inaccurately labeled pixels. 
\item \textbf{NUDT-SIRST Dataset (2021):} Li et al. \cite{DL-12} developed the NUDT -SIRST dataset with numerous categories of target, various target sizes, diverse clutter backgrounds, and accurate annotations. The NUDT-SIRST dataset contains 5 main background scenes including city, field, highlight, sea, and cloud. Each image is synthesized from real background with various targets (e.g., point, spot, and extended) under various SCR and rich poses. Most of the background images are collected by Li et al. \cite{DL-12}  and  a few field-type background images come from Hui et al. \cite{Dataset-1}. 
\item \textbf{IRSTD-1K Dataset (2022):} The IRSTD-1K dataset \cite{DL-211} contains a collection of 1,000 IR images obtained in real-world scenarios using an IR camera. These images have dimensions of $512 \times 512$ pixels. The targets have been labeled at the pixel level. A variety of small targets are presents such as drones, persons, vehicles and boats. These targets have been captured at different locations, and the images are acquired from considerable imaging distances. The dataset offers a various environments such as natural, urban and maritime under different atmospheric conditions. The backgrounds show notable clutter and noise increasing the difficulty for the detection and recognition of targets. Thus, this dataset is suitable for comprehensive assessments and method benchmarking.
\item \textbf{SISD Dataset (2022):} This dataset\footnote{\url{https://github.com/echo-sky/SISD}} proposed by Wu et al. \cite{Applications-S-44} is an infrared ship dataset in sea environments. 
\item \textbf{DenseSIRST Dataset (2024):} The DenseSIRST Dataset \footnote{https://github.com/GrokCV/DenseSIRST} released by Xiao et al. \cite{PSTD-40} addresses both the clustered target detection challenge and the background semantic modeling requirement. 
\begin{enumerate}
\item Dense Target Scenarios: Transitioning from sparse to clustered target configurations that reflect real-world deployment conditions.
\item Background Semantic Annotations: Offering detailed per-pixel semantic segmentation of background regions, classified into sky and non-sky classes, thereby allowing background semantic modeling. 
\end{enumerate}
\item \textbf{SIRST-V2 Dataset (2023):} The SIRST-V2\footnote{https://github.com/YimianDai/open-sirst-v2} provided by Dai et al. \cite{Dataset-200} includes 515 real infrared images selected from thousands of infrared sequences. Nevertheless, its moderate size still constrains the training of large capacity models without additional augmentation, and the variable resolution images introduce inconsistencies in evaluation protocols across methods that assume a fixed input size \cite{DL-211}. These problems may lead to the inapplicability toward robust SIRST  detection~\cite{PE-1}. 
\item \textbf{NCHU-SIRST Dataset (2023):} The NCHU-SIRST\footnote{https://github.com/PCwenyue/CAFF-Net-method-and-the-NCHU-SIRST-dataset} released by Shi et al. \cite{Dataset-100} is a real-world infrared small target dataset captured using an infrared camera, comprising 590 manually selected images from 6,300 raw infrared photographs. It contains 273 training frames and 317 test frames, covering six background scene categories: architecture, cloudless sky, complex clouds, continuous clouds, sea, and trees. The dataset includes aircraft, birds, and ships as target types, with target sizes ranging from 3×3 to 9×9 pixels in strict conformance with the SPIE small-target definition, providing a reliable real-world evaluation reference for infrared small target detection methods. \\
\item \textbf{MassMIND Dataset (2023)} This dataset\footnote{https://github.com/uml-marine-robotics/MassMIND} \cite{Dataset-500} of real-life Long Wave Infrared (LWIR) images consists of 2916 diverse LWIR images captured in and around the Boston Harbor in Massachusetts, USA over a span of 2 years. The images have been recorded in a busy marine area as well as in not so busy ocean waters during various seasons and times of the day creating seasonal and temporal diversity. Each image has been segmented with pixel level instance segmentation and labeled across 7 different classes. 
\item \textbf{MWIRSTD Dataset (2024):} The MWIRSTD\footnote{https://github.com/avinres/MWIRSTD} proposed by Kumar et al. \cite{Dataset-401} includes 1053 real Mid-Wave Infrared (MWIR) images drawn from 14 video sequences, annotated across three distinct classes of small objects captured using cooled MWIR imagers. 
Its limited scale still constrains the training of large capacity models without additional augmentation, and the fixed resolution of $640 \times 512$ pixels, while consistent, restricts generalization to other sensor configurations and imaging conditions. These problems may lead to inapplicability toward robust SIRST detection in diverse real-world scenarios \cite{Dataset-401}. 
\item \textbf{TSIRMT Dataset (2024):} The TSIRMT\footnote{https://github.com/lifier/LMAFormer} dataset, introduced by Huang et al. \cite{DL-46}, is a multi-frame sequential infrared benchmark designed to evaluate spatiotemporal detectors under conditions where simultaneous target and platform motion produce strong coupling between target signatures and background clutter. 
\item \textbf{DMIST dataset (2024):}
The DMIST dataset\footnote{https://github.com/UESTC-nnLab/DMIST} proposed by Chen et al. \cite{Dataset-300} group two special dense moving target datasets (DMIST-60 and DMIST-100) on DAUB data. They both contain far more than 50 infrared small targets per frame.
\item \textbf{DSISTD Dataset (2024):} The DSISTD dataset\footnote{https://github.com/UESTC-nnLab/DMIST} proposed by Chen et al. \cite{Dataset-300} is a drone swarm infrared small target dataset that integrates real drone targets into simulated infrared backgrounds. 
\item \textbf{SIRST-UAVB Dataset (2024):} The SIRST-UAVB\footnote{https://ieee-dataport.org/documents/sirst-uavb-single-frame-infrared-small-target-dataset-uav-and-birds} dataset, introduced by Yang et al. \cite{DL-104}, comprises 3,000 real single frame mid-wave infrared images captured using a cooled MWIR camera at a fixed resolution of 640$\times$512 pixels and shooting distances ranging from 100 to 800 metres. Targets, consisting of UAVs and birds, account for 94.3\% of the data and are presented against complex backgrounds. Nevertheless, the dataset exclusive focus on two target categories UAVs and birds, restricts its diversity relative to broader SIRST benchmarks. These factors may reduce its applicability toward comprehensive and robust SIRST detection across diverse operational conditions. 
\end{itemize}

\subsubsection{Large-Scale Datasets}
Large-scale datasets covers diverse scenarios in terms of environments, number of targets (sparse/dense) and number images/videos allowing training of data-based models.

\begin{itemize}
\item \textbf{IRDST Dataset (2023):} The IRDST dataset \cite{DL-33} dataset contains 142,727 frames. It includes 40,650 real frames from 85 scenes and 102,077 simulated frames across 317 scenes. Each frame is annotated with three kinds of labels ranging from fine to coarse: pixel-level masks, bounding boxes, and central pixels. Pixel level masks are obtained by labeling each target pixel individually. Bounding box labels are obtained by placing the box two pixels away from the edges of the pixel-level masks. Central pixel labels are determined by identifying the centroid of the pixel-level mask.
\item \textbf{WideIRSTD Dataset (2024):} This dataset\footnote{https://github.com/XinyiYing/WideIRSTD-Dataset} was delivered for the ICPR 2024 Resource-Limited Infrared Small Target Detection Challenge. It consists of seven public datasets: SIRST-V2, IRSTD-1K, IRDST, NUDT-SIRST, NUDT-SIRST-Sea, NUDT-MIRSDT, Anti-UAV, and a dataset developed by the team of National University of Defense Technology (NUDT) including simulated land-based and space-based data, and real manually annotated space-based data. The dataset contains images with various target shapes (e.g., point target, spotted target, extended target), wavelengths (e.g., near-infrared, shortwave infrared and thermal), image resolution (e.g., 256, 512, 1024, 3200, etc.), at varied imaging systems (e.g., land-based, aerial-based and space-based imaging systems). 
\item \textbf{IRSatVideo-LEO Dataset (2025):} The IRSatVideo-LEO\footnote{https://github.com/XinyiYing/IRSatVideo} released by Ying et al. \cite{MFD-13} is a large-scale, semi-simulated dataset designed for multi-frame infrared small target (MIRST) detection in Low Earth Orbit (LEO) satellite videos. It contains 200 image sequences, 91,021 frames, and 218,038 targets. The dataset includes synthesized satellite motion, target appearance, trajectory, and intensity, providing a standard toolbox for satellite video generation and a reliable evaluation platform. 
\item \textbf{IR-SatDense Dataset (2026):} The IR-SatDense Dataset developed by Zhang et al. \cite{DL-501} reflects realistic on-orbit imaging conditions with complex backgrounds. It consists of a total of 2154 background images from public satellite infrared imagery sources. These images cover diverse Earth observation scenarios and represent various background characteristics. All background images were cropped
and preprocessed to a uniform resolution of $512 \times 512$ pixels. According to background complexity, the dataset is categorized into four levels: easy, medium, complex, and extremely complex scenes. Compared to DMIST and DenseSIRST, IR-SatDense dataset provides higher target density and stronger realism. It is obtained with a Dense Single-Frame Target Dataset Generator (DSTDGen) based on real satellite infrared
backgrounds and is organized into multiple subsets corresponding to different density levels measured by the Average Minimum Inter-Target Distance (AMID). 
\end{itemize}

\subsubsection{Air Environments}
The available datasets dedicated to air environments can be listed as follows:
\begin{itemize}
\item \textbf{BSIRST-v1 Dataset (2024):} This dataset developed by Zhang et al. \cite{DL-542} is dedicated for testing algorithms in monitoring of birds in plateau lakes and wetlands. 
\item \textbf{BSIRST-v2 Dataset (2024):} The BSIRST-v2 Dataset \footnote{https://github.com/miliner/FATCNet} developed by Yang et al. \cite{DL-278} consists of  1000 authentic images with various target shapes, sizes, and cluttered backgrounds. The dataset includes rigorous pixel-level annotations.
\item \textbf{MIST Dataset (2026):} The MIST\footnote{https://github.com/GR-ray/MIST} dataset developed by Gao et al. \cite{DL-563} contains airborne infrared detection scenarios. The dataset is obtained with a synthetic data engine that models variations in pose, size, and intensity of moving targets while seamlessly blending them into real backgrounds for physical, geometric, and visual realism. Targets present low signal-to-clutter ratios and complex motion.  
\end{itemize}

\subsubsection{Maritime Environments}
The available datasets dedicated to maritime environements can be listed as follows:
\begin{itemize}
\item \textbf{ISDD Dataset (2021):} The Infrared Ship Detection Dataset (ISDD\footnote{https://github.com/yaqihan-9898/ISDD}) \cite{DL-30} contains 1284 infrared remote sensing images and 3061 ship instances. The image size is $500\times500$. ISSD consists of  images taken by the Landsat8 satellite. There are 373 inshore scenes containing 924 instances and 911 offshore scenes containing 2137 instances. 
\item \textbf{NUDT-SIRST-Sea Dataset (2023):} The NUDT-SIRST-sea Dataset\footnote{https://github.com/TianhaoWu16/Multi-level-TransUNet-for-Space-based-Infrared-Tiny-ship-Detection} developed by Wu et al. \cite{DL-340} consists of images with multiple targets. It lacks wave clutter background images, with over $80\%$ featuring simple backgrounds or land near port \cite{Applications-S-45}.
\item \textbf{IRSDSS Dataset (2024):} This dataset\footnote{https://github.com/greekinRoma/SMPISD-MTPNet} proposed by Hu et al. \cite{Applications-S-42} presents maritimes scenarios for infrared ship  and contains 1491 images. It contains only about $10\%$ of wave clutter and complex cloud formation backgrounds \cite{Applications-S-45}. 
\item \textbf{Maritime-SIRST Dataset (2024)} This dataset \footnote{https://github.com/peerless66/Maritime-SIRST/blob/main/README.md} developed by Guo et al. \cite{Applications-S-45} exploits publicly available images from the near-infrared band of the Landsat-8 remote sensing satellite. Within the Landsat-8 remote sensing images, IR band images have been selected from various regions worldwide, including ports, canals, and open seas across Asia, Africa, and North America. The selected images span from 2013 to 2021, ensuring temporal diversity. Moreover, images with different months and varying cloud cover percentages have been chosen. Maritime-SIRST dataset comprises 1131 images with 2647 targets. Each image sized is of $256 \times 256$ pixels. 
\end{itemize}

\subsection{Visual and Text Modalities Datasets}
Since 2025, visual and text modalities datasets have emerged according to the need of evaluation of large language models and vision-language models applied for infrared small target detection. These datasets can be listed as follows:
\begin{itemize}
\item \textbf{MIRSTD Dataset (2025):} The MIRSTD dataset developed by Zhang et al. \cite{LLM-1} consists of 427 and 1,000 real infrared images with one or more small targets, and 1,327 synthetic infrared images with small targets. All images are resized to $1024 \times 1024$. The images are divided into three disjoint subsets: $50\%$ for training, $30\%$ for validation, and $20\%$ for testing.
\item \textbf{IR-TXPair Dataset (2025):} The IR-TXPair dataset developed by Zhang et al. \cite{LLM-2} is obtained from the IRSTD-1k and NUAA-SIRST datasets and contains 1427 images. It contains various scenes like urban environments, landscapes with mountains, marine environments, and
desert environments, with backgrounds such as buildings, roads, vegetation, sea waves, and sandstorms. In addition, the dataset offers diverses infrared small targets (drones, vehicles, and ships). The dataset is provide with text descriptions for rich multimodal information. For each dataset,$50\%$ of the images are used as the training set and $30\%$ are employed for the validation set, and $20 \%$for the test set.
\item \textbf{FZDT Dataset (2025):} The FZDT dataset introduced by Huang et al. \cite{LLM-3}  consists of 2,755 infrared images of different scenarios with fuzzy semantic textual annotations. More specifically, iy consists of the three public image datasets NUAA-SIRST, NUDT-SIRST, IRSTD-1k, and corresponding fuzzy semantic text prompts. Each dataset is divided into training set, validation set and test set in the ratio of 6:2:2.
\item \textbf{LangIR Dataset (2026):}  The LangIR dataset developed by Chib and Singh \cite{LLM-4} contains two datasets, namely NUAA-SIRST and IRSTD-1k. The NUAA-SIRST subset of the LangIR dataset contains 427 text descriptions and the IRSTD-1k subset contains 1,001 text descriptions. These descriptions include textual information that localizes and describes the small target. In both the NUAA-SIRST and IRSTD-1k subsets, the most frequently occurring positional word is left, indicating that most small targets are located in the left most position of the image, followed by upper and then center. 
\end{itemize}

\begin{table*}[!t]
\centering
\scriptsize
\setlength{\tabcolsep}{3pt}
\renewcommand{\arraystretch}{1.1}
\begin{tabular}{|>{\raggedright\arraybackslash}p{4.2cm}|>{\raggedright\arraybackslash}p{3.4cm}|>{\raggedright\arraybackslash}p{3.0cm}|>{\raggedright\arraybackslash}p{3.3cm}|>{\raggedright\arraybackslash}p{1.9cm}|>{\raggedright\arraybackslash}p{1.0cm}|}
\hline
Datasets (Authors) (Date) & Background Features & Target Features & Images/Videos/Resolution & Single/Multi-Frame & Sparse/\newline Dense \\
\hline
\hline
\multicolumn{6}{|l|}{\textbf{Visual Modality Datasets (28)}} \\
\hline
NUST-SIRST Dataset (Wang et al. \cite{DL-15}) (2019) & Sky, clouds, ground, buildings & Tiny cats, cars, drones, and planes & 100 images, 11 sequences (2,098 frames), $256 \times 256$ & Single/Multi-Frame & Sparse \\
DAUB Dataset (Hui et al. \cite{Dataset-1}) (2019) & Sky and complex environments & Dim fixed-wing UAVs & 22 sequences, 16,177 frames & Multi-Frame & Sparse \\
NUAA-SIRST Dataset (Dai et al. \cite{DL-10}) (2021) & Sea, clouds and city environments & Extremely dim, tiny targets & 427 images, variable resolutions & Single-Frame & Sparse \\
NUDT-SIRST Dataset (Li et al. \cite{DL-12}) (2021) & Cloud, city, sea and field environments & Ships, planes, UAVs & 1,327 synthesized images, $256 \times 256$ & Single-Frame & Sparse \\
ISDD Dataset (Han et al. \cite{DL-30}) (2021) & Complex ocean, inshore and offshore environments & Ships & 1,284 images, $500 \times 500$ & Single-Frame & Mixed \\
IRSTD-1K Dataset (Zhang et al. \cite{DL-211}) (2022) & Rich clutter, diverse scenes & Drones, creatures, vessels, and vehicles & 1,000 images, $512 \times 512$ & Single-Frame & Sparse \\
SISD Dataset (Wu et al. \cite{Applications-S-44}) (2022) & Sea environments & Ships & 150 images, $512 \times 512$ & Single-Frame & Sparse \\
SIRST-V2 Dataset (Dai et al. \cite{Dataset-200}) (2023) & Complex urban scenarios & Dim small spots & 1,024 images, $1280 \times 1024$ & Single-Frame & Sparse \\
NCHU-SIRST Dataset (Shi et al. \cite{Dataset-100}) (2023) & Complex different scenes & Small weak signals & 590 images, $256 \times 256$ & Single-Frame & Sparse \\
MassMIND Dataset (Nirgudkar et al. \cite{Dataset-500}) (2023) & Coastal maritime environments & Inanimate and living obstacles & 2,916 LWIR images, $640 \times 512$ & Single-Frame & Dense \\
NCHU-Seg Dataset (Meng et al. \cite{Dataset-700}) (2023) & Different environments & Aircraft, birds, and ships & 590 images, $256 \times 256$ & Single-Frame & Sparse \\
IRDST Dataset (Sun et al. \cite{DL-33}) (2023) [Large-Scale] & Terrestrial and aerial environments & Dim aircraft & 142,727 frames, 5,930 videos, $512 \times 512$ & Multi-Frame & Sparse \\
NUDT-SIRST-Sea Dataset (Wu et al. \cite{DL-340}) (2023) & Ocean environments & Extremely small, dim ships & 48 real images, $10000 \times 10000$ & Single-Frame & Sparse \\
DenseSIRST Dataset (Xiao et al. \cite{PSTD-40}) (2024) & Aerial, maritime, and terrestrial environments & Densely clustered tiny spots & 1,024 images, $512 \times 512$ & Single-Frame & Dense \\
IRSTDID Dataset (Yuan et al. \cite{Dataset-600}) (2024) & Sky and ground environments & UAVs & 800 images & Single-Frame & Sparse \\
MWIRSTD Dataset (Kumar et al. \cite{Dataset-401}) (2024) & Terrestrial environments & Rockets, debris and vehicles & 1,053 images, $640 \times 512$ & Multi-Frame & Sparse \\
TSIRMT Dataset (Huang et al. \cite{DL-46}) (2024) & Terrestrial/aerial backgrounds & Aircraft, UAVs, ships, vehicles & 20,000 images, 200 videos, $150 \times 200$ & Multi-Frame & Sparse \\
DMIST Dataset (Chen et al. \cite{Dataset-300}) (2024) & Real ground/air environments & Synthetic moving spots & 13,778 images, $256 \times 256$ & Multi-Frame & Dense \\
DSISTD Dataset (Chen et al. \cite{Dataset-300}) (2024) & Synthetic environments & Real targets & -- & Single-Frame & Dense \\
SIRST-UAVB Dataset (2024) & Terrestrial and aerial backgrounds & UAVs and birds & 3,000 images, $640 \times 512$ & Single-Frame & Sparse \\
WideIRSTD Dataset (Li et al. \cite{Dataset-400}) (2024) [Large-Scale] & SIRST-V2, IRSTD-1K, IRDST, NUDT-SIRST, NUDT-SIRST-Sea, NUDT-MIRSDT, Anti-UAV & Diverse targets & Variable size & Single/Multi-Frame & Sparse \\
BSIRST-v1 Dataset (Zhang et al. \cite{DL-542}) (2024) & Plateau lakes and wetland & Birds & 427 images, $512 \times 512$ & Single-Frame & Sparse \\
BSIRST-v2 Dataset (Yang et al. \cite{DL-278}) (2024) & Plateau lakes and wetland & Birds & 1,000 images, $512 \times 512$ & Single-Frame & Sparse \\
IRSDSS Dataset (Hu et al. \cite{Applications-S-42}) (2024) & Inshore and offshore sea scenes & Ships & 1,491 images, $640 \times 640$ & Single-Frame & Sparse \\
Maritime-SIRST Dataset (Guo et al. \cite{Applications-S-45}) (2024) & Sea environments (diverse scenes) & Dim, textureless small ships & 1,131 images, $256 \times 256$ & Single-Frame & Sparse \\
IRSatVideo-LEO Dataset (Ying et al. \cite{MFD-13}) (2025) [Large-Scale] [Satellite] & Terrestrial and sea environments & Maneuvering aircraft and missile spots & 91,021 images, 200 videos, $1024 \times 1024$ & Multi-Frame & Sparse \\
IR-SatDense Dataset (Zhang et al. \cite{DL-501}) (2026) [Large-Scale] [Satellite] & Complex satellite backgrounds & Extremely tiny, low-SNR targets & 2,154 images, $512 \times 512$ & Single-Frame & Dense \\
MIST Dataset (Gao et al. \cite{DL-563}) (2026) & Airborne environments & Low-SCR targets & 120 video sequences & Multi-Frame & Sparse \\
\hline
\hline
\multicolumn{6}{|l|}{\textbf{Visual and Text Modality Datasets (4)}} \\
\hline
MIRSTD Dataset (Zhang et al. \cite{LLM-1}) (2025) & Sky, sea, and land environments & Dim targets, aircraft and ships & 1,000 images, $256 \times 256$ & Single-Frame & Sparse \\
IR-TXPair Dataset (Zhang et al. \cite{LLM-2}) (2025) & IRSTD-1K, NUAA-SIRST & Drones, airplanes, vehicles, and ships & 2,754 images, $1024 \times 1024$ & Single-Frame & Sparse \\
FZDT Dataset (Huang et al. \cite{LLM-3}) (2025) & NUAA-SIRST, NUDT-SIRST, IRSTD-1K & Dim, tiny targets & 2,755 images, $256 \times 256$ & Single-Frame & Sparse \\
LangIR Dataset (Chib and Singh \cite{LLM-4}) (2026) & NUAA-SIRST, IRSTD-1K & Idem & 427 images, 1,001 images & Single-Frame & Sparse \\
\hline
\end{tabular}
\caption{Existing Datasets in Infrared Small Target Detection: An Overview (32 Datasets).}
\label{OverviewDatasets}
\end{table*}

\section{Performance Evaluation}
\subsection{Conventional Protocols}
Conventional evaluation protocols have been established over several decades, originally within signal processing and subsequently adapted to the deep learning paradigm.  They are characterized by their reliance on either \emph{pixel-level} or \emph{block-level} (target-level) computational primitives, reflecting the two dominant formulations of the detection problem: as a per-pixel binary segmentation task or as a target-instance localisation task, respectively.

\subsubsection{Pixel-Level Metrics}
Pixel-level metrics treat IRSTD as a dense binary segmentation problem, evaluating the agreement between the predicted mask and the ground truth at individual pixel resolution.Each metric derives from the four entries of the binary confusion matrix: $\mathrm{TP}$, $\mathrm{FP}$, $\mathrm{TN}$,
and $\mathrm{FN}$ \cite{PE-1}. From these quantities, a set of metrics can be derived as follows.

\begin{itemize}
\item \textbf{Intersection over Union($\mathrm{IoU}$):}
The common pixel-level metric is the Intersection over Union (IoU). Also known as the Jaccard index, $\mathrm{IoU}$ measures the pixel wise overlap between the predicted and ground truth masks:
\begin{equation}
  \mathrm{IoU}
  = \frac{\mathrm{TP}}
         {\mathrm{TP} + \mathrm{FP} + \mathrm{FN}}.
  \label{eq:iou}
\end{equation}
It is the primary segmentation metric across IRSTD
benchmarks. Its main limitation is a size
aggregation bias. Since statistics are pooled globally before computing the ratio, large or numerous targets dominate the score while small, sparse targets contribute
negligibly \cite{PE-1}. \\

\item \textbf{Normalised IoU ($\mathrm{nIoU}$):}
To correct the aggregation bias of Equation \eqref{eq:iou},
the normalised IoU computes the ratio per sample before
averaging across the dataset:
\begin{equation}
  \mathrm{nIoU}
  = \frac{1}{N}\sum_{n=1}^{N}
    \frac{\mathrm{TP}_{n}}
         {\mathrm{TP}_{n}+\mathrm{FP}_{n}+\mathrm{FN}_{n}}.
  \label{eq:niou}
\end{equation}
This formulation ensures equal contribution from all samples regardless of target size. \\

\item \textbf{F1-Score / Dice ($\mathrm{F1}$):}
The pixel-level F1-score is the harmonic mean of pixel-level precision and recall:
\begin{equation}
  \mathrm{F1}
  = \frac{2\,\mathrm{TP}}
         {2\,\mathrm{TP}+\mathrm{FP}+\mathrm{FN}}.
  \label{eq:f1pix}
\end{equation}
While both $\mathrm{IoU}$ and $\mathrm{F1}$ mitigate foreground imbalance through relative area proportions, they remain limited to global pixel-wise analysis and cannot capture target level spatial localisation \cite{PE-1,SFD-LCM-90}. \\

\item \textbf{Pixel Accuracy (PA):}
\begin{equation}
  \mathrm{PA}
  = \frac{\mathrm{TP}+\mathrm{TN}}{N_{\mathrm{total}}}.
  \label{eq:pa}
\end{equation}
PA is rarely used as a primary metric in IRSTD. Since target pixels account for less than $0.1\%$ of the image, a null predictor achieves $\mathrm{PA}>99.9\%$, rendering
the metric uninformative under such extreme class
imbalance. \\

\item \textbf{Mean Pixel Accuracy (MPA):}
\begin{equation}
  \mathrm{MPA}
  = \frac{1}{C}\sum_{c=1}^{C}
    \frac{\mathrm{TP}_{c}}{\mathrm{TP}_{c}+\mathrm{FN}_{c}}.
  \label{eq:mpa}
\end{equation}
By computing per-class recall and averaging across the target and background classes, MPA directly corrects the imbalance situation of PA \cite{Survey-6}. \\

\item \textbf{Pixel-level Precision and Recall:}
Precision and recall at the pixel level are defined as:
\begin{equation}
  \mathrm{Prec}
  = \frac{\mathrm{TP}}{\mathrm{TP}+\mathrm{FP}},
  \label{eq:prec}
\end{equation}
\begin{equation}
  \mathrm{Rec}
  = \frac{\mathrm{TP}}{\mathrm{TP}+\mathrm{FN}}.
  \label{eq:rec}
\end{equation}
Both are reported in the work of Ni et al. \cite{DL-231}. Notably, when computed at the target level, $\mathrm{Rec}$ coincides with the probability of detection $P_d$.\\

\item \textbf{PSNR and SSIM:}
Two additional metrics assess the quality of background estimation and image reconstruction. The Peak Signal-to-Noise Ratio (PSNR) quantifies the reconstruction fidelity of the background model in decibels:
\begin{equation}
  \mathrm{PSNR}
  = 10\log_{10}\!\left(\frac{\mathrm{MAX}^{2}}{\mathrm{MSE}}\right)
  \quad [\mathrm{dB}],
  \label{eq:psnr}
\end{equation}
where $\mathrm{MAX}$ is the maximum possible pixel intensity
and $\mathrm{MSE}$ is the mean squared error between the
estimated and reference background. Complementarily, the
Structural Similarity Index Measure (SSIM) evaluates
perceptual fidelity by jointly assessing luminance, contrast,
and structural consistency between two images:
\begin{equation}
  \mathrm{SSIM}(x,y)
  = l(x,y)\cdot c(x,y)\cdot s(x,y).
  \label{eq:ssim}
\end{equation}
Both PSNR and SSIM are restricted to evaluating the fidelity of background suppression models and carry no direct interpretation as detection performance
measures.
\end{itemize}

\subsubsection{Target-Level Metrics}
The common target-level metrics are the Probability of detection (Pd) and the False-alarm rate (Fa). A set of block-level metrics can be derived as follows:
\begin{itemize}
\item \textbf{Probability of Detection ($P_d$):}
The most fundamental block-level metric is the probability of detection, which measures the fraction of ground truth targets successfully identified:
\begin{equation}
  P_d  = \frac{\mathrm{TP}}{\mathrm{TP} + \mathrm{FN}}.
  \label{eq:pd}
\end{equation}
$P_d$ measures detection completeness but ignores false positives, potentially overestimating performance in noisy environments \cite{PE-1}. It is the primary operational metric across the entire IRSTD literature \cite{Survey-6,DL-211,DL-12,SFD-LCM-90}.\\

\item \textbf{False Alarm Rate ($F_a$):}
Complementary to $P_d$, the false alarm rate quantifies
the spatial extent of false positive detections by
normalising their total area to the full image size:
\begin{equation}
  F_a = \frac{|\mathrm{area}(\mathrm{FP})|}{N_{\mathrm{pixels}}}
  \quad [\times10^{-6}\ \mathrm{per\ pixel}].
  \label{eq:fa}
\end{equation}
While $F_a$ captures the spatial footprint of false alarms, it may undercount small false positive instances while over penalising large ones. Additionally, $F_a$ disregards
pixel-level errors within correctly matched
targets \cite{Survey-6, DL-211, DL-12,SFD-LCM-90}. \\

\item \textbf{False Alarm Rate per Frame (FAPF):}
Rather than measuring the spatial area of false alarms,
FAPF counts the mean number of false detection events
per frame:
\begin{equation}
  \mathrm{FAPF}
  = \frac{N_{\mathrm{FP}}}{N_{\mathrm{frames}}}.
  \label{eq:fapf}
\end{equation}
This count-based variant is particularly informative in sequential detection scenarios where tracker workload depends on the number of spurious instances rather than their area \cite{Survey-6}. \\

\item \textbf{Target-level Precision and F1:}
Analogously to their pixel-level counterparts, precision
and F1 can be formulated at the target instance level:
\begin{equation}
  P_{\mathrm{tgt}}
  = \frac{\mathrm{TP}}{\mathrm{TP}+\mathrm{FP}},
  \label{eq:ptgt}
\end{equation}
\begin{equation}
  \mathrm{F1}
  = \frac{2\,P_{\mathrm{tgt}}\cdot P_d}
         {P_{\mathrm{tgt}}+P_d}.
  \label{eq:f1tgt}
\end{equation}
Notably, the recall in this context corresponds directly to the target-level $P_d$ defined in
Equation \eqref{eq:pd} \cite{PE-1,DL-231}. \\

\item \textbf{ROC Curve and AUC:}
The Receiver Operating Characteristic (ROC) curve plots $P_d$ as a function of $F_a$ across all detection thresholds, providing a threshold-independent view of the detection trade-off. The Area Under the Curve (AUC)
summarizes this trade-off in a single scalar:
\begin{equation}
  \mathrm{AUC}
  = \int P_d(t)\,\mathrm{d}[F_a(t)].
  \label{eq:auc}
\end{equation}
The larger the AUC, the better the overall detection
performance. The ROC curve is the standard evaluation
tool for all filter-based and model-driven IRSTD
methods \cite{Survey-6}. \\

\item \textbf{Signal-to-Clutter Ratio (SCR):}
Before any processing, the inherent detectability of a target in a given scene is characterised by the
Signal-to-Clutter Ratio (SCR):
\begin{equation}
  \mathrm{SCR}
  = \frac{\mu_t - \mu_{\mathrm{bg}}}{\sigma_{\mathrm{bg}}},
  \label{eq:scr}
\end{equation}
where $\mu_t$ is the mean intensity of the target region
and $\mu_{\mathrm{bg}}$, $\sigma_{\mathrm{bg}}$ are the
local background mean and standard deviation computed in
a surrounding window \cite{PE-2, SFD-RPCATensor-73, SFD-LCM-2}. SCR is widely used for dataset difficulty characterisation and serves as
the baseline reference quantity against which the
enhancement capability of an algorithm is measured,
as detailed in the following metric. \\

\item \textbf{Signal-to-Clutter Ratio Gain (SCRG):}
Building on the SCR definition in Equation \eqref{eq:scr},
the SCRG measures the degree to which an algorithm
amplifies the target signal relative to background
clutter:
\begin{equation}
  \mathrm{SCRG}
  = \frac{\mathrm{SCR}_{\mathrm{out}}}
         {\mathrm{SCR}_{\mathrm{in}}}.
  \label{eq:scrg}
\end{equation}
Larger values indicate stronger target enhancement.
It is important to note, that a high SCRG
does not guarantee successful detection; a proper
thresholding strategy must follow \cite{PE-2}. \\

\item \textbf{Background Suppression Factor (BSF):}
Complementary to SCRG, the BSF measures how effectively
the algorithm flattens background clutter texture,
independently of target enhancement:
\begin{equation}
  \mathrm{BSF}
  = \frac{\sigma_{\mathrm{bg,in}}}
         {\sigma_{\mathrm{bg,out}}}.
  \label{eq:bsf}
\end{equation}
SCRG and BSF are always reported jointly, as together
they distinguish between two complementary aspects of
background suppression: target contrast enhancement
and clutter flattening \cite{PE-2,Survey-6,SFD-LCM-104,SFD-LCM-105}. \\
\item \textbf{Signal-to-Noise Ratio (SNR)}
A global variant of SCR, the Signal-to-Noise Ratio
replaces the local background window with image-wide
noise statistics:
\begin{equation}
  \mathrm{SNR}
  = \frac{\mu_t - \mu_{\mathrm{global}}}
         {\sigma_{\mathrm{global}}}.
  \label{eq:snr}
\end{equation}
Although less precise than SCR in characterising
structured clutter, SNR is used in early IRSTD
literature and hardware-level system characterisation \cite{Survey-6,RT-10}. \\

\item \textbf{Local Contrast Measure (LCM):}
Motivated by the human visual system, the Local
Contrast Measure quantifies target saliency relative
to its immediate spatial neighborhood:
\begin{equation}
  \mathrm{LCM}(x,y)
  = \frac{I_{\mathrm{target}}}
         {\mu_{\mathrm{neighborhood}}}.
  \label{eq:lcm}
\end{equation}
Multiple variants exist in the literature, including
the nonlinear NLCM and the revised RLCM, each
modifying the neighborhood weighting \cite{Survey-6,SFD-LCM-106,SFD-LCM-72}. \\

\item \textbf{Centroid Localisation Error (CLE):}
Rather than measuring overlap or signal quality, the
CLE directly quantifies the spatial accuracy of a
detection by computing the Euclidean distance between
the predicted and ground-truth target centroids:
\begin{equation}
  \mathrm{CLE}
  = \sqrt{(\hat{x}-x)^{2}+(\hat{y}-y)^{2}}
  \quad [\mathrm{pixels}].
  \label{eq:cle}
\end{equation}
Mean CLE is typically reported per sequence in
tracking evaluations, where sub-pixel
accuracy is the target \cite{Survey-6, DL-231, SFD-RPCAMatrix-46}. \\

\item \textbf{Bounding-Box IoU Variants:}
For anchor based detection architectures, several
extensions of the standard IoU have been adapted to
the IRSTD setting to improve bounding box regression
for small targets. The Generalised IoU (GIoU)
penalises non-overlapping boxes through an enclosing
box term. The Distance IoU
(DIoU) additionally penalises the centroid offset
between predicted and ground-truth boxes. The Complete
IoU (CIoU) further incorporates aspect ratio
consistency \cite{Survey-6}. Finally, the Normalised
Gaussian Wasserstein Distance (NWD) replaces the
geometric overlap entirely by modeling each target
as a Gaussian distribution and measuring the
Wasserstein distance between them, which is more
suitable for targets with near-zero
overlap:
\begin{equation}
  \mathrm{NWD}(\mathcal{N}_p,\mathcal{N}_g)
  = \exp\!\left(-\frac{\sqrt{W_2^2(\mathcal{N}_p,\mathcal{N}_g)}}{C}\right),
  \label{eq:nwd}
\end{equation}
where $W_2$ is the Wasserstein-2 distance between
the two Gaussian representations and $C$ is a
normalisation constant. \\

\item \textbf{Mean Average Precision (mAP):}
Adopted from the general object detection literature,
mAP summarises the area under the precision--recall
curve across IoU thresholds:
\begin{equation}
  \mathrm{mAP}
  = \frac{1}{|\mathcal{T}|}\sum_{\tau \in \mathcal{T}}
    \int_0^1 P(R,\tau)\,\mathrm{d}R,
  \label{eq:map}
\end{equation}
where $\mathcal{T}$ is the set of IoU thresholds and
$P(R,\tau)$ is the precision at recall $R$ under
threshold $\tau$. mAP is used in anchor-based
YOLO-family and DETR-family IRSTD detectors and is
not applicable to segmentation-only
pipelines \cite{Survey-6}. \\
\end{itemize}

\subsubsection{Complexity Metrics}
Floating-point operations (FLOPs) and parameters (Params) are metrics employed to assess the complexity of deep neural networks models. These metrics allows to understand the computational demands and scale of a model, thereby aiding in model selection, optimization, and deployment. \\

\begin{itemize}
\item FLOPs indicates the number of floating-point operations a model performs for one forward pass. The magnitude of FLOPs reflects the complexity of a model. More the model has a higher number of FLOPs the more computationally complex the model is.
\item Params indicates the number of parameters the model needs to learn. The number of Params reflects the size of a model, and smaller models involve lower computational burdens.
\end{itemize}

\subsection{Common Ranking Protocols}
The linear weighted summation of IoU and Pd is used for ranking in the ICPR 2024 Resource-Limited Infrared Small Target Detection Challenge. Even if this ranking protocol appears efficient to obtain one best method, it is proved that this kind of ranking is not suitable and does not reflect the real respective performance \cite{MOD-Perspectives-100,MOD-Perspectives-101,MOD-Perspectives-102,MOD-Perspectives-103}. Indeed, the performance of a method depend of its environmental application. In addition, a method does not need to be the best in each challenge to be suitable for one kind of application. Thus, alternative ranking protocols need to be designed to provide a fair recommendation to the engineers. \\

\subsection{Disruptive Protocols}
The conventional evaluation protocols present limitations. First, they rely on fragmented pixel-level and target-level specific metrics, which fails to provide a comprehensive view of model capabilities. Second, an excessive emphasis on overall performance scores do not allows error analysis, which is crucial to identify failure modes and to improved real-world system performance. Third, the field predominantly adopts dataset-specific training-testing paradigms, hindering the understanding of model robustness and generalization across diverse infrared scenarios. One way to change the protocol evaluation is to change the metrics. An other way is to change the ranking protocol itself. 
\subsubsection{Hybrid-Level Metric}
Conventional IRSTD evaluation assesses pixel-level and target-level 
performance independently, creating a mismatch where missing a target and miscalculating its shape are never jointly penalised \cite{PE-1}. In 2025, Pang et al. \cite{PE-1} address this by proposing Hierarchical IoU (hIoU), a hybrid-level metric that unifies both aspects within a single score, supported by an open-source 
toolkit.\footnote{\url{https://github.com/lartpang/PyIRSTDMetrics}}.
hIoU jointly captures target-level localisation and pixel-level segmentation via the \emph{Overlap Priority with Distance Compensation} (OPDC) matching strategy \cite{PE-1}:
\begin{equation}
  \mathrm{hIoU}
  = \mathrm{IoU}^{\mathrm{loc}}_{\mathrm{tgt}}
    \times
    \mathrm{IoU}^{\mathrm{seg}}_{\mathrm{pix}},
  \label{eq:hiou}
\end{equation}
where $\mathrm{IoU}^{\mathrm{loc}}_{\mathrm{tgt}}$ and 
$\mathrm{IoU}^{\mathrm{seg}}_{\mathrm{pix}}$ denote target-level 
localisation and pixel-level segmentation IoU, respectively. The 
multiplicative form ensures that weakness in either component cannot be masked by the other \cite{PE-1}.\\

\subsubsection{Holistic SIRST Evaluation Metric}
In 2026, Yu et al. \cite{VLM-2} design an Holistic SIRST Evaluation (HSE) metric that performs multi-threshold integral evaluation at both pixel-level confidence and target-level robustness. More specifically, this metric consists of two complementary sub-metrics:
\begin{itemize}
\item \textbf{Pixel-level Average Precision ($\mathrm{HSE\text{-}P}$):}
HSE-P quantifies pixel-level confidence quality by 
computing the area under the precision recall curve derived from all 
flattened prediction ground-truth pairs \cite{VLM-2}:
\begin{equation}
  \mathrm{HSE\text{-}P} = \int_{0}^{1} P(R)\, \mathrm{d}R,
  \label{eq:hsep}
\end{equation}
where $P(R)$ denotes precision as a function of recall. HSE-P is more sensitive to false positive pixels under the 
extreme class imbalance inherent to IRSTD \cite{VLM-2}.\\

\item \textbf{Target-level Average Precision ($\mathrm{HSE\text{-}T}$):}
HSE-T quantifies target-level detection robustness 
via discrete integration of precision recall pairs across a threshold 
set $\mathcal{T} = \{t_1, t_2, \ldots, t_M\}$ \cite{VLM-2}:
\begin{equation}
  \mathrm{HSE\text{-}T} = \sum_{j=1}^{M} P(t_j)\,
  \bigl[R(t_j) - R(t_{j+1})\bigr],
  \label{eq:hset}
\end{equation}
where $P(t_j)$ and $R(t_j)$ denote precision and recall at threshold 
$t_j$, obtained via centroid-distance of target matching.\\
\end{itemize}
The two sub-metrics are fused multiplicatively into the final unified 
score:
\begin{equation}
  \mathrm{HSE} = \mathrm{HSE\text{-}P} \times \mathrm{HSE\text{-}T},
  \label{eq:hse}
\end{equation}
ensuring that neither pixel-level precision nor target-level 
detectability can mask deficiencies in the other \cite{VLM-2}.\\

\subsection{Protocols for Data-based Methods}
In addition to these conventional protocols, specific protocols are used for deep neural network supervised models; they are presented in "Traditional and Recent Approaches for Infrared Small Target Detection: Overview, Challenges and Perspectives (Part 2)". 

\section{Prospective Research}
We first provide an analysis of the current methods and evaluation protocols. Second, we review the current solved and unsolved challenges. Third, we provide theoretical perspectives for each category of methods.

\subsection{Analysis of Current Methods and Evaluation Protocols}
Even if many progress have been made, there are still improvements to investigate according to the analysis of strengths and weaknesses of the current methods. Regarding the current methods, the following remarks can be made as described as follows:
\begin{enumerate}
\item \textbf{Conventional Unsupervised Methods (Model-based):}
Filtering based approaches efficiently separate targets from the background
using prior knowledge but require manual tuning in complex scenes, limiting adaptability and potentially causing target loss when background noise varies. Visual-saliency-based methods improve the target contrast but struggle with complex backgrounds containing similar distractors, introducing noise that affects precise detection. All these methods are researcher-based methods using crisp, statistical or fuzzy concepts designed in the context of infrared small target detection. \\
\item \textbf{Low-rank and Sparse Decomposition Unsupervised Models (Model-based): }
Sparse and low-rank representation methods model targets as sparse components and backgrounds as low-rank structures, leveraging image properties for robust detection. They integrate spatial–temporal information but are computationally intensive, posing challenges for high dimensional data. Each method has strengths and limitations, influencing its effectiveness in different environments. However, all the variants of the IPI method are researcher-based improvements that choose or design dedicated norms and miminization  problems to handle specific challenges. \\
\item \textbf{Evaluation Protocols:} About the conventional evaluation protocols, several weaknesses appear to have a fair evaluation and comparison for ranking. First, they rely on fragmented pixel- and target-level specific metrics and thus they fail to provide a comprehensive view of model capabilities. Second, the emphasis on overall performance scores do not allow a precise error analysis required to identify failure modes and improve real-world system performance. Third, the employed dataset are dataset-specific training-testing paradigm and to no allow the understanding of model robustness and generalization across diverse infrared scenarios.
\end{enumerate}

\subsection{Solved and Unsolved Challenges}
Challenges can be investigated either according to single-frame/multi-frame detection and to sparse/dense targets detection as follows:
\begin{enumerate}
\item \textbf{Challenges in Single-frame/Multi-frame Detection:} Even if many progress have been made, there are still unsolved challenges both in single-frame and multi-frame detection. \\
\item \textbf{Challenges in Sparse/Dense Target Detection:}
In 2026, existing methods present the following limitations as developed by Zhang et al \cite{DL-501}:
\begin{itemize}
\item \textbf{Local spatial saliency limitation:} Most existing approaches employed local spatial saliency, improving targets by exploiting intensity contrast between a target and its surrounding background. When multiple adjacent targets are densely distributed in the same region, these methods struggle to separate subtle energy differences between targets and background, resulting in target adhesion, blurred responses, and incomplete
separation in the detection maps. 
\item \textbf{Density-aware semantic guidance deficiency:} Even if deep neural networks models employ powerful backbones and transformer-based heads, the backbone feature extraction process is bottom-up and does not explicitly encode target density. Features at different spatial locations are processed in a homogeneous manner, regardless of whether they
belong to dense target clusters or mostly background. 
\item \textbf{Dense-target structural limitation:} Existing sparse-target models do not explicitly model the structural hierarchy of densely distributed targets in the spatial domain. In multi target scenes with high spatial density, they generate ambiguous spatial structures and
mutual feature interference, which conducts to degraded localization accuracy and unstable detection performance.
\end{itemize}
These three limitations show the current unsolved challenges in the case of dense scenes.
\end{enumerate}

\subsection{Prospective Solutions}
\subsubsection{Perspectives for Researcher Solutions}
\begin{enumerate}
\item \textbf{Conventional Unsupervised Methods (Model-based):} This category seems to have fully exploited its potential. Nevertheless, recent methods based on hyperbolic clustering have shown high performance without training data. In addition, Dempster-Schäfer theory has not been investigated in infrared small target detection. \\
\item \textbf{Low-rank and Sparse Decomposition Unsupervised Models (Model-based):} Outliers can be classified in the subspace where they are the more prominent as developed in Brahma et al.\cite{MOD-Theory-2-1}:
\begin{itemize}
\item Outliers in the original observation space (OS) (that are commonly addressed) are relatively visible due to their outlying values. They can be effectively handled by PCP.
\item Outliers in the orthogonal complement (OC) subspace (which is the space orthogonal to the primary principal component subspace) for which PCP may fail. OC outliers are observations that have arbitrarily large magnitude when projected onto the OC subspace.
\end{itemize}
In literature, all the current works using LRSD for infrared small target detection only considered OS outliers. To address this issue, future works may concern a dedicated method for OC outliers or both OS and OC outliers in the low-rank component. Furthermore, outliers can affect both the low-rank and sparse components. Most of the works only considered outliers in the low-rank components. Future works  may also concern dedicated methods for outliers in the sparse components or for the simultaneous presence of both sources of corruption.  \\
\end{enumerate}

\subsubsection{Perspectives for AI-Assisted Research Solutions}
Previous solutions are designed by researchers using mathematical tools, machine learning tools and signal processing tools. In this context, researchers attempt to develop methods using these modeling tools acting on their characteristics according to the challenges they wish to handle. Low-rank and sparse decomposition based methods differ from their decomposition, their minimization problems, their loss functions and their solvers. Deep neural networks methods differ from their conceptual model (CNNs, Transformers,...), their architecture, their optimization, their loss function and their training. In each case, researchers propose a designed combination of the corresponding characteristics of the chosen modeling tools. However, it is not sure that all the combinations have been investigated to obtain the most suitable combination for the infrared small target detection task. In this context, an AI-Assisted research tools addressing this issue would be helpful.

\subsection{Perspectives for Performance Evaluation}
For performance evaluation and fair comparison, Pang et al. \cite{PE-1} propose an hybrid-level metric incorporating pixel- and target-level performance. Thus, a systematic error analysis method is proposed emphasizing the importance of cross-dataset evaluation. An open-source toolkit is released to facilitate standardized benchmarking. \\

Moreover, performance evaluation has to be revisited in the context of high activities research and the diversity of deployment of infrared sensors. Indeed, the research of one top method is not relevant in the context because numerous methods can be more suitable depending the environments. In this context, recent works \cite{MOD-Perspectives-100,MOD-Perspectives-101,MOD-Perspectives-102} on performance evaluation conducted in the close field of moving object detection can alleviate the biases of the current evaluation protocols both for ranking according to the diversity of environments and also in unseen scenarios \cite{MOD-Perspectives-103}.

\section{Conclusion}
\label{Conclusion}
In this paper, we have firstly presented a full review about infrared small target detection by covering 30 years of research. In this context, we categorized the existing methods into two main categories, namely traditional unsupervised models and low-rank and sparse decomposition unsupervised models, deep learning models being reviewed in Part 2. For each categories, the methods have been surveyed in a rationale and historical manner with the corresponding first works, the similar research, their variants/improvements and their last developments. Moreover, we have presented the different datasets, the metrics and the ranking protocols employed for performance evaluation. Finally, we provide theoretical perspectives as well as AI-assisted research solutions.

\bibliographystyle{IEEEtran}
\bibliography{1-SFD,2-MFD,3-RSTD,4-GraphMethods,5-PSTD,6-RealTime,7-Applications,10-Survey,11-PerformanceEvaluation,12-Datasets,20-BackgroundSubtraction,21-Perspectives,30-Miscellaneous,31-Sensors}

\end{document}